\relax
\documentclass[letterpaper]{article} 

\usepackage{styleis}  
\usepackage{times}  
\usepackage{helvet} 
\usepackage{courier}  
\usepackage[hyphens]{url}  
\usepackage{graphicx} 
\usepackage{natbib}  
\usepackage{caption} 

\usepackage{helvet} 
\usepackage{courier}  
\usepackage[hyphens]{url}  
\usepackage{times}
\usepackage{amsmath}
\usepackage{amssymb}
\usepackage{amsopn}
\usepackage{epsfig}
\usepackage{microtype}
\usepackage{subfigure}
\usepackage{booktabs}
\usepackage{booktabs}
\usepackage{amsfonts}
\usepackage{nicefrac}
\usepackage{microtype}
\usepackage{algorithm}
\usepackage{algorithmic}
\usepackage{xcolor}
\usepackage{nicefrac}
\usepackage{stmaryrd}
\usepackage{lipsum}
\usepackage{rotating}
\usepackage{phonetic}
\usepackage{fge}
\usepackage{yfonts}

\let\origbar\bar
\let\origdot\dot
\let\origstar\star
\usepackage{allrunes}
\let\bar\origbar
\let\dot\origdot
\let\star\origstar

\everypar{\looseness=-1}

\newenvironment{proof}{\paragraph{Proof:}}{\hfill$\square$}

\newtheorem{def_krein}{Definition}
\newtheorem{def_rkks}[def_krein]{Definition}

\newtheorem{th_indefrep}{Theorem}
\newtheorem{th_pushforward}[th_indefrep]{Lemma}
\newtheorem{th_collapse}[th_indefrep]{Theorem}
\newtheorem{cor_collapse}[th_indefrep]{Theorem}
\newtheorem{cor_collapsed}[th_indefrep]{Theorem}

\newtheorem{th_kernels}[th_indefrep]{Theorem}
\newtheorem{th_boundsvm}[th_indefrep]{Theorem}
\newtheorem{th_tightboundnn}[th_indefrep]{Theorem}
\newtheorem{th_lownormbound}[th_indefrep]{Theorem}
\newtheorem{cor_sparsity}[th_indefrep]{Corollary}

\newtheorem{th_feature_krein}[th_indefrep]{Theorem}

\newcommand{\tsp}[0]{{\rm T}}

\newcommand{\lip}[0]{\langle}
\newcommand{\rip}[1]{\rangle_{#1}}
\newcommand{\liip}[0]{[}
\newcommand{\riip}[1]{]_{#1}}

\newcommand{\lmip}[0]{\langle\!\langle}
\newcommand{\rmip}[1]{\rangle\!\rangle_{#1}}
\newcommand{\lmiip}[0]{[\![}
\newcommand{\rmiip}[1]{]\!]_{#1}}

\newcommand{\featmap}[0]{\phi}
\newcommand{\featmapmkern}[0]{\varphi}

\newcommand{\dsum}[0]{{{\scriptstyle{[}}\!\!+\!\!{\scriptstyle{]}}}\,}

\newcommand{\infset}[1]{{\mathbb{#1}}}

\newcommand{\NN}[0]{{\mbox{\tiny\tt NN}}}
\newcommand{\SVM}[0]{{\mbox{\tiny\tt SVM}}}
\newcommand{\NTK}[0]{{\mbox{\tiny\tt NTK}}}
\newcommand{\NNSVM}[0]{{\mbox{\tiny\tt M}}}

\newcommand{\rkhs}[1]{\mathcal{H}_{#1}}
\newcommand{\subrkhs}[1]{\mathcal{H}_{#1}}

\newcommand{\frkhs}[1]{\mathcal{F}_{#1}}

\newcommand{\rkks}[1]{\mathcal{K}_{#1}}
\newcommand{\subrkks}[1]{\mathcal{K}_{#1}}
\newcommand{\subsubrkks}[1]{\mathcal{K}_{#1}}
\newcommand{\frkks}[1]{\mathcal{F}_{#1}}

\newcommand{\unitnormballkrein}[1]{\textara{f}_{\!#1}}

\newcommand{\RKHS}[0]{RKHS}

\newcommand{\RKKS}[0]{RKKS}

\newcommand{\krein}[0]{Kre{\u{\i}}n}

\newcommand{\bg}[1]{{\mbox{\boldmath $#1$}}}

\newcommand{\loss}[0]{\ell}
\newcommand{\reg}[0]{r}

\DeclareMathOperator\sgn{sgn} 
\DeclareMathOperator\erf{erf} 

\newcommand{\nnregscale}[0]{\frac{1}{d}}

\newcommand{\GM}[0]{{\rm GM}}

\definecolor{Gray}{gray}{0.9}

\title{From deep to Shallow: Equivalent Forms of Deep Networks in Reproducing Kernel {\krein} Space and Indefinite Support Vector Machines}

\author{
Alistair Shilton, 
Sunil Gupta, 
Santu Rana, 
Svetha Venkatesh \\}

\affiliations{
 Applied Artificial Intelligence Institute (A2I2) \\
 Deakin University, Geelong, Australia \\
 {\tt alistair.shilton@deakin.edu.au}
}

\begin{document}
\maketitle

\begin{abstract}
In this paper we explore a connection between deep networks and learning in reproducing kernel {\krein} space.  Our approach is based on the concept of push-forward - that is, taking a fixed non-linear transform on a linear projection and converting it to a linear projection on the output of a fixed non-linear transform, {\em pushing} the weights {\em forward} through the non-linearity.  Applying this repeatedly from the input to the output of a deep network, the weights can be progressively ``pushed'' to the output layer, resulting in a flat network that has the form of a fixed non-linear map (whose form is determined by the structure of the deep network) followed by a linear projection determined by the weight matrices - that is, we take a deep network and convert it to an equivalent (indefinite) kernel machine.  We then investigate the implications of this transformation for capacity control and uniform convergence, and provide a Rademacher complexity bound on the deep network in terms of Rademacher complexity in reproducing kernel {\krein} space.  Finally, we analyse the sparsity properties of the flat representation, showing that the flat weights are (effectively) $L_p$-``norm'' regularised with $p \in (0,1)$ (bridge regression).
\end{abstract}

\section{Introduction}

In machine learning, a clear distinction is often drawn between kernel methods 
such as support vector machines, which were overwhelmingly popular in the 
early-mid 2000s, and deep networks that have come to dominate the field since.  
Kernel methods are often characterised as elegant but limited - founded on 
beautiful mathematical theory (reproducing kernel Hilbert space etc), and 
intuitive (max-margin in feature space, geometric interpretation of support 
vectors etc), but inflexible and incapable of scaling to the needs of big-data 
- while deep networks are characterised as utilitarian but superior in terms of 
performance, scalability, and flexibility.  So deep networks now dominate in 
many areas, while kernel methods survive in niche applications.

An argument often made to explain the superior expressive power and performance 
of deep networks is the apparent complexity (and hence capacity) of such 
networks.  Kernel methods learn a linear relation in a feature space, with all 
nonlinearity contained in the fixed map from input space to feature space; 
while deep networks are built from many layers of non-linearity interspersed 
with linear maps (weight matrices).  Thus it may appear 
that (a) there is little or no crossover between the two methods, and (b) that 
deep networks are naturally more flexible and expressive.

In this paper we show that the distinction is not clear-cut.  In particular, a 
large family of deep networks can be precisely represented as single-layer 
networks of the SVM type - single-layer networks consisting of a fixed 
non-linear layer (a feature map encoded by a {\krein} kernel) followed by a 
trainable linear projection.  The structure of the deep network (number and 
width of layers, activation functions) is precisely encoded by a {\krein} 
kernel.  We show that the set of possible trained networks 
is in fact {\em smaller} than the set of possible trained machines for the 
corresponding single-layer network, which will allow us to analyse the capacity 
and generalisation of deep networks.

With regard to capacity analysis and uniform convergence bounds, in recent 
years a significant body of literature has been generated with bounds based on 
various assumptions 
\cite{Ney1,Ney2,Ney3,Ney4,Har2,Bar8,Gol3,Aro3,Zey1,Dra2,Li8,Nag1,Nag2,Zho5}.  
In this paper we approach the problem 
indirectly, which both simplifies the derivation and generalises the results.  By 
constructing an equivalence between deep networks and kernel methods using 
indefinite support vector machines, we are able to analyse the capacity of a 
deep network by bounding it by the capacity of a corresponding 
indefinite SVM.  Assuming the deep network is regularised using 
Frobenius norm on the weight matrices (weight decay), we give 
an equivalent regularisation scheme for the ``flat'' deep network 
representation.  We then show that the resulting (effective) regularisation 
term imposed by the deep network weight regularisation places an upper bound on 
the corresponding (naive) regularisation term for an SVM-type approach.  This 
allows us to show that the set of reachable functions in the deep network with 
bounded (norm) weight matrices is a subset of the corresponding set of 
reachable functions in the SVM approach.  Thus we can bound for example the 
Rademacher complexity of deep networks in terms of the Rademacher complexity of 
a corresponding indefinite ({\krein}) SVM, allowing a set of results to be 
directly transferred from the SVM context to the deep network context.

We finish by considering sparsity in the flat representation.  Sparsity in 
neural networks is reduces the complexity, but can also improve accuracy and 
robustness \cite{Wei4,Guo1}.  In this paper we show that simply applying 
standard, $L_2$-norm (weight) regularisation leads to sparsity in the flat 
representation by effectively applying bridge regression \cite{Fra2} 
($L_p$-norm regularisation for $p \in (0,1)$) to the flat weights.  This is 
particularly interesting when we consider recent results \cite{Ber2,Has4}, 
where it was shown that bridge regularisation can perform significantly better 
than alternatives e.g. $L_1$- or $L_2$-norm regularisation.

\subsection{Notation} \label{sec:sipsiip}

We use $\infset{N} = \{ 0,1,\ldots \}$, $\infset{N}_+ = \{ 1,2,\ldots \}$, 
$\bar{\infset{N}} = \infset{N} \cup \{ \infty \}$, $\infset{N}_n = \{ 0,1, 
\ldots,n-1 \}$, $\infset{R}_+ = \{ x \in \infset{R} | x > 0 \}$.  
Hilbert spaces are denoted $\rkhs{}$ and {\krein} spaces $\rkks{}$.  
%
%
%
%
For (countable) vectors ${\bf a}, {\bf b}$, $a_i$ denotes the $i^{\rm th}$ 
element of ${\bf a}$, ${\bf a} \odot {\bf b}$ is the elementwise product, 
${\bf a}^{\odot c}$ the elementwise power, $| {\bf a} |$ the elementwise 
absolute, $\sgn ({\bf a})$ the elementwise sign, and ${\rm sum} ({\bf a}) = 
\sum_i a_i$.  
%
%
We define $(a)_+ = \max \{ a,0 \}$, and $({\bf a})_+$ elementwise. 
%
%
We use a number of variations of inner product, denoted as follows 
\cite{Hor1,Dra1,Sal1,Sal2,Cra4}:
\[
 {\small
 \begin{array}{ccc}
                                       & \mbox{Definite} & \mbox{Indefinite} \\
  \mbox{Inner-product:}                & \lip   \cdot,\cdot  \rip{}   : \infset{V} \times \infset{V} \to \infset{R} & \liip  \cdot,\cdot  \riip{}  : \infset{V} \times \infset{V} \to \infset{R} \\
  \mbox{$m$-inner-product:}            & \lmip  \cdot,\ldots \rmip{}  : \infset{V}^m                 \to \infset{R} & \lmiip \cdot,\ldots \rmiip{} : \infset{V}^m                 \to \infset{R} \\
 \end{array}
 }
\]
all of which are symmetric and multilinear, and $\liip a, a' \riip{} = 0$ or 
$\lmiip a, a', \ldots \rmiip{} = 0$ $\forall a',\ldots \Rightarrow a = 0$.  
The ($m$-) inner product 
is norm-inducing ($\| x \|^2 = \lip x,x \rip{}$ and $\| x \|^m = \lmip x,x, 
\ldots \rmip{}$), and must satisfy the Cauchy-Schwarz inequality $|\lmip 
a,a',\ldots \rmip{}|^m \leq |\lmip a,a,\ldots \rmip{} \lmip a',a', \ldots 
\rmip{} \ldots|$.  
We also define weighted indefinite and definite $m$-inner products on 
$\infset{R}^n$ (see \cite{Dra1} regarding $\lmip \ldots \rmip{m,{\bf g}}$):
\[
 {\small
 \begin{array}{rll}
  \!\!\!\lmiip {\bf x}, {\bf x}', \ldots, {\bf x}'''' \rmiip{m,{\bf g}} &\!\!\!= \sum_i g_i x_i x'_i \ldots x''''_i \mbox{ (where }{\bf g} \in \infset{R}^n\mbox{)}\\
  \!\!\!\lmip  {\bf x}, {\bf x}', \ldots, {\bf x}'''' \rmip{m,{\bf g}}  &\!\!\!= \sum_i g_i x_i x'_i \ldots x''''_i \mbox{ (}{\bf g} \in \infset{R}^n, {\bf g} \geq {\bf 0}\mbox{)} \\
 \end{array}
 }
\]
Likewise $\liip {\bf x}, {\bf x}' \riip{\bf g} = \lmiip {\bf x}, {\bf x}' 
\rmiip{2,{\bf g}}$, $\lip {\bf x}, {\bf x}' \rip{\bf g} = \lmip {\bf x}, {\bf 
x}' \rmip{2,{\bf g}}$.\footnote{Technically $\lip {\bf x}, {\bf x}' \rip{{\bf 
g}}$ defined here is a positive semidefinite Hermitian form (inducing a 
seminorm rather than a norm) and not an inner product unless ${\bf g} > {\bf 
0}$.  However this makes no substantive difference to our results, so we use 
the less verbose definition.}

\section{Related Work} \label{sec:relwork}

The study of the connection between kernel methods and deep networks has a 
long history.  In \cite{Nea1} it was shown that, as the width of a single-layer 
neural network goes to infinity, and assuming iid random weights, the network 
converges to a draw from a Gaussian process.  This was extended to 
multi-layered nets \cite{Lee8,Mat5} by assuming random weights up to (but not 
including) the output layer.  Indeed, deriving approximate kernels through 
random weights is a popular means of linking deep networks and kernel methods 
\cite{Rah2,Bac3,Bac4,Dan1,Dan2}.

More recently, neural tangent kernels \cite{Jac2,Aro4} have been investigated.  
If $f (\cdot;\theta) : \infset{R}^D \to \infset{R}$ is a neural network 
parameterised by $\theta$ (weight matrices), the neural tangent kernel is the 
kernel associated with the feature map $x \to \bg{\nabla}_{\theta} 
f(x;\theta)$ via $K_{\NTK} (x,x') = \lip \bg{\nabla}_{\theta} 
f(x;\theta), \bg{\nabla}_{\theta} f(x';\theta) \rip{}$.   Neural tangent 
kernels allow us to analyse the generalization features of deep networks, 
particularly in the infinite width case where $K_{\NTK}$ converges to an 
explicit limit that does not change during training.  However neural tangent 
kernels do not provide a 1-1 equivalence in general, which is our goal here.  
Arc-cosine kernels \cite{Cho8} work on a similar premise.  For activation 
functions of the form $\sigma (\xi) = (\xi)_+^n$, $n = 0,1,2,\ldots$, letting 
the width of the network go to infinity, arc-cosine kernels capture the feature 
map of the network (depth is achieved by composition of kernels), effectively 
flattening it.  However once again this approach is restricted to networks of 
infinite width, whereas our approach works for arbitrary networks.

\section{Preliminaries I: Deep Networks} \label{sec:deepprelim}

\begin{figure}
 \centering 
 \includegraphics[width=0.34\textwidth]{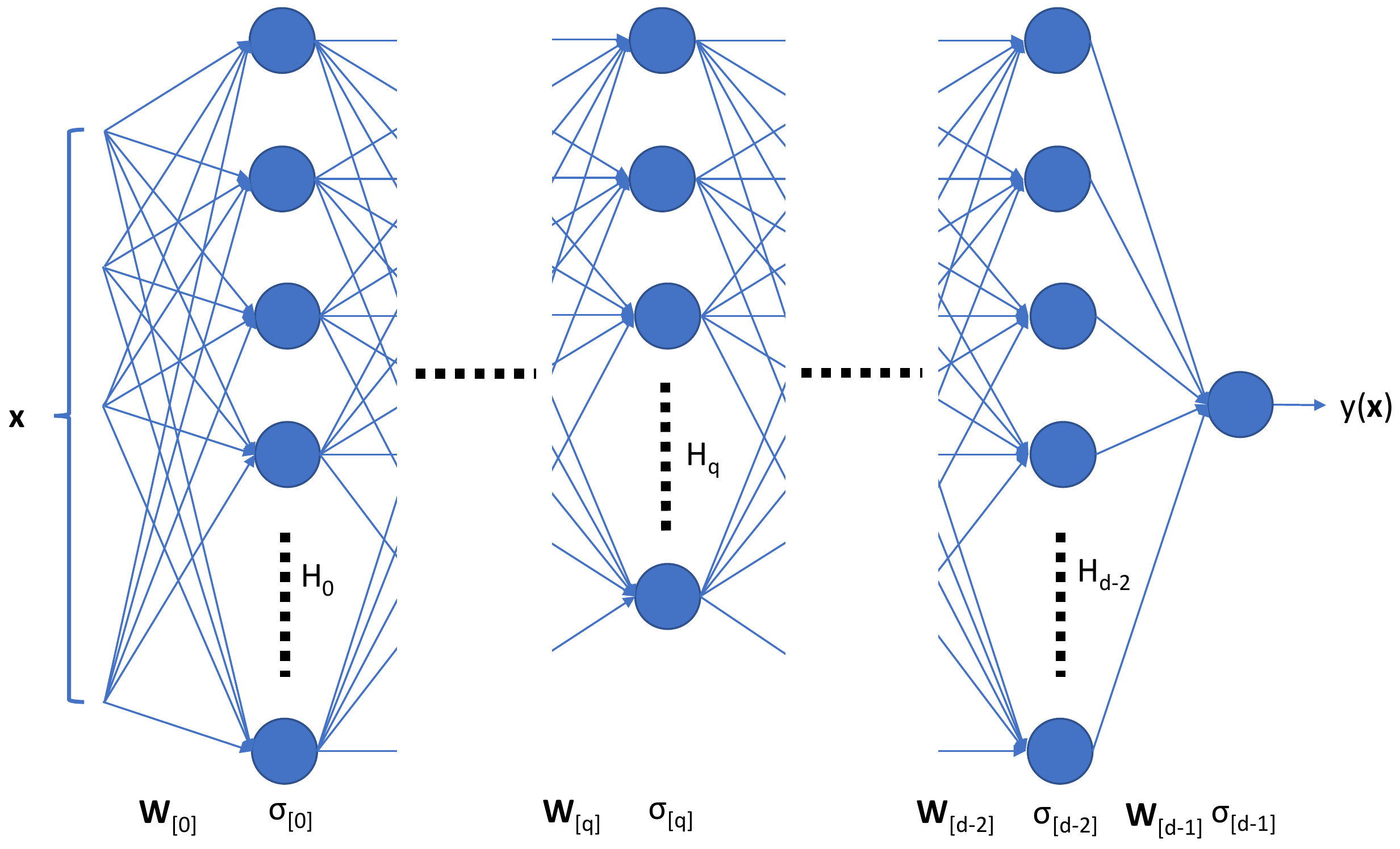} $\!\!\!$
 \includegraphics[width=0.1\textheight]{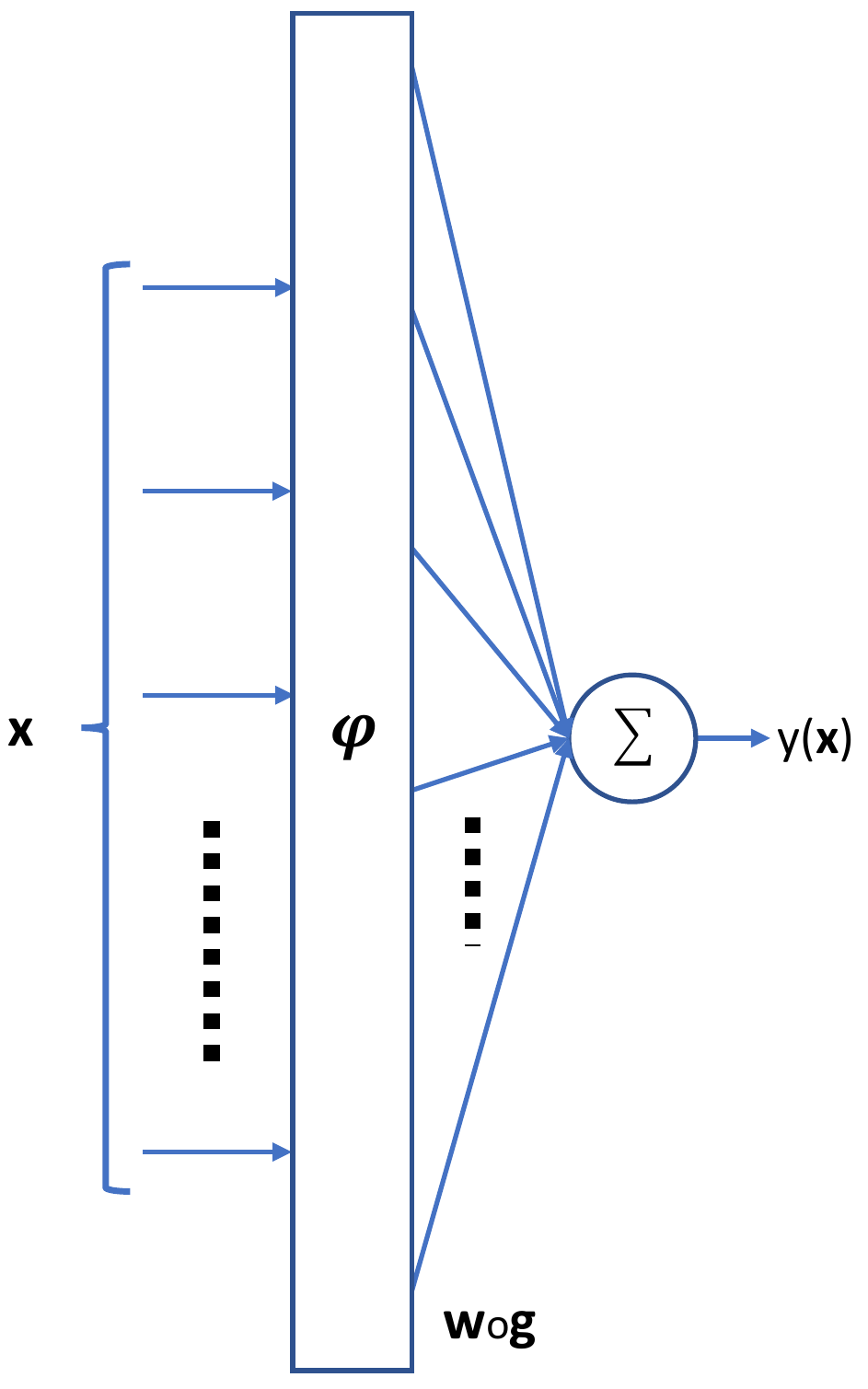} \\
 \caption{Machine learning architectures.  Left shows the physical deep network 
          architecture, and right shows the {\em representation} of the same 
          deep network in ``flat'' (feature space) form consisting of feature 
          map ${\bg{\featmapmkern}}_{\NN} : \infset{R}^D \to \frkks{}$ and a 
          linear projection onto $\infset{R}$, where the weight vector ${\bf w} 
          \in \frkks{}$ is the ``push-forward'' of the weight matrices in the 
          deep network.}
 \label{fig:deepstandard}
\end{figure}

For the purposes of this paper, a (fully connected and layered) $d$-layer 
feedforward neural network computes a function $f : \infset{R}^D \to 
\infset{R}$ is as shown in figure \ref{fig:deepstandard}, where the layers are 
indexed $0,1,\ldots,d-1$, and layer $q$ has width $H_q$.  We assume that all 
nodes in layer $q$ share the same activation function $\sigma_{q} : \infset{R} 
\to \infset{R}$. Given input ${\bf x} \in \infset{R}$, the output is:
\begin{equation}
 {\small\!\!\!\!
 \begin{array}{l}
  f \!\left( {\bf x} \right) = \sigma_{d-1} \!\left( 
  {\bf W}_{[d-1]} \sigma_{d-2} \!\left(
  {\bf W}_{[d-2]}
  \ldots \sigma_{0} \!\left(
  {\bf W}_{[0]} {\bf x}
  \right)
  \right)
  \right)_{{\;}_{\;}}
 \end{array}
 \!\!\!\!}
 \label{eq:deepnetform}
\end{equation}
where $\sigma_{q}$ operates elementwise and ${\bf W}_{[q]} \in \infset{R}^{H_q 
\times H_{q-1}}$ is the weight matrix for each layer $q$; and we let $H_{-1} = 
D$ and $H_{d-1} = 1$ (scalar output).  Weight matrices are chosen to solve the 
regularised risk minimisation problem:
\begin{equation}
 \begin{array}{l}
  \!\!\!\!\!\!\mathop{\min}\limits_{{\bf W}_{[q]} \in \infset{R}^{H_q \times H_{q-1}}} \!\!\frac{1}{N} \!\sum_i \!\loss \!\left( y_i, f \left( {\bf x}_i \right) \right) \!+\! \lambda \nnregscale \!\sum_q \!\left\| {\bf W}_{[q]} \right\|_F^2\!\!\!\!\!\!
 \end{array}
 \label{eq:origdeepreg}
\end{equation}
where the first term is the empirical risk ($\loss$ is the loss function, which 
will vary depending on the purpose of the network (classification, regression 
etc)) and the second term is a regularisation penalty.  As noted in \cite{Bis4}, if $\loss$ is quadratic and the training data 
       is noisy then we do not require the explicit 
       regularisation term as there is an {\em implicit} Tikhonov 
       regularisation present.  Note that:
\begin{enumerate}
 \item We use minimise in the local sense, as local minima suffice.
 \item Any topologically equivalent norm can be substituted for the Frobenius 
       norm $\| \cdot \|_F$, with the effect of introducing additional 
       constants into certain bounds but otherwise with no substantive change.
 \item We assume the activation functions $\sigma_{q}$ are increasing, 
       entire,\footnote{We discuss how the entire function requirement may be 
       relaxed in the supplementary.} positive at $0$ and Lipschitz on 
       $\infset{R}_+$ with constant $L_q$.  Hence $\sigma_q$ has an everywhere 
       convergent Taylor expansion $\sigma_q (\xi) = \sum_i a_{[q]i} \xi^i$, 
       where $a_{[q]0}, a_{[q]1} > 0$ $\forall q$.
 \item For each layer $q \in \infset{N}_d$ we define an associated (convex) 
       activation function $\bar{\sigma}_{[q]} (\xi) = \sum_i |a_{[q]i}| \xi^i$, where 
       $\bar{\sigma}_q = \sigma_q$ if $a_{[q]i} \geq 0$ $\forall i$.  Note that 
       $\bar{\sigma}_q$ is Lipschitz on any finite interval $[0,m]$ with 
       constant $\bar{L}_q$, where $\bar{L}_q \ne L_q$ in general.
 \item When discussing the network width and Lipschitz constants of 
       $\sigma_q$ and $\bar{\sigma}_q$ we find it most convenient to use the 
       geometric mean, which we write as $H = \GM(H_0,H_1,\ldots)$, $L = 
       \GM(L_0,L_1,\ldots)$ and $\bar{L} = \GM(\bar{L}_0,\bar{L}_1,\ldots)$.
\end{enumerate}

We will show that the deep network (\ref{eq:deepnetform}) can be rewritten in 
feature-space form as per figure \ref{fig:deepstandard}:
\[
 \begin{array}{l}
  f \left( {\bf x} \right) 
  = \liip {\bf v}, {\bg{\featmapmkern}} \left( {\bf x} \right) \riip{{\bf g}}
  = \sum_i g_i v_i {\featmapmkern}_i \left( {\bf x} \right)
 \end{array}
\]
which will allow us to build a connection between deep networks and support 
vector machines.  Before proceeding, however, we first present some background 
on the theory of indefinite ({\krein}) support vector machines.

\section{Preliminaries II: Indefinite SVMs} \label{sec:prelimsvm}

Indefinite (or {\krein}) support vector machines (SVMs) 
\cite{Lin11,Lus1,Haa1,Yin2,Sch27} are an extension of support vector machines 
\cite{Cor1,Bur1,Smo4,Cri4,Ste3} that relax the usual positive definiteness 
requirement on the kernel, based on the observation that indefinite kernels, 
naively applied, outperform positive definite kernels in some cases.  They may 
be interpreted \cite{Ong3,Ogl1,Ogl2} as a form of regularised learning in 
reproducing kernel {\krein} space \RKKS \cite{Bog1,Azi1}.  Typically, 
indefinite SVMs are introduced without reference to the primal formulation 
often found in standard SVM theory (for example \cite{Cor1}), but as we require 
the primal formulation here we now give a brief introduction from this 
perspective using the {\krein}-kernel trick.  Our approach is loosely based on 
\cite{Cor1}, extended to the indefinite case.  For a more conventional 
introduction see the supplementary material.

We consider a function of the simple, linear form:
\begin{equation}
 \begin{array}{l}
  f \left( {\bf x} \right) 
  = \liip {\bf v}, {\bg{\featmapmkern}} \left( {\bf x} \right) \riip{{\bf g}}
  = \sum_i g_i v_i {\featmapmkern}_i \left( {\bf x} \right)
 \end{array}
 \label{eq:wfinalformb}
\end{equation}
where, denoting the feature space by $\frkks{}$, the feature map 
${\bg{\featmapmkern}} : \infset{R}^D \to \frkks{}$ and the metric ${\bf 
g} \in \frkks{}$ are defined a-priori (implicitly, as we will see, by a 
{\krein} kernel).  We note that this is the same as the primal form of the 
trained machine in SVM theory, excepting that it involves a weighted indefinite 
inner product rather than the usual inner product; that is, it is an {\em 
indefinite} SVM primal.  In SVM learning, as in deep networks, the goal is to 
mimic the input/output relation embodied by the training set $\{ ({\bf x}_i, 
y_i) \in \infset{R}^D \times \infset{R} | i \in \infset{N}_N \}$.  In an 
indefinite SVM this is done by minimising the stabilised risk minimisation 
problem (\cite[equation (1)]{Ogl2}, \cite{Loo1}), noting that the 
regularisation penalty $\liip {\bf v}, {\bf v} \riip{\bf g}$ here is not a 
norm (it may be positive, negative or zero):
\begin{equation}
 \begin{array}{l}
  \mathop{\min}\limits_{{\bf v} \in \frkks{}} \frac{1}{N} {\sum}_i \loss \left( y_i, \liip {\bf v}, {\bg{\featmapmkern}} \left( {\bf x}_i \right) \riip{{\bf g}} \right) + \lambda \liip {\bf v}, {\bf v} \riip{{\bf g}} \\
 \end{array}
 \label{eq:trainSVM}
\end{equation}
where once again we use min in the loose sense, as local minima are allowed 
(see \cite{Loo1} for discussion, as well as an alternative notation).  
Representor theory follows as usual (proof in supplementary):
\begin{th_indefrep}[Representor Theory]
 Any solution ${\bf v}^\star$ to (\ref{eq:trainSVM}) can be represented as 
 ${\bf v}^\star = \sum_i \alpha_i {\bg{\featmapmkern}} ({\bf x}_i)$, 
 where ${\bg{\alpha}} \in \infset{R}^N$.  Defining $K ({\bf x}, 
 {\bf x}') = \liip {\bg{\featmapmkern}} ({\bf x}), 
 {\bg{\featmapmkern}} ({\bf x}') \riip{{\bf g}}$, the optimal 
 $f^\star : \infset{R}^D \to \infset{R}$ is $f^\star ({\bf 
 x}) = \sum_i \alpha_i^{\star} K ({\bf x}, {\bf x}_i)$.
 \label{th:indefrep}
\end{th_indefrep}
Note that, for $K$ as per theorem \ref{th:indefrep}, the stabilised risk 
minimisation problem (\ref{eq:trainSVM}) can be rewritten in terms of 
${\bg{\alpha}}$ as:
\begin{equation}
 {\!\!\!\!\!\!\!\!\small{
 \begin{array}{rl}
  \mathop{\min}\limits_{{\bg{\alpha}} \in \mathbb{R}^N} \!\frac{1}{N} \!{\sum}_i \loss \!\left( \!y_i, {\sum}_j \alpha_j K\! \left( {\bf x}_i, {\bf x}_j \right) \!\right) \!+\! \lambda {\sum}_{i,j} \alpha_i \alpha_j K\! \left( {\bf x}_i, {\bf x}_j \right) \\
 \end{array}
 }\!\!\!\!}
 \label{eq:trainSVMalpha}
\end{equation}
In this formulation $K$ is a {\em {\krein} kernel}.  That is, $K : \infset{R}^D 
\times \infset{R}^D \to \infset{R}$ that can be written as a difference $K = 
K_+ - K_-$ between positive definite kernels $K_\pm$ \cite[Proposition 
7]{Ong3}.  Note that $K$ in theorem \ref{th:indefrep} can be split in this 
manner (writing $(a)_+ = {\rm max} \{ 0,a \}$ and $({\bf a})_+$ elementwise):
\begin{equation}
 \begin{array}{l}
  K_{\pm} \left( {\bf x}, {\bf x}' \right) = \lip {\bg{\featmapmkern}} \left( {\bf x} \right), {\bg{\featmapmkern}} \left( {\bf x}' \right) \rip{(\pm {\bf g})_+}
 \end{array}
 \label{eq:Kpm_basics}
\end{equation}
where $K_{\pm}$ are trivially positive definite.  Conversely, given a {\krein} 
kernel $K$, by definition there exist positive definite $K_\pm$ (non-uniquely) 
such that $K = K_+ - K_-$.  Hence there exists implicit, finite or countably 
infinite dimensional expansions: 
\[
 \begin{array}{l}
  K_{\pm} \left( {\bf x}, {\bf x}' \right) = \lip {\bg{\featmapmkern}}_\pm \left( {\bf x} \right), {\bg{\featmapmkern}}_\pm \left( {\bf x}' \right) \rip{{\bf 1}}
 \end{array}
\] 
(using Mercer's theorem), so $K ({\bf x}, {\bf x}') = \liip {\bg{\featmapmkern}} 
({\bf x}), {\bg{\featmapmkern}} ({\bf x}') \riip{{\bf g}}$ where 
${\bg{\featmapmkern}} ({\bf x}) = [ {\bg{\featmapmkern}}_+ ({\bf x}), 
{\bg{\featmapmkern}}_- ({\bf x}) ]$ and ${\bf g} = [ +{\bf 1}, -{\bf 1} ]$.  
Consequently, as for standard SVMs, we don't need to know the feature map and 
metric; rather, we just need a {\krein} kernel to implicitly define a feature 
map and metric.  We call this the {\krein} kernel trick by analogy with the 
more familiar (non-{\krein}) kernel trick commonly used in kernel methods.

When analysing the {\em capacity} of indefinite SVMs we also need to define the 
associated kernel.  Given a {\krein} kernel $K = K_+ - K_-$, the associated 
kernel is $\bar{K} = K_+ + K_-$, which we note is positive definite.  In terms 
of the metric ${\bf g}$ if $K ({\bf x},{\bf x}') = \liip \bg{\featmapmkern} 
({\bf x}), \bg{\featmapmkern} ({\bf x}') \riip{\bf g}$ then, using 
(\ref{eq:Kpm_basics}):
\[
 \begin{array}{l}
  \bar{K} \left( {\bf x}, {\bf x}' \right) = \lip \bg{\featmapmkern} \left( {\bf x} \right), \bg{\featmapmkern} \left( {\bf x}' \right) \rip{\left| {\bf g} \right|}
 \end{array}
\]

As discussed in the supplementary, the {\krein} kernel $K$ defines a 
reproducing kernel {\krein} space ({\RKKS}) $\rkks{K}$, and the associated 
kernel $\bar{K}$ defines a reproducing kernel Hilbert space ({\RKHS}) 
$\rkhs{\bar{K}}$ \cite{Ong3}:
\[
 \begin{array}{rl}
  \rkks{K}       &\!\!\!\!= \left\{ \left. f \left( \cdot \right) = \liip {\bf v}, {\bg{\featmapmkern}} \left( \cdot \right) \riip{{\bf g}} \right| {\bf v} \in \frkks{} \right\} \\
  \rkhs{\bar{K}} &\!\!\!\!= \left\{ \left. f \left( \cdot \right) = \lip {\bf v}, {\bg{\featmapmkern}} \left( \cdot \right) \rip{|{\bf g}|} \right| {\bf v} \in \frkhs{} \right\} \\
 \end{array}
\]
where $\rkks{K}$ is is equipped with an indefinite inner product $\liip 
\liip {\bf v}, {\bg{\featmapmkern}} (\cdot) \riip{{\bf g}}, \liip {\bf 
v}', {\bg{\featmapmkern}} (\cdot) \riip{{\bf g}} \riip{\subrkks{K}} = 
\liip {\bf v}, {\bf v}' \riip{{\bf g}}$ and $\rkhs{\bar{K}}$ is equipped with 
an inner product $\lip \lip {\bf v}, {\bg{\featmapmkern}} (\cdot) \rip{|{\bf 
g}|}, \lip {\bf v}', {\bg{\featmapmkern}} (\cdot) \rip{|{\bf g}|} 
\rip{\subrkhs{\bar{K}}} = \lip {\bf v}, {\bf v}' \rip{|{\bf g}|}$ ($\rkks{K}$ 
and $\rkhs{\bar{K}}$ coincide if $K$ is positive definite).  Hence $f \in 
\rkks{K}$ and (\ref{eq:trainSVM}) can be rewritten:
\begin{equation}
 \begin{array}{rl}
  \mathop{\min}\limits_{f \in \subrkks{K}} \frac{1}{N} \sum_i \loss \left( y_i, f \left( {\bf x}_i \right) \right) + \lambda \liip f,f \riip{\subrkks{K}} 
 \end{array}
 \label{eq:trainSVMfunc}
\end{equation}

\section{Flat Representation for Deep Network} \label{sec:assump} \label{sec:flattenit}

We aim to show that the deep network (\ref{eq:deepnetform}) can be rewritten in 
the simpler, flattened representation:
\begin{equation}
 \begin{array}{rll}
  f \left( {\bf x} \right) 
  &\!\!\!\!= \liip {\bf v}_{\NN}, {\bg{\featmapmkern}}_{\NN} \left( {\bf x} \right) \riip{{\bf g}_{\NN}} 
  &\!\!\!\!= \sum_i g_{\NN i} v_{\NN i} {\featmapmkern}_{\NN i} \left( {\bf x} \right)
 \end{array}
 \label{eq:wfinalform}
\end{equation}
where $\liip \cdot,\cdot \riip{{\bf g}_{\NN}}$ is an indefinite-inner-product, 
${\bg{\featmapmkern}}_{\NN} : \infset{R}^D \to \frkks{}$ is a feature 
map and ${\bf g}_{\NN}$ is a metric; ${\bg{\featmapmkern}}_{\NN}$ and 
${\bf g}_{\NN}$ are defined by the network structure; and ${\bf v}_{\NN} \in 
\frkks{\NN}$ is a weight vector that solves the regularised risk minimisation 
problem:\footnote{We use the subscript $\mbox{\tt NN}$ on ${\bf v}_{\NN}$ as a 
visual reminder that ${\bf v}_{\NN}$ corresponds, possibly non-uniquely, to 
some set of weight matrices ${\bf W}_{[0]}$, ${\bf W}_{[1]}$, $\ldots$ in a 
deep network satisfying our assumptions such that (\ref{eq:deepnetform}) and 
(\ref{eq:wfinalform}) are functionally equivalent}
\begin{equation}
 {\!\!\!\!\!\!
 \begin{array}{l}
  \mathop{\min}\limits_{{\bf v}_{\NN} \in \frkks{\NN} \subseteq \frkks{}} \frac{1}{N} {\sum}_i \loss \left( y_i, f \left( {\bf x}_i \right) \right) + \lambda \reg_{\NN} \left( {\bf v}_{\NN} \right) \\
 \end{array}
 \!\!\!\!\!\!}
 \label{eq:deepfeatminprob_prelim}
\end{equation}
for appropriate $\frkks{\NN} \subset \frkks{}$ and $\reg_{\NN} : \frkks{} \to 
\infset{R}$; such that the trained networks (\ref{eq:deepnetform}) and 
(\ref{eq:wfinalform}) are functionally equivalent.  This representation is 
analogous to the trained indefinite SVM primal (\ref{eq:wfinalformb}), which 
will allow us to analyse deep networks from the same perspective as indefinite 
SVMs.

As an intermediate step we begin showing that the deep network 
(\ref{eq:deepnetform}) can be rewritten in a semi-flat form:
\begin{equation}
 \begin{array}{rl}
  f \left( {\bf x} \right) 
  &\!\!\!\!= \lmiip {\bf v}_{[0]}, {\bf v}_{[1]}, \ldots, {\bf v}_{[d-1]}, {\bg{\featmapmkern}}_{\NN} \left( {\bf x} \right) \rmiip{d+1,{\bf g}_{\NN}} \\
  &\!\!\!\!= \sum_i g_{\NN i} v_{[0]i} v_{[1]i} \ldots v_{[d-1]i} {\featmapmkern}_{\NN i} \left( {\bf x} \right)
 \end{array}
 \label{eq:semideepflat}
\end{equation}
where $\lmiip \cdot,\cdot, \ldots \rmiip{d+1,{\bf g}_{\NN}}$ is an indefinite 
$(d+1)$-inner-product and, for all $q \in \infset{N}_d$, the weight vectors 
${\bf v}_{[q]} \in \frkks{q} \subset \frkks{}$ solve the regularised risk 
minimisation problem:
\begin{equation}
 {\!\!\!\!\!\!
 \begin{array}{l}
  \mathop{\min}\limits_{{\bf v}_{[q]} \in \frkks{q}} \frac{1}{N} {\sum}_i \loss \left( y_i, f \left( {\bf x}_i \right) \right) + \lambda \sum_q \reg_{q} \left( {\bf v}_{[q]} \right) \\
 \end{array}
 \!\!\!\!\!\!}
 \label{eq:deepfeatminprob_prelim_semi}
\end{equation}

Central to our approach is the push-forward operation, converting a nonlinear 
function of a multilinear product of vectors to a multilinear product of the 
non-linear images of the original vectors - that is:
\begin{th_pushforward}
 Let $\sigma$ be an entire function with Taylor expansion $\sigma(\xi) = \sum_i 
 a_i \xi^i$, and let $\lmiip \cdot, \cdot, \ldots \rmiip{m, 
 {\bg{\scriptstyle\mu}}}$ be an $m$-indefinite-inner-product defined by 
 metric ${\bg{\mu}} \in \infset{R}^n$ (section \ref{sec:sipsiip}).  
 Then:
 \begin{equation}
  \begin{array}{rl}
   \!\!\!\!\!\!\sigma \!\left( \lmiip {\bf x}, \ldots, {\bf x}'''' \rmiip{m,{\bg{\scriptstyle\mu}}} \right) 
   \!=\! \lmiip {\bg{\featmap}} \!\left( {\bf x} \right), \ldots, {\bg{\featmap}} \!\left( {\bf x}'''' \right) \rmiip{m,{\bg{\scriptstyle\gamma}} \odot {\bg{\scriptstyle\featmap}} ({\bg{\scriptstyle\mu}})}\!\!\!\!\!\! \\
  \end{array}
  \label{eq:define_feature}
 \end{equation}
 where $\bg{\featmap} : \infset{R}^n \to \frkks{}$ is a feature map and 
 $\bg{\gamma} \in \frkks{}$, both independent of $m$ and ${\bg{\mu}}$.  
 Using multi-index notation, ${\bg{\featmap}} ({\bf x}) = [ 
 {{\featmap}}_{\bf i} ({\bf x})]_{{\bf i} \in \infset{N}^n}$ and 
 ${\bg{\gamma}} = [ {{\gamma}}_{\bf i}]_{{\bf i} \in \infset{N}^n}$, 
 where:
 \begin{equation}
  \begin{array}{l}
   {\featmap}_{\bf i} \left( {\bf x} \right) = \prod_j x_j^{i_j}, \;\;\;\; {\gamma}_{\bf i} = \left( \frac{{\rm sum} \left( {\bf i} \right)}{\prod_j i_j!} \right) a_{{\rm sum} \left( {\bf i} \right)}
  \end{array}
  \label{eq:featmapform}
 \end{equation}
 \label{th:pushforward}
\end{th_pushforward}
\begin{proof}
Equations (\ref{eq:define_feature}) and (\ref{eq:featmapform}) follows from the 
multinomial expansion of $\sigma$ and subsequent collection of terms.  See 
supplementary for details.
\end{proof}

\begin{figure}
 \centering
 \includegraphics[width=0.44\textwidth]{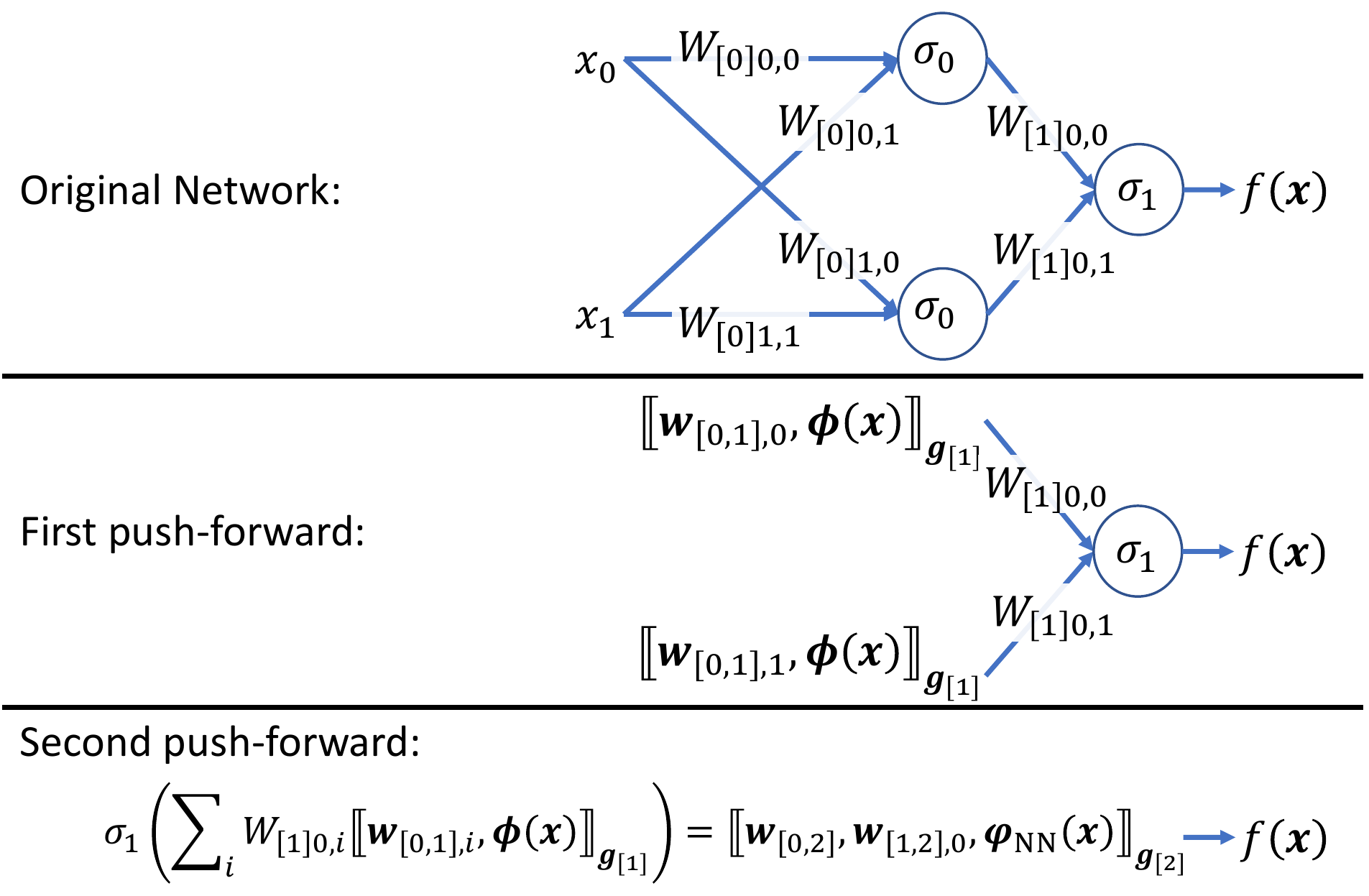}
 \caption{Push-forward on a simple $2$-layer neural network.  Starting with the 
          original network (top), we apply push-forward to layer $0$, so by 
          (\ref{eq:pushforwardsigma}) the output of neuron $i$ in layer $0$ is 
          $o_{[0],i} ({\bf x}) = \sigma_0 (\lmiip {\bf W}_{[0]i,:},{\bf x} 
          \rmiip{2,{\bf 1}}) = \lmiip {\bf w}_{[0,1] i}, \bg{\featmap} 
          ({\bf x}) \rmiip{2,{\bf g}_{[1]}}$.  Repeating for layer $1$ 
          (with some housekeeping as per the supplementary) we find $f ({\bf 
          x}) = \lmiip {\bf w}_{[0,2]}, {\bf w}_{[1,2] 0}, 
          \bg{\featmapmkern}_{\NN} ({\bf x}) \rmiip{3,{\bf g}_{[2]}}$.} 
 \label{fig:pushforward}
\end{figure}

We call the image ${\bg{\featmap}} ({\bf x})$ of ${\bf x}$ the {\em 
push-forward} of ${\bf x}$, as it heuristically represents the result of 
pushing ${\bf x}$ forwards through $\sigma$.  Recalling that we are assuming 
entire activation functions, by (\ref{eq:define_feature}) and 
(\ref{eq:featmapform}), using multi-index notation, the activation functions 
satisfy the following:
\begin{equation}
 {\!\!\!\!\!\!\small
 \begin{array}{l}
  \sigma_{q} \!\left( \lmiip {\bf x}, \ldots, {\bf x}'''' \rmiip{m,{\bg{\scriptstyle\mu}}\!} \right) \!=\! 
  \lmiip {\bg{\featmap}} \left( {\bf x} \right)\!, \ldots, \!{\bg{\featmap}} \!\left( {\bf x}'''' \right) \rmiip{m,{\bg{\scriptstyle\gamma}}_{[q]} \odot {\bg{\scriptstyle\featmap}} ({\bg{\scriptstyle\mu}})} \\
  \gamma_{[q] \bf i} = \left( \frac{{\rm sum} \left( {\bf i} \right)}{\prod_j i_j!} \right) a_{[q]{\rm sum} \left( {\bf i} \right)}
 \end{array}
 \!\!\!\!\!\!}
 \label{eq:pushforwardsigma}
\end{equation}
where the indices $q \in \infset{N}_d$ again denotes layer $q$ (we do not place 
a subscript on ${\bg{\featmap}}$ as, by Lemma \ref{th:pushforward}, this 
depends on the input dimension, not the activation function $\sigma_{q}$).  The 
next step is to apply push-forward repeatedly, starting with ${\bf x}$ and 
${\bf W}_{[0]}$ at the input to layer $0$, applying (\ref{eq:pushforwardsigma}) 
to obtain the push-foward representation at the input to layer $1$, and 
repeating until the output layer is reached.  
This is shown in figure \ref{fig:pushforward} for a simple $2$-layer network 
with $d = D = H_0 = 2$.  
The result of this procedure is the semi-flat form (\ref{eq:semideepflat}), 
where the feature map is (theorem \ref{th:collapse} in supplementary):
\begin{equation}
 {\!\!\!\small{
  \begin{array}{l}
   {\bg{\featmapmkern}}_{\NN} \left( {\bf x} \right) = {\bf 1}_{H_{d-1}} \otimes {\bg{\featmap}} \left( {\bf 1}_{H_{d-2}} \otimes {\bg{\featmap}} \left( \ldots {\bf 1}_{H_0} \otimes {\bg{\featmap}} \left( {\bf x} \right) \right) \right) \\
  \end{array}
 }\!\!\!}
 \label{eq:featmapis}
\end{equation}
with weight vectors and metric:
\begin{equation}
 {\!\!\!\!\!\!\small{
  \begin{array}{rl}
   {\bf v}_{[q]} &\!\!\!\!= {\bg{\featmap}}_{[q+]} \left( {\bf W}_{[q]}^\tsp \right) \\
   {\bf g}_{\NN} &\!\!\!\!= {\bg{\gamma}}_{[d-1]} \!\odot\! {\bg{\featmap}} \left( {\bf 1}_{H_{d-2}} \!\otimes\! \left( \ldots {\bf 1}_{H_0} \!\otimes\! \left( {\bg{\gamma}}_{[0]} \!\odot\! {\bg{\featmap}} \left( {\bf 1}_D \right) \right) \right) \right) \\
  \end{array}
 }\!\!\!\!\!\!}
 \label{eq:wform}
\end{equation}
where, writing ${\bf W}_{i,:}$ for row $i$ of matrix ${\bf W}$:
\begin{equation}
 {\!\!\!\!\!\!\!\!\small{
 \begin{array}{l}
 \begin{array}{rl}
  \!\!\!{\bg{\featmap}}_{[q+]} \!\!\left( \!{\bf W}^\tsp \!\right) 
  &\!\!\!\!\!\!=\! {\bf 1}_{\!H_{d-1}} \!\!\!\otimes\! {\bg{\featmap}} \!\!\left(\!\! \ldots {\bf 1}_{\!H_{q+1}} \!\!\!\otimes\! {\bg{\featmap}} \!\!\left( \!\left[\!\!\! \begin{array}{c} {\bg{\featmap}} \left( {\bf W}_{\!0,:} \!\otimes \!{\bf e}_{[q]} \right) \\ {\bg{\featmap}} \left( {\bf W}_{\!1,:} \!\otimes \!{\bf e}_{[q]} \right) \\ \vdots \\ \end{array} \!\!\!\right] \!\right) \!\!\right) \\
 \end{array} \\
 \begin{array}{l}
  {\bf e}_{[q]} \!=\! {\bf 1}_{H_{q-1}} \otimes {\bg{\featmap}} \left( \ldots {\bf 1}_{H_0} \otimes {\bg{\featmap}} \left( {\bf 1}_D \right) \right)^{{\;}^{{\;}^{{\;}^{{\;}^{\;}}}}}\!\!\!\!\!\!\!\! \\
 \end{array}
 \end{array}
 }\!\!\!\!\!\!\!\!}
 \label{eq:featmapdetail}
\end{equation}

To rewrite the original regularised risk minimisation problem 
(\ref{eq:origdeepreg}) in terms of the weight vectors ${\bf v}_{[q]}$ in 
(\ref{eq:deepfeatminprob_prelim_semi}) we can simply note the correspondence 
(\ref{eq:wform}) between ${\bf v}_{[q]} \in \frkks{q}$ and ${\bf W}_{[q]}$ for 
all $q \in \infset{N}_d$ and define the regularisation penalty as:
\begin{equation}
 {\small{\!\!\!
  \begin{array}{l}
   \reg_{q} \left( {\bf v}_{[q]} \right) = \mathop{\rm sel}\limits_{\!\!\!\!\!{{\bf W}_{[q]} \in \infset{R}^{H_q \times H_{q-1}} :}  {{\bf v}_{[q]} = \bg{\featmap}_{[q+]} ({\bf W}_{[q]})}\!\!\!} \nnregscale \left\| {\bf W}_{[q]} \right\|_{[q]}^2 \\
  \end{array}
 \!\!\!}}
 \label{eq:reginsemiflat}
\end{equation}
where ${\rm sel}$ means ``select'' (this may not be unique), and:
\begin{equation}
 {\small{
  \begin{array}{l}
   \frkks{q} = \left\{\! \left. {\bg{\featmap}}_{[q+]} \left( {\bf W}_{[q]} \right) \right| {\bf W}_{[q]} \in \infset{R}^{H_q \times H_{q-1}} \right\}
  \end{array}
 }}
\end{equation}

Having derived a semi-flat representation it is straightforward to derive 
the (fully) flat form (\ref{eq:wfinalform}) by noting that $\lmiip {\bf 
v}_{[0]}, {\bf v}_{[1]}, \ldots, {\bf v}_{[d-1]}, 
\bg{\featmapmkern}_{\NN} ({\bf x}) \rmiip{m+1,{\bf g}} = \liip {\bf 
v}_{\NN}, \bg{\featmapmkern}_{\NN} ({\bf x}) \riip{\bf g}$, where ${\bf 
v}_{\NN} = \bigodot_q {\bf v}_{[q]} \in \frkks{\NN}$ and:
\begin{equation}
 {\small{
 \begin{array}{l}
  \frkks{\NN} = \left\{ \left. \bigodot_q {\bg{\featmap}}_{[q+]} \left( {\bf W}_{[q]} \right) \right| {\bf W}_{[q]} \in \infset{R}^{H_q \times H_{q-1}} \right\}
 \end{array}
 }}
 \label{eq:flatspacedef}
\end{equation}
Thus we see that (\ref{eq:semideepflat}) reduces to (\ref{eq:wfinalform}) 
where ${\bg{\featmapmkern}}_{\NN}$ is as per (\ref{eq:featmapis}).  To 
derive an appropriate regularisation penalty to ensure that solving 
(\ref{eq:deepfeatminprob_prelim}) makes (\ref{eq:deepnetform}) and 
(\ref{eq:wfinalform}) functionally identical, we can once again note the 
correspondence between ${\bf v}_{\NN}$ and $\{ {\bf W}_{[0]}, {\bf W}_{[1]}, 
\ldots \}$ and define the regularisation penalty (non-uniquely - see below) as:
\begin{equation}
 {\!\!\!\!\!\!\small{
  \begin{array}{l}
   \reg_{\NN} \left( {\bf v}_{\NN} \right) = \mathop{\rm sel}\limits_{{{\bf v}_{\NN} = \bigodot_q {\bf v}_{[q]}} : {\bf v}_{[q]} \in \frkks{q}} \nnregscale \sum_q \left\| {\bf W}_{[q]} \right\|_{[q]}^2 \\
  \end{array}
 }\!\!\!\!\!\!}
 \label{eq:reginflat}
\end{equation}

We note that neither the flat or semi-flat are intended for direct 
application.  Rather, they will (a) allow us to derive {\krein} kernels that 
will allow us to construct indefinite SVMs with the same feature map in flat 
form (but different regularisation) as the deep network, and subsequently (b) 
allow us to analyse the properties of the deep network in terms of complexity 
analysis from a novel angle.

\subsection{Regularisation Properties}

As presented in the (\ref{eq:reginsemiflat}-\ref{eq:reginflat}) in the previous 
section, the flat and semi-flat regularised risk penalties are uninformative.  
Ideally we would prefer to regularise in terms of ${\bf v}_{[q]}$ or ${\bf 
v}_{\NN}$ directly without reference to the corresponding weight matrices.  
While this does not appear to be precisely possible, the following theorem 
shows that we can bound the regularisation penalties in terms of either $\liip 
{\bf v}_{[q]}, {\bf v}_{[q]} \riip{\bf g}$ or $\lip {\bf v}_{[q]}, {\bf 
v}_{[q]} \rip{|{\bf g}|}$ (see supplementary for proof):
\begin{cor_collapse}
 Recalling that $\sigma_{q} (\xi) = \sum_i a_{[q]i} \xi^i$, for all $q \in 
 \infset{N}_d$, and $\bar{\sigma}_{q} (\xi) = \sum_i |a_{[q]i}| \xi^i$.  
 Defining 
 $\liip {\bf v}_{[q]}, {\bf v}_{[q]} \riip{{\bf g}_{\NN}} = p_q({\bf W}_{[q]})$ 
 and $\lip {\bf v}_{[q]}, {\bf v}_{[q]} \rip{|{\bf g}_{\NN}|} = \bar{p}_q ({\bf 
 W}_{[q]})$, 
 where:
 \begin{equation}
  {\small\!\!\!\!\!\!\!\!\!{
  \begin{array}{l}
   p_q \left( {\bf W}_{[q]} \right)
   \!=\! \sigma_{d-1} \Big( H_{d-2} \sigma_{d-2} \Big( H_{d-3} \sigma_{d-3} \Big( \ldots \\
   H_{q+1} \sigma_{q+1} \Big( \sum_{i_{q}} \sigma_{q} \Big( \left\| {\bf W}_{[q]i_{q},:} \right\|_2^2 \sigma_{q-1} \Big( \ldots 
   \!H_0 \sigma_{0} \!\Big( D \Big) \!\ldots \!\Big) \\ 
  \end{array}
  }\!\!\!\!\!\!}
  \label{eq:vqiip}
 \end{equation}
 \begin{equation}
  {\small\!\!\!\!\!\!\!\!\!{
  \begin{array}{l}
   \bar{p}_q \left( {\bf W}_{[q]} \right)
   \!=\! \bar{\sigma}_{d-1} \Big( H_{d-2} \bar{\sigma}_{d-2} \Big( H_{d-3} \bar{\sigma}_{d-3} \Big( \ldots \\
   H_{q+1} \bar{\sigma}_{q+1} \Big( \sum_{i_{q}} \bar{\sigma}_{q} \Big( \left\| {\bf W}_{[q]i_{q},:} \right\|_2^2 \bar{\sigma}_{q-1} \Big( \ldots 
   \!H_0 \bar{\sigma}_{0} \!\Big( D \Big) \!\ldots \!\Big) \\ 
  \end{array}
  }\!\!\!\!\!\!}
  \label{eq:vqip}
 \end{equation}
 we have that:
 \[
 {\!\!\small{
 \begin{array}{rcl}
  0 &\!\!\!\!\leq     {p}_q \left( {\bf W}_{[q]} \right) &\!\!\!\!\leq \left(H    {L} \right)^{d} \frac{D}{H_q H_{q-1}} \left\| {\bf W}_{[q]} \right\|^2_{F_{{\;}_{\;}}} \\
  0 &\!\!\!\!\leq \bar{p}_q
   \left( {\bf W}_{[q]} \right) &\!\!\!\!\leq \left(H\bar{L} \right)^{d} \frac{D}{H_q H_{q-1}} \left\| {\bf W}_{[q]} \right\|^2_{F_{{\;}_{\;}}} \\
 \end{array}
 }}
 \]
 where $H$, $L$ and $\bar{L}$ are geometric means of $H_q$, $L_q$ and $\bar{L}_q$.
 \label{cor:collapse}
\end{cor_collapse}

Note that $\sqrt{p_{[q]}}$ is an $F$-norm and $\sqrt{\bar{p}_{[q]}}$ is a 
quasi-$F$-norm\footnote{A quasi-$F$-norm is like an $F$-norm, except 
that it satisfies a weaker form of the triangle inequality 
$\sqrt{\bar{p}_{[q]}} ({\bf W}+{\bf W}') \leq c (\sqrt{\bar{p}_{[q]}} ({\bf W}) 
+ \sqrt{\bar{p}_{[q]}} ({\bf W}'))$ for some $c > 0$.} on weight-matrix space 
if $\sigma_q (0) = 0$ and $\sigma_q$ is concave, both being topologically 
equivalent to the Frobenius norm $\| \cdot \|_F$ (see supplementary).  The 
analogous result for the flat representation is as follows ($p_{\NN}$ and 
$\bar{p}_{\NN}$ are not norms - see supplementary for proof):
\begin{cor_collapsed}
 Using the notation of theorem \ref{cor:collapse}, defining 
 $\liip {\bf v}_{\NN}, {\bf v}_{\NN} \riip{{\bf g}_{\NN}} = p_{\NN} ({\bf 
 W}_{[0]}, {\bf W}_{[1]}, \ldots)$ and $\lip {\bf v}_{\NN}, {\bf v}_{\NN} 
 \rip{|{\bf g}_{\NN}|} = \bar{p}_{\NN} ({\bf W}_{[0]}, {\bf W}_{[1]}, \ldots)$, 
 where:
 \begin{equation}
  {\!\!\!\!\!\!\small{
  \begin{array}{l}
   p_{\NN} \left( {\bf W}_{[0]}, {\bf W}_{[1]}, \ldots \right)
   \!=\! \sigma_{d-1} \Big( {\sum}_{i_{d-2}} \left| W_{[d-1] 0,i_{d-2}} \right|^2 \ldots \\
   \sigma_{d-2} \Big( \ldots {\sum}_{i_0} \left| W_{[1]i_1,i_0} \right|^2 \sigma_{0} \Big( \left\| {\bf W}_{[0],i_0,:} \right\|_2^2 \Big) \Big) \Big) 
  \end{array}
  }\!\!\!\!\!\!}
  \label{eq:viip}
 \end{equation}
 \begin{equation}
  {\!\!\!\!\!\!\small{
  \begin{array}{l}
   \bar{p}_{\NN} \left( {\bf W}_{[0]}, {\bf W}_{[1]}, \ldots \right)
   \!=\! \bar{\sigma}_{d-1} \Big( {\sum}_{i_{d-2}} \left| W_{[d-1] 0,i_{d-2}} \right|^2 \ldots \\
   \bar{\sigma}_{d-2} \Big( \ldots {\sum}_{i_0} \left| W_{[1]i_1,i_0} \right|^2 \bar{\sigma}_{0} \Big( \left\| {\bf W}_{[0],i_0,:} \right\|_2^2 \Big) \Big) \Big) 
  \end{array}
  }\!\!\!\!\!\!}
  \label{eq:vip}
 \end{equation}
 we have that:
 \[
  {\small{
  \begin{array}{rccl}
   0 &\!\!\!\!\leq \liip {\bf v}_{\NN},{\bf v}_{\NN} \riip{\bf g}    &\!\!\!\!\leq     {L}^d \prod_q \left\|{\bf W}_{[q]}\right\|_{[q]_{{\;}_{\;}}\!\!\!\!}^2 &\!\!\!\!\leq \left( \frac{    {L}}{d} \sum_q \left\| {\bf W}_{[q]} \right\|_{[q]}^2 \right)^d \\
   0 &\!\!\!\!\leq \lip  {\bf v}_{\NN},{\bf v}_{\NN} \rip{|{\bf g}|} &\!\!\!\!\leq \bar{L}^d \prod_q \left\|{\bf W}_{[q]}\right\|_{[q]                    }^2 &\!\!\!\!\leq \left( \frac{\bar{L}}{d} \sum_q \left\| {\bf W}_{[q]} \right\|_{[q]}^2 \right)^d \\
  \end{array}
  }}
 \]
 where $L$ and $\bar{L}$ are geometric means of $L_q$ and $\bar{L}_q$.
 \label{cor:collapsed}
\end{cor_collapsed}

\subsection{Equivalent SVMs for Deep Networks} \label{sec:equivsvm}

We have shown that any deep network satisfying our assumptions can be flattened 
to obtain an equivalent flat representation (\ref{eq:wfinalform}) with feature 
map and metric defined by 
(\ref{eq:featmapis}) and (\ref{eq:wform}); and training the deep network is 
functionally equivalent to solving the regularised risk minimisation problem 
(\ref{eq:deepfeatminprob_prelim}):
\[
 \begin{array}{l}
  \mathop{\min}\limits_{{\bf v} \in \frkks{\NN}} \frac{1}{N} {\sum}_i \loss \left( y_i, \liip {\bf v}, {\bg{\featmapmkern}}_{\NN} \left( {\bf x}_i \right) \riip{{\bf g}_{\NN}} \right) + \lambda \reg_{\NN} \left( {\bf v} \right) \\
 \end{array}
\]
where $\frkks{\NN} \subseteq \frkks{}$ and $\reg_{\NN}$ are defined by (\ref{eq:flatspacedef}) and 
(\ref{eq:reginflat}).  We define an {\em equivalent} (indefinite) SVM for a 
given deep network to be an indefinite SVM using the same feature map 
${\bg{\featmapmkern}}_{\NN} : \infset{R}^D \to \frkks{}$ and metric 
${\bf g}_{\NN}$ as the deep network (in flat form) that solves the regularised 
risk minimisation problem:
\begin{equation}
 {\small{
 \begin{array}{l}
  \mathop{\min}\limits_{{\bf v} \in \frkks{\SVM}} \frac{1}{N} {\sum}_i \loss \left( y_i, \liip {\bf v}, {\bg{\featmapmkern}}_{\NN} \left( {\bf x}_i \right) \riip{{\bf g}_{\NN}} \right) + \lambda \reg_{\SVM} \left( {\bf v} \right) \\
 \end{array}
 }}
 \label{eq:equivsvmtrain}
\end{equation}
where $\frkks{\SVM} = \frkks{}$ and $\reg_{\SVM} ({\bf v}) = \liip {\bf v}, 
{\bf v} \riip{{\bf g}_{\NN}}$.  Clearly the feature map is countably infinite 
dimensional, so the primal form of the equivalent SVM is not useful; however we 
may use the {\krein} kernel trick to encapsulate the feature map in a {\krein} 
kernel and then solve (\ref{eq:trainSVMalpha}) to get $f ({\bf x}) = \sum_i 
\alpha_i K_{\NN} ({\bf x}, {\bf x}_i)$:
\begin{th_kernels}
 Let the feature map ${\bg{\featmapmkern}}_{\NN} : \infset{R}^D \to 
 \frkks{}$ and metric ${\bf g}_{\NN}$ be defined by the deep network 
 (\ref{eq:deepnetform}) by (\ref{eq:featmapis}), (\ref{eq:wform}).  Then:
 \begin{equation}
  \begin{array}{l}
   K_{\NN} \left( {\bf x}, {\bf x}' \right)
   = \liip {\bg{\featmapmkern}}_{\NN} \left( {\bf x} \right), {\bg{\featmapmkern}}_{\NN} \left( {\bf x}' \right) \riip{{\bf g}_{\NN}}
   = \sigma_{d-1} ( \ldots \\
   \!\!\!H_{d-2} \sigma_{d-2} ( H_{d-3} \ldots H_1 \sigma_{1} ( H_0 \sigma_{0} ( \lip {\bf x}, {\bf x}' \rip{{\bf 1}} ) ) ) ) \\
  \end{array}
  \label{eq:whatisK}
 \end{equation}
 is the corresponding {\krein} kernel, and:
 \begin{equation}
  {\!\!
  \begin{array}{l}
   \bar{K}_{\NN} \left( {\bf x}, {\bf x}' \right)
   = \lip {\bg{\featmapmkern}}_{\NN} \left( {\bf x} \right), {\bg{\featmapmkern}}_{\NN} \left( {\bf x}' \right) \rip{|{\bf g}_{\NN}|}
   = \bar{\sigma}_{d-1} ( \ldots \\
   \!\!\!H_{d-2} \bar{\sigma}_{d-2} ( H_{d-3} \ldots H_1 \bar{\sigma}_{1} ( H_0 \bar{\sigma}_{0} ( \lip {\bf x}, {\bf x}' \rip{{\bf 1}} ) ) ) ) \\
  \end{array}
  \!\!}
  \label{eq:whatisbarK}
 \end{equation}
 is the associated kernel ($\bar{\sigma}_{q}$ is the associated activation 
 function - that is, if $\sigma_{q} (\xi) = \sum_i a_{[q]i} \xi^i$ 
 then $\bar{\sigma}_{q} (\xi) = \sum_i |a_{[q]i}| \xi^i$ (e.g. see table 1 in 
 the supplementary)).  Note that if $\sigma_q$ is convex (eg linear, 
 exponential) then $K = \bar{K}$.
 \label{th:kernels}
\end{th_kernels}
\begin{proof}
The proof follows by direct application of the definitions (theorem 
\ref{th:collapse} and lemma \ref{th:pushforward}).  See supplementary.
\end{proof}

An indefinite SVM using {\krein} kernel $K_{\NN}$ trained on a 
particular dataset will learn a relation $f: \infset{R}^D \to \infset{R}$ of 
the same form (in primal representation), but with different weights, as that 
learned by the deep network (flat representation) from whose structure 
(\ref{eq:whatisK}) was derived and that has been trained on the same dataset.

The differences between the deep network and its equivalent SVM are (a) the 
definition of the restricted feature space $\frkks{\NNSVM}$ and (b) the form of 
regularisation $\reg_{\NNSVM}$.  Note that the space of $\frkks{\NN}$ of 
realisable weights of the deep network is smaller than the space of realisable 
weights for the equivalent SVM.  
We may therefore 
expect that the capacity of the equivalent SVM will be larger than the capacity 
of the deep network from which it was derived (a fact that we demonstrate 
shortly).

\section{Capacity, Sparsity and Convergence} \label{sec:indefsvm}

In this section we use the flat (and semi-flat) representations of deep 
networks, and the observed connection between these and indefinite SVMs, to 
analyse the capacity, sparsity and convergence of deep networks.  First we 
apply Rademacher complexity theory to the flat representation of the deep 
network and show that it is bounded by the (known \cite{Ong3}) Rademacher 
complexity of the equivalent indefinite SVM, with depth and width dependence 
similar to those reported elsewhere \cite{Bar8,Ney2}.  Next, we derive a 
stronger bound in the case where the activation functions $\sigma_q$ are 
concave (which is typical) using the properties of the weight space 
$\frkks{\NN}$, which decouples capacity and network width if $\sigma_q$ is 
bounded (tanh-like).  Finally, we use the semi-flat representation of the deep 
network to derive a bound on the $L_p$-``norm'', with $p = \frac{2}{d} \in 
(0,1]$, of the weight vector ${\bf v}$, demonstrating that deep networks 
actually implement a form of bridge regression \cite{Fra2} approaching the 
best-subset limit \cite{Bea1,Hoc3} on the flat representation, which we find 
particularly interesting in light of recent developments regarding the promise 
of best-subset selection, particularly in noisy scenarios \cite{Ber2,Has4}.

\subsection{Rademacher Complexity Analysis}

The Rademacher complexity $\mathcal{R}_N (\unitnormballkrein{})$ of a 
hypothesis space $\unitnormballkrein{}$ of real-valued functions is a measure 
of its capacity.  Let $\{ {\bf x}_i \sim \nu : i \in \infset{N}_N \}$ be a set 
of vectors drawn from distribution $\nu$ and let $\epsilon_0, \epsilon_1, 
\ldots \in \{ -1,1 \}$ be Rademacher random variables.  Then by definition 
\cite{Men2}:
\begin{equation}
 \begin{array}{rl}
  \mathcal{R}_{N} \left( \unitnormballkrein{} \,\right) 
  &\!\!\!\!= \mathbb{E}_{\nu,\epsilon} \left[ {\rm sup}_{f \in \unitnormballkrein{}} \left| \frac{1}{N} {\sum}_i \epsilon_i f \left( {\bf x}_i \right) \right|  \right] \\
 \end{array}
 \label{eq:deepnetcompbound}
\end{equation}
Rademacher complexity may be used in uniform convergence analysis to bound 
how quickly the empirical loss $\hat{\mathcal{L}} (f) = \frac{1}{N} \sum_i 
\loss (y_i,f({\bf x}_i))$ converges to the expected loss $\mathcal{L} (f) = 
\mathbb{E} [\loss (y,f({\bf x}))]$ for a given $f \in \unitnormballkrein{}$ 
as the dataset size increases.  For example, if $\loss$ is Lipschitz with 
constant $L_\loss$ and bounded by $c$ then for all $\delta \in \infset{R}_+$, 
with probability $>1-\delta$:
\[
 \begin{array}{l}
  \mathcal{L} \!\left( f \right) \leq \hat{\mathcal{L}} \!\left( f \right) + 2 L_{\loss} \mathcal{R}_N \!\left( \unitnormballkrein{} \right) + c \sqrt{\frac{\log \left( \frac{1}{\delta} \right)}{2N}}
 \end{array}
\]

It is well known (eg \cite[Theorem 12]{Bar1})) that if $\unitnormballkrein{} 
\subseteq \bar{\unitnormballkrein{}}$ then $\mathcal{R}_N(\unitnormballkrein{}) 
\leq \mathcal{R}_N (\bar{\unitnormballkrein{}})$.  Thus we may bound Rademacher 
complexity by showing that the hypothesis space is a subset of a larger 
hypothesis space whose Rademacher complexity is known.  We will bound 
the complexity of the deep network by showing that its hypothesis space is a 
subset of the hypothesis space of the associated equivalent (indefinite) SVM.

In the usual, non-flat form a trained deep network has the form 
(\ref{eq:deepnetform}), where the weight matrices ${\bf W}_{[q]}$ are selected 
to solve the regularised risk minimisation problem (\ref{eq:origdeepreg}).  If 
we interpret $\lambda \geq 0$ in (\ref{eq:origdeepreg}) as a Lagrange 
multiplier, this is equivalent to the constrained optimisation problem:
\begin{equation}
 \begin{array}{l}
  \!\!\!\!\mathop{\arg\min}\limits_{{\bf W}_{[q]} \in \infset{R}^{H_q \times H_{q-1}} : \nnregscale \sum_q \left\| {\bf W}_{[q]} \right\|_{[q]}^2 \leq R_{\NN}\!\!\!\!} \frac{1}{N} \sum_i \loss \left( y_i, f \left( {\bf x}_i \right) \right)\!\!\!\!
 \end{array}
 \label{eq:trainingprobNN}
\end{equation}
for appropriate $R_{\NN}$.  Hence the hypothesis space is:
\begin{equation}
 {\!\!\!\small{
 \begin{array}{l}
  \unitnormballkrein{\NN} = \left\{ \!\!\!\begin{array}{r} \left. f \left( {\bf x} \right) = \sigma_{d-1} \left( {\bf W}_{[d-1]} \ldots \sigma_0 \left( {\bf W}_{[0]} {\bf x} \right) \right) \right| \ldots \\ \ldots \nnregscale \sum_q \left\| {\bf W}_{[q]} \right\|_{[q]}^2 \leq R_{\NN} \end{array} \!\!\!\right\} 
 \end{array}
 }\!\!\!}
 \label{eq:hypothesisNN}
\end{equation}

Likewise, the regularised risk minimisation problem (\ref{eq:equivsvmtrain}) 
for the equivalent SVM defined by the deep network can be rewritten as a 
constrained optimisation problem:\footnote{In formulating this we use the fact 
that we are using min in the loose sense (allowing local minima), so we may 
apply Lagrange multiplier theory, which in this case guarantees only local 
optima as the regularisation term is non-convex.}
\begin{equation}
 \begin{array}{l}
  \mathop{\arg\min}\limits_{{\bf v} \in \frkks{\NN} : \liip {\bf v}_{\NN}, {\bf v}_{\NN} \riip{{\bf g}_{\NN}} \leq R_{\SVM}} \frac{1}{N} \sum_i \loss \left( y_i, f \left( {\bf x}_i \right) \right)
 \end{array}
 \label{eq:trainingprobequivSVM}
\end{equation}
for some $R_{\SVM}$, so the corresponding hypothesis space is:
\begin{equation}
 {\small
 \!\!\!\!\!\!\!\!\!\!\!\!\!\begin{array}{rl}
  \unitnormballkrein{\SVM} 
  &\!\!\!\!= \left\{ \left. f = \liip {\bf v}, {\bg{\featmapmkern}}_{\NN} \left( \cdot \right) \riip{{\bf g}_{\NN}} \right| {\bf v} \in \frkks{} \wedge \liip {\bf v}, {\bf v} \riip{{\bf g}_{\NN}} \leq R_{\SVM} \right\} \\
  &\!\!\!\!= \{ \left. f \in \rkks{K_{\NN}} \right| \liip f,f \riip{\rkks{K_{\NN}}} \leq R_{\SVM} \} \\
 \end{array}\!\!\!\!\!\!\!\!\!\!\!\!\!
 }
 \label{eq:hypothesisSVM}
\end{equation}
which is a ball of radius $R_{\SVM}$ in {\RKKS} $\rkks{{K}_{\NN}}$ specified by 
the {\krein} kernel defined by the form of the deep network as per 
(\ref{eq:whatisK}) in theorem \ref{th:kernels}.

In the equivalent, flat form, a trained deep network has the form 
(\ref{eq:wfinalform}), where the weight vector ${\bf v}_{\NN} \in \frkks{\NN}$ 
are selected to solve the regularised risk minimisation problem 
(\ref{eq:deepfeatminprob_prelim}).  Applying the usual procedure, the 
hypothesis space is:
\begin{equation}
 {
 \!\!\!\!\!\!\!\!\!\!\!\!\!\begin{array}{rl}
  \unitnormballkrein{\NN} 
  &\!\!\!\!= \left\{ \left. f = \liip {\bf v}, {\bg{\featmapmkern}}_{\NN} \left( \cdot \right) \riip{{\bf g}_{\NN}} \right| {\bf v} \in \frkks{\NN} \wedge \reg_{\NN} \left( {\bf v} \right) \leq\! R_{\NN} \right\} \\
 \end{array}\!\!\!\!\!\!\!\!\!\!\!\!\!
 }
 \label{eq:hypothesisNNflat}
\end{equation}
noting that (\ref{eq:hypothesisNN}) and (\ref{eq:hypothesisNNflat}) are in fact 
identical as the conditions in (\ref{eq:hypothesisNNflat}) simply assert that 
${\bf v}_{\NN}$ corresponds to some set of weight matrices ${\bf W}_{[q]}$ 
satisfying the conditions of (\ref{eq:hypothesisNN}), where $f$ is functionally 
equivalent for either representation.\footnote{That is, ${\bf v}_{\NN} = 
\bigodot_q \bg{\featmap}_{q+} ({\bf W}_{[q]})$ as per 
(\ref{eq:featmapis}-\ref{eq:featmapdetail}).}

Given the above, using theorem \ref{cor:collapsed} and 
(\ref{eq:trainingprobNN}), we have that:
\[
 {\small{
 \begin{array}{l}
  \liip {\bf v}_{\NN}, {\bf v}_{\NN} \riip{\bf g} \leq \left( \frac{L}{d} \sum_q \left\| {\bf W}_{[q]} \right\|_{[q]}^2 \right)^d \leq \left( L R_{\NN} \right)^d
 \end{array}
 }}
\]
which, recalling that $\frkks{\NN} \subset \frkks{\SVM}$ and using 
(\ref{eq:hypothesisSVM}), implies that:
\begin{equation}
 {\!\!\small{
 \begin{array}{l}
  \unitnormballkrein{\NN} \subseteq \unitnormballkrein{{\SVM}} \; \big| R_{\SVM} = \left( LR_{\NN} \right)^d 
 \end{array}
 }}
 \label{eq:firstNNradbound}
\end{equation}
and hence $\mathcal{R}_N (\unitnormballkrein{\NN}) \leq \mathcal{R}_N 
(\unitnormballkrein{{\SVM}})$ if $R_{\SVM} = (LR_{\NN})^d$.  Moreover as noted 
in \cite[Lemma 9]{Ong3}, the Rademacher complexity in {\RKKS} $\rkks{K_{\NN}}$ 
is equivalent to the Rademacher complexity in the associated {\RKHS} 
$\rkhs{\bar{K}_{\NN}}$, so we can bound 
$\mathcal{R}_N (\unitnormballkrein{{\SVM}})$ as per the following theorem 
\cite{Ong3}:
\begin{th_boundsvm}
 Let $K_{\NN}$ be a {\krein} kernel and $\bar{K}_{\NN}$ be its associated 
 kernel, such that ${\bf x} \to \bar{K}_{\NN} ({\bf x}, {\bf x}) \in L_1 
 (\infset{R}^D,\nu)$ and $\bar{K}_{\NN} ({\bf x}, {\bf x}) \geq 0$ $\forall {\bf 
 x} \in \infset{R}^D$.  Then:
 \begin{equation}
  {\small{
  \begin{array}{l}
   \mathcal{R}_{N} \left( \unitnormballkrein{\SVM} \,\right) \leq \frac{1}{\sqrt{N}} \left( R_{\SVM} \int_{{\bf x} \in \infset{R}^D} \bar{K}_{\NN} \left( {\bf x}, {\bf x} \right) d\nu ({\bf x}) \right)^{\frac{1}{2}}
  \end{array}
  }}
  \label{eq:radboundsvm}
 \end{equation}
 \label{th:boundsvm}
\end{th_boundsvm}
\begin{proof}
See \cite{Ong3}.  Alternatively we provide a weight-space proof in the 
supplementary material.
\end{proof}

Combining (\ref{eq:firstNNradbound}) and theorem \ref{th:boundsvm} we obtain 
the following bound on the Rademacher complexity of the deep network:
\begin{equation}
{{
 \begin{array}{rl}
  \mathcal{R}_N \left( \unitnormballkrein{\NN} \right)  
\leq \sqrt{\frac{\left( LR_{\NN} \right)^{d}}{N} \int_{{\bf x} \in \infset{R}^D} \bar{K}_{\NN} \left( {\bf x}, {\bf x} \right) d\nu \left( {\bf x} \right)} \\
 \end{array}
}}
 \label{eq:NNriskbound}
\end{equation}
This bound grows exponentially with depth $d$ and polynomially (order 
$\frac{d}{2}$) with activation function Lipschitz constant $L$.  Width and data 
distribution dependence come through the integral of the equivalent associated 
kernel $\bar{K}_{\NN}$ and thus depend on the network structure.  For example, 
a linear network has $K_{\NN} ({\bf x}, {\bf x}') = \bar{K}_{\NN} ({\bf x}, 
{\bf x}) = H^d \| {\bf x} \|_2^2$, so (\ref{eq:NNriskbound}) reduces to:
\begin{equation}
 \begin{array}{rl}
  \mathcal{R}_N \!\left( \unitnormballkrein{\NN} \right)  \leq
  \sqrt{\frac{\left( HLR_{\NN} \right)^{d}}{{N}} \mathbb{E}_\nu \left[ \left\| X \right\|_2^2 \right]} \\
 \end{array}
 \label{eq:NNriskbound_linear}
\end{equation}
which grows polynomially (order $\frac{d}{2}$) with width, which is similar to 
bounds reported elsewhere \cite{Bar8,Ney2}.  However it follows from the 
convexity of $\bar{K}$ that (\ref{eq:NNriskbound_linear}) is the best-case 
behaviour of the bound (\ref{eq:NNriskbound}), and table 1 in the supplementary 
indicates that the bound can be very loose.

The difficulty with (\ref{eq:NNriskbound}) arises from the presense of the 
associated kernel $\bar{K}_{\NN}$.  Roughly speaking, $\bar{K}_{\NN}$ enters 
the picture in the proof of theorem \ref{th:boundsvm} when we apply the 
Cauchy-Schwarz inequality to separate out the feature-map dependence - that 
is, $\liip {\bf v}, \bg{\featmapmkern}_{\NN} ({\bf x}) \riip{\bf g}^2 \leq \lip 
{\bf v}, {\bf v} \rip{|{\bf g}|} \lip \bg{\featmapmkern} ({\bf x}), 
\bg{\featmapmkern} ({\bf x}) \rip{|{\bf g}|}$.  There is no obvious way around 
this in the general case, but as an alternative we may use the fact that ${\bf 
v}_{\NN} \in \frkks{\NN}$ to cast $\liip {\bf v}, \bg{\featmapmkern}_{\NN}({\bf 
x}) \riip{\bf g}$ into weight matrix space before separating factors, which 
leads to the following in the concave case (proof in supplementary):
\begin{th_tightboundnn}
 Let $\sigma_q$ be concave on $\infset{R}_+$, $\sigma_q(0) = 0$ and $\sigma_q 
 (-\xi) = -\sigma_q (\xi)$ in addition to the usual assumptions.  Let:
 \[
  {\small{
  \begin{array}{l}
   \!\!\chi_{\NN} \!\left( \xi \right) \!=\! \sigma_{d-1} \!\left( d \sqrt{H_{d-2}} \sigma_{d-2} \!\left( d \sqrt{H_{d-3}} \sigma_{d-3} \!\left( \ldots d \sqrt{H_0} \sigma_{0} \!\left( \xi \right) \right) \right) \right) \\
  \end{array}
  }}
 \]
 If ${\bf x} \to \chi_{\NN} (\| {\bf x} \|_2) \in L_1 (\infset{R}^D, \nu)$ then:
 \[
  {\small{
  \begin{array}{rl}
   \mathcal{R}_N \left( \unitnormballkrein{\NN} \,\right) \leq \frac{\max \{ 1,R_{\NN}^d \}}{\sqrt{N}} \sqrt{\int_{{\bf x} \in \infset{R}^D} \chi_{\NN}^2 \left( \left\| {\bf x} \right\|_2 \right) d\nu \left( {\bf x} \right)} \\
  \end{array}
  }}
 \]
 Moreover if $\sigma_q$ is unbounded for all $q$ then:
 \[
  {\small{
   \begin{array}{l}
    \mathcal{R}_N \left( \unitnormballkrein{\NN} \right) \leq \frac{\max \left\{ 1,R_{\NN}^d \right\}}{\sqrt{N}} \left( d\sqrt{H}L \right)^d \sqrt{ \mathbb{E}_{\nu} \left[ \left\| X \right\|_2^2 \right]}
   \end{array}
  }}
 \]
 and otherwise, if $\sigma_{q} (\xi) \leq 1$ $\forall \xi \in \infset{R}_+$ for 
 some $q \in \infset{N}_d$ then:
 \[
  {\small{
  \begin{array}{rl}
   \mathcal{R}_N \left( \unitnormballkrein{\NN} \,\right) 
   &\!\!\!\!\leq \frac{\max \left\{ 1,R_{\NN}^d \right\}}{\sqrt{N}} \left( d \sqrt{H_{[q+]}} L_{[q+]} \right)^{d-q-1} 
  \end{array}
  }}
 \]
 where $H_{[q+]} = \GM (H_{q+1},H_{q+2},\ldots,H_{d-1})$ and $L_{[q+]} = \GM 
 (L_{q+1},L_{q+2},\ldots,L_{d-1})$ are geometric means.
 \label{th:tightboundnn}
\end{th_tightboundnn}

Neglecting width dependence, like (\ref{eq:NNriskbound}), this bound on 
Rademacher complexity grows exponentially with depth $d$ as $R_{\NN}^d$.  
However, unlike (\ref{eq:NNriskbound}), the {\em worst-case} width dependency 
(not best case, as in (\ref{eq:NNriskbound})) is polynomial of order $d$, where 
the order decreases if bounded activation functions are used in the network and 
vanishes entirely if the output layer activation function $\sigma_{d-1}$ is 
bounded.

\subsection{Sparsity Analysis}

Considering the form of the semi-flat (\ref{eq:semideepflat}) and flat 
(\ref{eq:wfinalform}) representations of the deep network, and in light of 
theorem \ref{cor:collapse}, we see that the (flat) weight vector ${\bf 
v}_{\NN}$ is an elementwise product of a set of $d$ vectors with constrained 
(regularised) $2$-norms.  Thus we might expect that ${\bf v}_{\NN}$ will have a 
constrained $\frac{2}{d}$-``norm'', which would imply a form of (effectively) 
sparsity-inducing regularisation in the flat representation.  Precisely (proof 
in supplementary material):
\begin{th_lownormbound}
 For a given deep network satisfying our assumptions, using the notations 
 described, we have that $\| {\bf v}_{[q]} \|_\infty \leq dR_{\NN}$, $\| {\bf 
 v}_{\NN} \|_\infty \leq R_{\NN}^d$, and:
 \begin{equation}
  \begin{array}{l}
   \left\| {\bf g}_{\NN} \odot {\bf v}_{\NN} \right\|_{2/d} \leq \left( H\bar{L} \right)^d \frac{DR_{\NN}}{H^2} \\
  \end{array}
  \label{eq:coolnormbound}
 \end{equation}
 where $\| {\bf a} \|_\gamma = \sum_i |a_i|^\gamma$ is the $L_\gamma$-``norm'' 
 $\forall \gamma \in [0,1]$.
 \label{th:lownormbound}
\end{th_lownormbound}

Thus we see that when we train the deep network using weight-decay we are 
effectively selecting the weight vector in the flat representation using 
$\frac{2}{d}$-``norm'' regularised (bridge) regression \cite{Fra2}, approaching 
best-subset regression \cite{Bea1,Hoc3} for larger $d$, which has been shown 
\cite{Ber2,Has4} to have significant benefits, particularly for noisy data, as 
it may help explain the remarkable performance of deep networks.  We finish 
with the following corollary of theorem \ref{th:lownormbound}, which 
characterises the distribution of magnitudes of weights in ${\bf v}_{\NN}$ 
(proof in supplementary):
\begin{cor_sparsity}
 The total weight vector ${\bf g}_{\NN} \odot {\bf v}_{\NN}$ of the flat 
 representation is $\epsilon$-sparse - that is, there are at most $\lfloor 
 (H\bar{L} )^d \frac{DR_{\NN}}{H^2} \epsilon^{-\frac{2}{d}} \rfloor$ elements 
 in this vector with magnitude $| g_{\NN i} v_{\NN i} |$ greater than $\epsilon 
 \in (0, dR_{\NN} \| {\bf g}_{\NN} \|_\infty]$.
 \label{cor:sparsity}
\end{cor_sparsity}

That is, as $\epsilon$ increases, the number of elements in the (total) weight 
vector ${\bf g}_{\NN} \odot {\bf v}_{\NN}$ that have magnitude greater than 
$\epsilon$ will decrease as the reciprocal of $\epsilon^{2/d}$.  So we 
may expect a relatively small number (at most $\lfloor (H\bar{L} )^d 
\frac{DR_{\NN}}{H^2} \rfloor$) of dominant features with (relatively, in 
magnitude) large total weight, combined with a much larger number of features 
with relatively small total weight (we call this sort of ``approximate'' form 
of sparsity $\epsilon$-sparsity).  This happens even though the feature map is 
infinite dimensional in general.

\section{Conclusions}

We have explored a novel connection between deep networks and learning in 
reproducing kernel {\krein} space.  We have shown how a deep network can be 
converted to an equivalent (flat) form consisting of a fixed non-linear feature 
map followed by a learned linear projection onto $\infset{R}$, which is 
functionally identical to an indefinite SVM.  Using this, we have explored 
capacity and generalisation in deep networks by bounding in terms of capacity 
in regularised learning in reproducing kernel {\krein} space; and shown that 
the flat form is effectively implementing sparsity-inducing bridge regression, 
approaching best-subset regression as the depth of the network increases.

{\small{
\bibliographystyle{plainnat}
\bibliography{universalmod}
}}

\clearpage
\normalsize
\setcounter{section}{0}
\setcounter{page}{1}

\twocolumn[\centering\bf\LARGE From deep to Shallow: Equivalent Forms of Deep Networks in Reproducing Kernel {\krein} Space and Indefinite Support Vector Machines - Supplementary Material \\ $\;$ \\  $\;$ \\  $\;$ \\ ]

\section{Supplementary: Reproducing Kernel {\krein} Space - Standard Approach}

Reproducing kernel Hilbert space theory \cite{Aro1,Ste3,Sha3} is ubiquitous in 
machine learning \cite{Cor1,Cho7,Cri4,Gen2,Gon7,Her2,Li7,Mul2,Sch3,Sch6,Sch24,Sha3,Smo12}.  
Motivated by the observation that indefinite kernels outperform {\RKHS} kernels 
in some cases \cite{Lin11,Lus1,Haa1,Yin2}, reproducing kernel {\krein} spaces 
({\RKKS}s) have been studied in \cite{Ong3,Ogl1,Ogl2,Loo1,Sch27}.  In the 
supplementary we present a quick overview of reproducing kernel {\krein} space 
theory (see \cite{Bog1,Azi1,Ong3,Ogl1,Ogl2} for alternatives) from a more 
conventional standpoint than in the body of the paper.  As per 
\cite{Bog1,Azi1}, we being by defining {\krein} space:
\begin{def_krein}
 A {\krein} space $\{ \rkks{}, \liip \cdot,\cdot \riip{\rkks{}} \}$ is a vector 
 space $\rkks{}$ over $\infset{R}$ equipped with an indefinite inner product 
 $\liip \cdot, \cdot \riip{\rkks{}} : \rkks{} \times \rkks{} \to \infset{R}$ 
 that may be decomposed into a direct difference $\rkks{} = \rkhs{+} \ominus 
 \rkhs{-}$ of $\liip \cdot, \cdot \riip{\rkks{}}$-orthogonal Hilbert spaces 
 $\rkhs{\pm}$ (that is, $\liip f_+,g_- \riip{\rkks{}} = 0$ $\forall f_+ \in 
 \rkhs{+}, g_- \in \rkhs{-}$) such that:
 \[
  \begin{array}{l}
   \liip f,g \riip{\rkks{}} = \lip f_+,g_+ \rip{\rkhs{+}} - \lip f_-,g_- \rip{\rkhs{-}}
  \end{array}
 \]
 where $f = f_+ \dsum f_-$, $g = g_+ \dsum g_-$, and $f_{\pm}, g_{\pm} \in 
 \rkhs{\pm}$ (here $\dsum$ denotes the orthogonal sum).  The {\em associated 
 Hilbert space} $\{ \bar{\rkks{}}, \lip \cdot,\cdot \rip{\bar{\rkks{}}} \}$ 
 is a Hilbert space $\bar{\rkks{}} = \rkhs{+} \oplus \rkhs{-}$ over 
 $\infset{R}$ with:
 \[
  \begin{array}{l}
   \lip f,g \rip{\bar{\rkks{}}} = \lip f_+,g_+ \rip{\rkhs{+}} + \lip f_-,g_- \rip{\rkhs{-}}
  \end{array}
 \]
 The {\em strong topology} on $\rkks{}$ is induced by the metric $d^2(f,g) = \| 
 f-g \|_{\bar{\subrkks{}}}^2 = \lip f-g,g-g \rip{\bar{\rkks{}}}$.
 \label{def:krein}
\end{def_krein}

Note that the decomposition of $\rkks{}$ into $\rkhs{\pm}$ is not unique in 
general.  However the strong topology induced by the associated Hilbert space 
norm is independent of the decomposition \cite{Ogl2}.  Reproducing kernel 
{\krein} space is defined as \cite{Alp2,Ong3}:
\begin{def_rkks}[Reproducing Kernel {\krein} Space]
 A {\em reproducing kernel {\krein} space ({\RKKS})} $\rkks{}$ is a {\krein} 
 space of functions $f : \infset{X} \to \infset{R}$ such that $\forall x \in 
 \infset{X}$ the point evaluation functional $L_x : \rkhs{} \to \infset{R}$, 
 $L_x (f) = f(x)$, is continuous with respect to the strong topology.

 (\cite[Proposition 6]{Ong3}) 
 For every {\RKKS} $\rkks{}$ there exists a symmetric reproducing ({\krein}) 
 kernel $K : \infset{X} \times \infset{X} \to \infset{R}$, where $f(x) = \liip 
 f,K(x,\cdot) \riip{\rkks{}}$ $\forall f \in \rkks{}$ and $K (x,x') = \liip K 
 (x,\cdot), K(\cdot,x') \riip{\rkks{}}$, and $K$ can be decomposed as $K = K_+ 
 - K_-$ where $K_{\pm}$ are positive-definite reproducing kernels for 
 $\rkhs{\pm}$.  The associated Hilbert space $\bar{\rkks{}}$ is a {\RKHS} with 
 reproducing kernel (associated kernel) $\bar{K} = K_+ + K_-$.
 \label{def:rkks}
 \label{th:rkks}
\end{def_rkks}

Any {\krein} kernel $K$ that can be decomposed as $K = K_+ - K_-$ defines a 
reproducing kernel {\krein} space $\rkks{K}$, and it can be shown that any 
symmetric, jointly analytic $K : \infset{X} \times \infset{X} \to \infset{R}$ 
is a {\krein} kernel \cite{Alp2}.  In this paper we are primarily concerned 
with {\krein} kernels of the form:
\[
 \begin{array}{l}
  K \left( {\bf x}, {\bf x}' \right) = k \left( \liip {\bf x}, {\bf x}' \riip{{\bf g}} \right)
 \end{array}
\]
where we use the notation $\liip {\bf x}, {\bf x}' \riip{{\bf g}} = \sum_i 
g_i x_i x'_i$ to represent the weighted indefinite inner product (in the 
special case ${\bf g} > {\bf 0}$ we instead write $\lip {\bf x}, {\bf x}' 
\rip{{\bf g}}$ to emphasise that this is an inner-product in this case).  
Clearly if $k$ is analytic then $K$ must be a {\krein}-kernel.  Indeed, if $k$ 
is {\em entire} then we can construct the Taylor expansion $k (\chi) = \sum_i 
a_i \chi^i$, and it follows that:
\[
 \begin{array}{rlll}
  K \left( {\bf x}, {\bf x}' \right) 
  &\!\!\!= k \left( \liip {\bf x}, {\bf x}' \riip{{\bf g}} \right) \\
  &\!\!\!= k_+ \left( \liip {\bf x}, {\bf x}' \riip{{\bf g}} \right) - k_- \left( \liip {\bf x}, {\bf x}' \riip{{\bf g}} \right) \\
  &\!\!\!= \liip \bg{\featmapmkern} ({\bf x}), \bg{\featmapmkern} ({\bf x}') \riip{\bg{\scriptstyle\gamma} \odot \bg{\scriptstyle\featmapmkern} ({\bf g})} \\
  \bar{K} \left( {\bf x}, {\bf x}' \right) 
  &\!\!\!= \bar{k} \left( \liip {\bf x}, {\bf x}' \riip{{\bf g}} \right) \\
  &\!\!\!= k_+ \left( \liip {\bf x}, {\bf x}' \riip{{\bf g}} \right) + k_- \left( \liip {\bf x}, {\bf x}' \riip{{\bf g}} \right) \\
  &\!\!\!= \lip \bg{\featmapmkern} ({\bf x}), \bg{\featmapmkern} ({\bf x}') \rip{|\bg{\scriptstyle\gamma} \odot \bg{\scriptstyle\featmapmkern} ({\bf g})|}\\
 \end{array}
\]
where, using multi-index notation ${\bg{\featmapmkern}} ({\bf x}) = [ 
{{\featmapmkern}}_{\bf i} ({\bf x})]_{{\bf i} \in \infset{N}^{{\rm dim} ({\bf x})}}$ and 
${\bg{\gamma}} = [ {{\gamma}}_{\bf i}]_{{\bf i} \in \infset{N}^{{\rm dim} ({\bf x})}}$, 
where ${\featmapmkern}_{\bf i}({\bf x}) = \prod_j x_j^{i_j}$ and ${\gamma}_{\bf 
i} = (\frac{{\rm sum} ({\bf i})}{\prod_j i_j!}) a_{{\rm sum} ({\bf i})}$.  We 
may further note that:
\[
 \begin{array}{rll}
  K_\pm \left( {\bf x}, {\bf x}' \right) 
  &\!\!\!= k_\pm \left( \liip {\bf x}, {\bf x}' \riip{{\bf g}} \right) 
  &\!\!\!= \lip \bg{\featmapmkern} ({\bf x}), \bg{\featmapmkern} ({\bf x}') \rip{(\pm\bg{\scriptstyle\gamma} \odot \bg{\scriptstyle\featmapmkern} ({\bf g}))_+} \\
 \end{array}
\]
where $(a)_+ = \max \{ a,0 \}$ and $({\bf a})_+ = [ (a_0)_+, (a_1)_+, \ldots 
]$.  See table \ref{tab:actfuns} for examples of functions $k$ defining {\krein} 
kernels, along with the functions $\bar{k}$ defining the associated kernel.  
Importantly in our context, we note that this expansion applies to more general 
entire functions of $m$-indefinite-inner-products, specifically:
\begin{equation}
 \begin{array}{l}
  k \left( \lmiip {\bf x}, {\bf x}', \ldots, {\bf x}'''' \rmiip{m,{\bf g}} \right) \\
  \;= k_+ \!\left( \lmiip {\bf x}, {\bf x}', \ldots, {\bf x}'''' \rmiip{m,{\bf g}} \right) - k_- \!\left( \lmiip {\bf x}, {\bf x}', \ldots, {\bf x}'''' \rmiip{m,{\bf g}} \right) \\
  \;= \lmiip \bg{\featmapmkern} ({\bf x}), \bg{\featmapmkern} ({\bf x}'), \ldots, \bg{\featmapmkern} ({\bf x}') \rmiip{m,\bg{\scriptstyle\gamma} \odot \bg{\scriptstyle\featmapmkern} ({\bf g})} \\
  \bar{k} \left( \lmiip {\bf x}, {\bf x}', \ldots, {\bf x}'''' \rmiip{m,{\bf g}} \right) \\
  \;= k_+ \!\left( \lmiip {\bf x}, {\bf x}', \ldots, {\bf x}'''' \rmiip{m,{\bf g}} \right) + k_- \!\left( \lmiip {\bf x}, {\bf x}', \ldots, {\bf x}'''' \rmiip{m,{\bf g}} \right) \\
  \;= \lmip \bg{\featmapmkern} ({\bf x}), \bg{\featmapmkern} ({\bf x}'), \ldots, \bg{\featmapmkern} ({\bf x}') \rmip{m,|\bg{\scriptstyle\gamma} \odot \bg{\scriptstyle\featmapmkern} ({\bf g})|}\\
 \end{array}
 \label{eq:kexpand}
\end{equation}
where ${\featmapmkern}_{\bf i}$ and ${\gamma}_{\bf i}$ are as before, 
independent of $m$ and ${\bf g}$, and:\footnote{We note in passing that 
$\bar{K} ({\bf x}, {\bf x}', \ldots) = \bar{k} (\lmiip {\bf x}, {\bf x}', 
\ldots \rmiip{m,{\bf g}})$ is an $m$-kernel \cite{Shi23} (tensor kernel 
\cite{Sal1,Sal2}, moment function \cite{Der1}).}
\[
 \begin{array}{l}
  k_\pm \left( \lmiip {\bf x}, {\bf x}', \ldots, {\bf x}'''' \rmiip{m,{\bf g}} \right) \\
  \;= \lmip \bg{\featmapmkern} ({\bf x}), \bg{\featmapmkern} ({\bf x}'), \ldots, \bg{\featmapmkern} ({\bf x}') \rmip{m,(\pm\bg{\scriptstyle\gamma} \odot \bg{\scriptstyle\featmapmkern} ({\bf g}))_+} \\
 \end{array}
\]

\begin{table*}
\centering
\begin{tabular}{| l || l | l | l | l |}
\hline
 & $\;\;\;\;\;k$ & $\;\;\;\;\;\bar{k}^{{\;}^{{\;}^{{\;}^{\;}}}}\!\!\!\!\!$ & $\;\;\;\;\;a_i$ & $\!r\!\!$ \\
\hline
\hline
$\!$Linear         & $\!k(\xi) = \xi$                                                           & $\!\bar{k} (\xi) = \xi^{{\;}^{{\;}^{{\;}^{\;}}}}$                                & $\!\delta_{i,1}$ & $\!\!\infty\!\!$ \\
\hline
$\!$Erf            & $\!k(\xi) = {\rm erf}(\xi)$                                                & $\!\bar{k} (\xi) = {\rm erfi}(\xi)$                                              & $\!\left\{ \!\!\!\begin{array}{ll} \frac{2}{\sqrt{\pi}} \frac{(-1)^{\frac{i-1}{2}}}{i(\frac{i-1}{2})!} & \mbox{if } i \mbox{ odd} \\ 0 & \mbox{otherwise} \\ \end{array} \right.$ & $\!\!\infty\!\!$ \\
\hline
$\!$Tanh           & $\!k(\xi) = \tanh (\xi)$                                                   & $\!\bar{k} (\xi) = \tan (\xi)$                                                   & $\!\left\{ \!\!\!\begin{array}{ll} \frac{2}{\sqrt{\pi}} \frac{2^{k+1}(2^{k+1}-1)B_{k+1}}{(k+1)!} & \mbox{if } i \mbox{ odd} \\ 0 & \mbox{otherwise} \\ \end{array}\!\!\!\!\!\! \right.$ & $\!\!\frac{\pi}{2}\!\!$ \\
\hline
$\!$Logistic$\!\!$ & $\!k(\xi) = \frac{1}{1+\exp(-\xi)}$                                        & $\!\bar{k} (\xi) = \frac{1}{2} (1+\tan(\frac{\xi}{2}))\!\!$                      & $\!\left\{ \!\!\!\begin{array}{ll} \frac{1}{2} & \mbox{if } i = 0 \\ \frac{2}{\sqrt{\pi}} \frac{(2^{k+1}-1)B_{k+1}}{(k+1)!} & \mbox{if } i \mbox{ odd} \\ 0 & \mbox{otherwise} \\ \end{array} \right.\!\!\!$ & $\!\!\frac{\pi}{2}\!\!$ \\
\hline
\end{tabular}
\caption{Expansion series for {\krein} kernels.  In each case $K ({\bf x}, {\bf 
         x}') = k (\liip {\bf x}, {\bf x}' \riip{{\bf g}})$ is a {\krein} 
         kernel with associated kernel $\bar{K} ({\bf x},{\bf x}') = \bar{k} 
         (\liip {\bf x}, {\bf x}' \riip{{\bf g}})$ (the latter were obtained by 
         comparison of the adjusted Taylor series $\bar{k} (\xi) = \sum_i |a_i| 
         \xi^i$ with Taylor series of known functions - eg \cite{Gra1,Olv1,Abr2,Jan2}) on $\{ {\bf x} \in 
         \infset{X} | \liip {\bf x}, {\bf x} \riip{{\bf g}} \leq r \}$.  
         Taylor series expansions are $k (\xi) = \sum_i a_i \chi^i$, $\bar{k} 
         (\xi) = \sum_i |a_i| \chi^i$, so $k (\lmiip {\bf x}, {\bf x}', \ldots, 
         {\bf x}'''' \rmiip{m,{\bf g}}) = \lmiip \bg{\featmapmkern} 
         ({\bf x}), \bg{\featmapmkern} ({\bf x}'), \ldots, 
         \bg{\featmapmkern} ({\bf x}') \rmiip{m, 
         \bg{\scriptstyle\gamma} \odot 
         \bg{\scriptstyle\featmapmkern} ({\bf g})}$, $\bar{k} (\lmiip 
         {\bf x}, {\bf x}', \ldots, {\bf x}'''' \rmiip{m,{\bf g}}) = \lmiip 
         \bg{\featmapmkern} ({\bf x}), \bg{\featmapmkern} ({\bf 
         x}'), \ldots, \bg{\featmapmkern} ({\bf x}') \rmiip{m, 
         |\bg{\scriptstyle\gamma} \odot 
         \bg{\scriptstyle\featmapmkern} ({\bf g})|}$, where, in multi 
         index notation, ${\bg{\featmapmkern}} ({\bf x}) = [ 
         {{\featmapmkern}}_{\bf i} ({\bf x})]_{{\bf i} \in \infset{N}^{{\rm dim} ({\bf x})}}$, 
         ${\bg{\gamma}} = [ {{\gamma}}_{\bf i}]_{{\bf i} \in 
         \infset{N}^{{\rm dim} ({\bf x})}}$, where ${\featmapmkern}_{\bf i} ({\bf x}) = 
         \prod_j x_j^{i_j}$, ${\gamma}_{\bf i} = (\frac{{\rm sum} ({\bf 
         i})}{\prod_j i_j!}) a_{{\rm sum} ({\bf i})}$.}
\label{tab:actfuns}
\end{table*}

The map $\bg{\featmapmkern} : \infset{X} \to \frkks{}$ is an example 
of a feature map to a feature ({\krein}) space $\{ \frkks{}, \liip \cdot, 
\cdot \riip{\bg{\scriptstyle\gamma} \odot 
\bg{\scriptstyle\featmapmkern} ({\bf g})} \}$.  As for positive definite 
kernels, it is natural to think of {\krein} kernels being associated 
(non-uniquely) with feature maps in this way:
\begin{th_feature_krein}[Feature Maps]
 Let $\bg{\featmap} = \bg{\featmap}_+ \dsum\bg{\featmap}_- 
 : \infset{X} \to \frkks{} = \frkks{+} \ominus \frkks{-}$ (where 
 $\bg{\featmap}_\pm : \infset{X} \to \frkks{\pm}$) be a feature map from 
 input space $\infset{X}$ to {\krein} feature space $\frkks{}$, where the 
 Hilbert spaces $\frkhs{\pm}$ are imbued with inner-products $\lip \cdot,\cdot 
 \rip{\frkhs{\pm}}$ and $\frkks{}$ is imbued with indefinite inner product 
 $\liip {\bf v}, {\bf v}' \riip{\frkks{}} = \lip {\bf v}_+, {\bf v}'_+ 
 \rip{\frkhs{+}} - \lip {\bf v}_-, {\bf v}'_- \rip{\frkhs{-}}$.  Let 
 $\bar{\frkks{}} = \frkks{+} \oplus \frkks{-}$ be the associated Hilbert 
 feature space, imbued with inner product 
 $\lip {\bf v}, {\bf v}' \rip{\bar{\frkks{}}} = \lip {\bf v}_+, {\bf v}'_+ 
 \rip{\frkhs{+}} + \lip {\bf v}_-, {\bf v}'_- \rip{\frkhs{-}}$.  Then:
 \begin{itemize}
  \setlength{\parskip}{0pt}
  \setlength{\itemsep}{1pt plus 1pt}
  \item $K (x,x') = \liip \bg{\featmap} (x), \bg{\featmap} (x') 
        \riip{\frkks{}}$ is a {\krein} kernel for {\RKKS} 
        $\rkks{K} = \{ \liip {\bf v}, \bg{\featmap} (\cdot) 
        \riip{\frkks{}} | {\bf v} \in \frkks{} \}$.
  \item $\bar{K} (x,x') = \lip \bg{\featmap} (x), \bg{\featmap} (x') 
        \rip{\bar{\frkks{}}}$ is a kernel for associated {\RKHS} 
        $\bar{\rkks{}}{}_{\bar{K}} = \{ \lip {\bf v}, \bg{\featmap} (\cdot) 
        \rip{\bar{\frkks{}}} | {\bf v} \in \bar{\frkks{}} \}$.
  \item $\liip \lip {\bf v}, \bg{\featmap}(\cdot) \rip{{\frkks{}}}, \lip 
        {\bf v}', \bg{\featmap}(\cdot) \rip{{\frkks{}}} \riip{{\rkks{K}}} 
        = \liip {\bf v}, {\bf v}' \riip{{\frkks{}}}$. 
  \item $\lip \lip {\bf v}, \bg{\featmap}(\cdot) 
        \rip{\bar{\frkks{}}}, \lip {\bf v}', \bg{\featmap}(\cdot) 
        \rip{\bar{\frkks{}}} \rip{\bar{\rkks{}}_{\bar{K}}} = \lip {\bf v}, {\bf 
        v}' \rip{\bar{\frkks{}}}$. 
 \end{itemize}
 where the vectors ${\bf v}, {\bf v}'$ are called weight vectors.
\label{th:feature_krein}
\end{th_feature_krein}

Regularised risk minimization in {\RKKS} can be formulated in a number of 
ways \cite{Loo1,Ong3,Ogl2,Ogl1}.  In \cite{Ong3} a stabilised risk minimisation 
problem is given that combines empirical risk minimisation with a 
regularisation term of the form $\lambda \liip f,f \riip{\rkks{K}}$.  The 
result is non-convex, but nevertheless representor theory applies to all saddle 
points.  Alternatively, \cite{Ogl2} 
apply regularisation via the associated {\RKHS} norm - that is, a 
regularisation term of the form $\lambda \lip f,f \rip{\bar{\rkks{}} 
{}_{\bar{K}}}$.  Once again the problem is non-convex, but superior results are 
reported.  Following \cite{Ong3}, consider the following (equivalent) 
regularised risk minimisation problems:
\begin{equation}
 \begin{array}{rcllll}
  \!\!\!\!\!\!\!\!\!{\bf v}^\star &\!\!\!\!\!\!=\!\!\! &\!\!\!\mathop{\min}\limits_{{\bf v} \in \frkks{}} &\!\!\!\!\!\!{\sum}_i \loss \left( \liip {\bf v}, \bg{\featmap} (x_i) \riip{\frkks{}} - y_i \right) + \lambda h \left( \liip {\bf v}, {\bf v} \riip{{\frkks{}}} \right) \!\!\!\!\!\!\!\!\!\\ & & & \mbox{(weight-centric form)} \\ \\
  \!\!\!\!\!\!\!\!\!f^\star       &\!\!\!\!\!\!=\!\!\! &\!\!\!\mathop{\min}\limits_{f \in \rkks{K}}       &\!\!\!\!\!\!{\sum}_i \loss \left( o (x_i) - y_i \right) + \lambda h \left( \liip f,f \riip{\rkks{K}} \right) \!\!\!\!\!\!\!\!\!\\ & & & \mbox{(function-centric form)} \\
 \end{array}
 \label{eq:rkks_reg_minprob}
\end{equation}
where $\{ (x_i,y_i) : i \in \infset{N}_n \}$ is some training set, $\loss$ is a 
(differentiable) loss function, and $h$ is differentiable.  As per \cite{Ong3}, 
it is not difficult to see that this has a solution of the form $f^\star 
(\cdot) = \sum_i \alpha_i K (\cdot,x_i)$ (or, equivalently in weight-centric 
notation, ${\bf v}^\star = \sum_i \alpha_i \bg{\featmap} (x_i)$), where 
$\bg{\alpha} \in \infset{R}^n$.  Note that, while 
(\ref{eq:rkks_reg_minprob}) appears directly analogous to a typical regularised 
risk minimisation problem in reproducing kernel Hilbert space, the 
non-convexity of this form makes finding ${\mbox{\boldmath $\alpha$}}$ somewhat 
complicated \cite{Ong3}, which may explain why it does not appear to have been 
widely adopted despite promising performance.

\section{Supplementary: Non-Entire Activation Functions} \label{sec:approxit}

While the ``entire function'' requirement on the activation functions 
$\sigma_{q}$ is necessary, we note that more general concave functions 
$\sigma_{q}$ can be approximated to arbitrary precision using an entire, 
concave surrogate.  For example, if $\sigma_{q}$ is continuous then it may 
always be approximated to arbitrary precision by a finite sum 
$\tilde{\sigma}_{q} (\cdot) = \sum_i \beta_i \kappa (\cdot,\xi_i)$, where 
$\beta_i,\xi_i \in \infset{R}$ and $\kappa$ is an entire, concave, integrated 
universal kernel \cite{Mic1} (for example, $\kappa (\xi) = \erf (\xi)$).  In 
this way we may construct arbitrarily close entire approximations to e.g. the 
tanh activation function.  Thus, though our analysis is restricted to entire 
activation functions, this should not be seen as a serious limiting factor.

\subsection{A Note on the ReLU Activation Function} \label{sec:relunote}

The ReLU (Rectified Linear Unit) activation $\sigma_+ (\xi) = (\xi)_+$ function 
is popular in deep networks, so it is worth considering it in more detail.  It 
is not entire, but can be approximated to arbitrary accuracy by $\sigma_{c+} 
(\xi) = \lim_{c \to 0_+} \frac{1}{2} \xi (1+\erf(\frac{1}{c}\xi))$, which is an 
entire function.  When discussing ReLU networks we implicitly mean the limit of 
some sequence $\sigma_{c_0+}, \sigma_{c_1+}, \ldots$, where $c_0 \geq c_1 \geq 
\ldots \to 0$.

\section{Supplementary: Details of Proofs}

In this section we present the full proofs for the theorems presented in the 
body of the paper.

\subsection{Preliminaries II: Indefinite SVMs}


{\bf Theorem \ref{th:indefrep} (Representor Theory) } 
{\it
 Any solution ${\bf v}^\star$ to (\ref{eq:trainSVM}) can be represented as 
 ${\bf v}^\star = \sum_i \alpha_i {\bg{\featmapmkern}} ({\bf x}_i)$, 
 where ${\bg{\alpha}} \in \infset{R}^N$.  Defining $K ({\bf x}, 
 {\bf x}') = \liip {\bg{\featmapmkern}} ({\bf x}), 
 {\bg{\featmapmkern}} ({\bf x}') \riip{{\bf g}}$, the optimal 
 $f^\star : \infset{R}^D \to \infset{R}$ is $f^\star ({\bf 
 x}) = \sum_i \alpha_i^{\star} K ({\bf x}, {\bf x}_i)$.
}
\begin{proof}
Applying first order stationarity conditions, denoting the derivative of 
$\loss$ as $\loss'$, we have:
\[
 \begin{array}{r}
  \frac{\partial \loss}{\partial v_k} \!= \!0 \!= \!\frac{1}{N} \!\sum_i \!\loss' \!\left( y_i, \!\liip {\bf v}, {\bg{\featmapmkern}} ({\bf x}_i) \riip{{\bf g}} \right) g_k \featmapmkern_k ({\bf x}_i) \!+\! 2 \lambda g_k v_k \\
 \end{array}
\]
and so ${\bf v} = \sum_i \alpha_i {\bg{\featmapmkern}} ({\bf x}_i)$ for 
${\bg{\alpha}} \in \infset{R}^N$, where $\alpha_i = \frac{1}{2 \lambda N} 
\sum_i \loss' (y_i, \!\liip {\bf v}, {\bg{\featmapmkern}} ({\bf x}_i) 
\riip{{\bf g}})$.  Substituting into (\ref{eq:wfinalformb}) we have $f ({\bf 
x}) = \sum_i \alpha_i K ({\bf x}, {\bf x}_i)$ for $K$ defined.  
\end{proof}

\subsection{Flat Representations of Deep Networks}


{\bf Lemma \ref{th:pushforward} (Extended Version):} 
{\it
 Let $\sigma$ be an entire function with Taylor expansion $\sigma(\xi) = \sum_i 
 a_i \xi^i$, and let $\lmiip \cdot, \cdot, \ldots \rmiip{m, 
 {\bg{\scriptstyle\mu}}}$ be an $m$-indefinite-inner-product defined by 
 metric ${\bg{\mu}} \in \infset{R}^n$ (section \ref{sec:sipsiip}).  Then:
 \[
  \begin{array}{rl}
   \!\!\!\!\!\!\sigma \!\left( \lmiip {\bf x}, \ldots, {\bf x}'''' \rmiip{m,{\bg{\scriptstyle\mu}}} \right) 
   \!=\! \lmiip {\bg{\featmap}} \!\left( {\bf x} \right), \ldots, {\bg{\featmap}} \!\left( {\bf x}'''' \right) \rmiip{m,{\bg{\scriptstyle\gamma}} \odot {\bg{\scriptstyle\featmap}} ({\bg{\scriptstyle\mu}})}\!\!\!\!\!\! \\
  \end{array}
  \;\; (\ref{eq:define_feature})
 \]
 where $\bg{\featmap} : \infset{R}^n \to \frkks{}$ is a feature map and 
 $\bg{\gamma} \in \frkks{}$, both independent of $m$ and ${\bg{\mu}}$.  
 Using multi-index notation, ${\bg{\featmap}} ({\bf x}) = [ 
 {{\featmap}}_{\bf i} ({\bf x})]_{{\bf i} \in \infset{N}^n}$ and 
 ${\bg{\gamma}} = [ {{\gamma}}_{\bf i}]_{{\bf i} \in \infset{N}^n}$, 
 where:
 \[
  \begin{array}{l}
   {\featmap}_{\bf i} \left( {\bf x} \right) = \prod_j x_j^{i_j}, \;\;\;\; {\gamma}_{\bf i} = \left( \frac{{\rm sum} \left( {\bf i} \right)}{\prod_j i_j!} \right) a_{{\rm sum} \left( {\bf i} \right)}
  \end{array}
  \;\;\;\;\;\; (\ref{eq:featmapform})
 \]
 Moreover, $\forall {\bf t} \in \infset{R}^D$, $\forall {\bf u}_{(0)}, \ldots, 
 {\bf u}_{(n-1)}, {\bf v}, {\bf v}' \ldots, {\bg{\mu}} 
 \in \infset{R}^p$:
 \begin{equation}
  \!\!\!\begin{array}{l}
   \sigma \left( \sum_j t_{j} \lmiip {\bf u}_{(j)}, {\bf v}, {\bf v}', \ldots \rmiip{m,{\bg{\scriptstyle\mu}}} \right) 
   = \ldots \\
   \lmiip {\bg{\featmap}} \left( {\bf t}_\bullet \right), {\bg{\featmap}} \left( {\bf u}_\bullet \right), {\bg{\featmap}} \left( {\bf v}_\bullet \right), {\bg{\featmap}} \left( {\bf v}'_\bullet \right), \ldots \rmiip{m+1,{\bg{\scriptstyle\gamma}} \odot {\bg{\scriptstyle\featmap}} ({\bg{\scriptstyle\mu}}_\bullet)\!\!\!\!\!\!\!\!} \\
  \end{array}
  \label{eq:propeqn}
 \end{equation}
 where ${\bf t}_\bullet = {\bf t} \otimes {\bf 1}_p$ and ${\bf u}_\bullet = 
 [ {\bf u}_{(0)}^\tsp \; {\bf u}_{(1)}^\tsp \ldots \; ]^\tsp$, ${\bf v}_\bullet 
 = {\bf 1}_n \otimes {\bf v}, {\bf v}'_\bullet = {\bf 1}_n \otimes {\bf v}', 
 \ldots$, ${\bg{\mu}}_\bullet = {\bf 1}_n \otimes {\bg{\mu}}$.  
}
\begin{proof}
Equation (\ref{eq:define_feature}) follows directly by substituting the 
$m$-indefinite-inner-product into the Taylor expansion of $\sigma$ and applying 
the multinomial expansion.  For (\ref{eq:propeqn}) we expand, noting that:
\[
 \begin{array}{l}
  t_{\bullet i} = t_{\left\lfloor \frac{i}{p} \right\rfloor}, 
  u_{\bullet i} = u_{\left(\left\lfloor \frac{i}{p} \right\rfloor\right), i-p\left\lfloor \frac{i}{p} \right\rfloor}, 
  v_{\bullet i} = v_{i-p\left\lfloor \frac{i}{p} \right\rfloor}, \\
  v'_{\bullet i} = v'_{i-p\left\lfloor \frac{i}{p} \right\rfloor}, \ldots, 
  \mu_{\bullet i} = \mu_{i-p\left\lfloor \frac{i}{p} \right\rfloor} 
 \end{array}
\] 
where $\lfloor \cdot \rfloor$ is floor:
\[
 \begin{array}{l}
  \sum_{j \in \infset{N}_n} t_{j} \lmiip {\bf u}_{(j)}, {\bf v}, {\bf v}', \ldots \rmiip{m,{\bg{\scriptstyle\mu}}} \\
   \;\;\;= \sum_{j \in \infset{N}_n, k \in \infset{N}_p} t_{j} u_{(j),k} v_{k} v'_{k} \ldots \mu_{k} \\
   \;\;\;= \sum_{i \in \infset{N}_{np}} t_{\left\lfloor \frac{i}{p} \right\rfloor } u_{\left(\left\lfloor \frac{i}{p} \right\rfloor\right), i-p\left\lfloor \frac{i}{p} \right\rfloor} v_{i-p\left\lfloor \frac{i}{p} \right\rfloor} v'_{i-p\left\lfloor \frac{i}{p} \right\rfloor} \ldots  \mu_{i-p\left\lfloor \frac{i}{p} \right\rfloor} \\
   \;\;\;= \sum_{i \in \infset{N}_{np}} t_{\bullet i} u_{\bullet i} v_{\bullet i} v'_{\bullet i} \ldots  \mu_{\bullet i} \\
   \;\;\;= \lmiip {\bf t}_\bullet, {\bf u}_\bullet, {\bf v}_\bullet, {\bf v}'_\bullet, \ldots \rmiip{m+1,{\bg{\scriptstyle\mu}}_\bullet} \\
 \end{array}
\]
Substituting and apply (\ref{eq:define_feature}):
\[
 \begin{array}{l}
  \sigma \left( \sum_{j \in \infset{N}_n} t_{j} \lmiip {\bf u}_{j}, {\bf v}, {\bf v}', \ldots \rmiip{m,{\bg{\scriptstyle\mu}}} \right) \\
   \;\;\;= \lmiip {\bg{\featmapmkern}} \left( {\bf t}_\bullet \right), {\bg{\featmap}} \left( {\bf u}_\bullet \right), {\bg{\featmap}} \left( {\bf v}_\bullet \right), {\bg{\featmap}} \left( {\bf v}'_\bullet \right), \ldots \rmiip{m+1,{\bg{\scriptstyle\gamma}} \odot {\bg{\scriptstyle\featmap}} \left( {\bg{\scriptstyle\mu}}_\bullet \right)} \\
 \end{array}
\]
which completes the proof.
\end{proof}

$\;$

\begin{th_collapse}
 The deep network (\ref{eq:deepnetform}) has equivalent form 
 (\ref{eq:semideepflat}), where ${\bg{\featmapmkern}}_{\NN} ({\bf x}) = 
 {\bf 1}_{H_{d-1}} \otimes {\bg{\featmap}} ({\bf 1}_{H_{d-2}} \otimes 
 {\bg{\featmap}} (\ldots {\bf 1}_{H_0} \otimes {\bg{\featmap}} 
 ({\bf x})))$ is the feature map and the weight vectors ${\bf v}_{[0]}$, ${\bf 
 v}_{[1]}$, $\ldots$ in (\ref{eq:semideepflat}) correspond to the weight 
 matrices in (\ref{eq:deepnetform}) via:
 \begin{equation}
  {\!\!\!\!\!\!\!\!\!\small{
  \begin{array}{rcl}
   {\bf v}_{[q]} &\!\!\!\!\!=&\!\!\!\!\! {\bg{\featmap}}_{[q+]} \left( {\bf W}_{[q]}^\tsp \right) \\
   {\bf g}_{\NN} &\!\!\!\!\!=&\!\!\!\!\! {\bg{\gamma}}_{[d-1]} \!\odot\! {\bg{\featmap}} \left( {\bf 1}_{H_{d-2}} \!\otimes\! \left( \ldots {\bf 1}_{H_0} \!\otimes\! \left( {\bg{\gamma}}_{[0]} \!\odot\! {\bg{\featmap}} \left( {\bf 1}_D \right) \right) \right) \right) \\
  \end{array}
  }\!\!\!\!\!\!}
  \label{eq:wformx}
 \end{equation}
 (which depend only on the deep network structure) where:
 \[
  \small{
  \begin{array}{l}
  \begin{array}{rl}
   \!\!\!{\bg{\featmap}}_{[q+]} \!\left( {\bf W}^\tsp \right) 
   &\!\!\!\!\!\!=\! {\bf 1}_{\!H_{d-1}} \!\!\otimes\! {\bg{\featmap}} \!\!\left(\!\! \ldots {\bf 1}_{\!H_{q+1}} \!\!\otimes\! {\bg{\featmap}} \!\!\left( \!\left[\!\!\! \begin{array}{c} {\bg{\featmap}} \left( {\bf W}_{\!0,:} \!\otimes \!{\bf e}_{[q]} \right) \\ {\bg{\featmap}} \left( {\bf W}_{\!1,:} \!\otimes \!{\bf e}_{[q]} \right) \\ \vdots \\ \end{array} \!\!\!\right] \!\right) \!\!\right) \\
  \end{array} \\
  \begin{array}{l}
   {\bf e}_{[q]} \!=\! {\bf 1}_{H_{q-1}} \otimes {\bg{\featmap}} \left( \ldots {\bf 1}_{H_0} \otimes {\bg{\featmap}} \left( {\bf 1}_D \right) \right)^{{\;}^{{\;}^{{\;}^{{\;}^{\;}}}}}\!\!\!\!\!\!\!\! \\
  \end{array}
  \end{array}
  }
 \] 
 (here ${\bf W}_{i,:}$ is row $i$ of matrix ${\bf W}$ (Matlab style notation)). 
 \label{th:collapse}
\end{th_collapse}
\begin{proof}
Let $m_0 = 1$ and $m_q = {\rm dim} ( {\bf x}_{[q-1]} )$ (see below).  
%
Let ${\bf w}_{[q],(i)} = {\bf W}_{[q] i,:}$, and let ${\bf o}_{[q]} ({\bf x})$ 
denote the output of layer $q$.  We proceed as follows:
\begin{description}
 \item[Layer 0:] As per figure \ref{fig:deepstandard} and equation 
       (\ref{eq:define_feature}), the output of layer $0$ is:
       \begin{equation}
        {\small{
        \begin{array}{rl}
         {\bf o}_{[0]} \left( {\bf x} \right)
         &\!\!\!= \left[ \begin{array}{c} 
                  \sigma_{0} ( \lmiip {\bf w}_{[0],(0)}, {\bf x} \rmiip{2,{\bf 1}} ) \\
                  \sigma_{0} ( \lmiip {\bf w}_{[0],(1)}, {\bf x} \rmiip{2,{\bf 1}} ) \\
                  \vdots \\
                  \end{array} \right] \\ 
         &\!\!\!= \left[ \begin{array}{c}
                  \lmiip {\bf w}_{[0,1],(0)}, {\bf x}_{[1]\bullet} \rmiip{2,{\bf g}_{[1]\bullet}} \\
                  \lmiip {\bf w}_{[0,1],(1)}, {\bf x}_{[1]\bullet} \rmiip{2,{\bf g}_{[1]\bullet}} \\
                  \vdots \\
                  \end{array} \right] 
        \end{array}
        }}
        \label{eq:push0}
       \end{equation}
       where ${\bf x}$ has been propogated through layer $0$ to obtain ${\bf 
       x}_{[1]}$, and likewise ${\bf W}_{[0]}$ has been proprogated through 
       layer $0$ to obtain ${\bf w}_{[0,1]}$.  Specifically (${\bf 1}_{m_0} = 
       [1]$):
       \[
        \!\!\!\!\!\!\!\!\!\!\!\!\!\!\!\!\!\!{\small{
        \begin{array}{l}
         \left.
         \begin{array}{rl}
          {\bf x}_{[1]\bullet} &\!\!\!\!= {\bg{\featmap}} \left( {\bf x} \right) \\
          {\bf g}_{[1]\bullet} &\!\!\!\!= {\bg{\gamma}}_{[0]} \odot {\bg{\featmap}} \left( {\bf 1}_D \right) \\
          {\bf w}_{[0,1],(i)}  &\!\!\!\!= {\bg{\featmap}} \left( {\bf w}_{[0],(i)} \otimes {\bf 1}_{m_0} \right) \!\!\!\! \\
         \end{array}
         \right|\!\!\!
         \begin{array}{rl}
          {\bf x}_{[1]}   &\!\!\!\!= {\bf 1}_{H_0} \otimes {\bf x}_{[1]\bullet} \\
          {\bf g}_{[1]}   &\!\!\!\!= {\bf 1}_{H_0} \otimes {\bf g}_{[1]\bullet} \\
          {\bf w}_{[0,1]} &\!\!\!\!= [ {\bf w}_{[0,1],(0)}^\tsp \; {\bf w}_{[0,1],(1)}^\tsp \; \ldots ]^\tsp \\
         \end{array}
        \end{array}
        }}
       \]
 \item[Layer 1:] As per figure \ref{fig:deepstandard} and equation 
       (\ref{eq:propeqn}), the output of layer $1$ is (where ${\bf 1}_{m_1} = 
       {\bf 1}_{H_0} \otimes \bg{\featmap} \left( {\bf 1}_D \right)$):
       \[
        \!\!\!\!\!{\small{
        \begin{array}{rl}
         {\bf o}_{[1]} \left( {\bf x} \right)
         &\!\!\!= \left[ \begin{array}{c} 
                  \sigma_{1} ( \sum_i w_{[1],(0) i} \lmiip {\bf w}_{[0,1],(i)}, {\bf x}_{[1]\bullet} \rmiip{2,{\bf g}_{[1]\bullet}} ) \\
                  \sigma_{1} ( \sum_i w_{[1],(1) i} \lmiip {\bf w}_{[0,1],(i)}, {\bf x}_{[1]\bullet} \rmiip{2,{\bf g}_{[1]\bullet}} ) \\
                  \vdots \\
                  \end{array} \right] \\
         &\!\!\!= \left[ \begin{array}{c} 
                  \sigma_{1} ( \lmiip {\bf w}_{[1],(0)} \!\otimes\! {\bf 1}_{m_1}, {\bf w}_{[0,1]}, {\bf x}_{[1]} \rmiip{3,{\bf g}_{[1]}} ) \\
                  \sigma_{1} ( \lmiip {\bf w}_{[1],(1)} \!\otimes\! {\bf 1}_{m_1}, {\bf w}_{[0,1]}, {\bf x}_{[1]} \rmiip{3,{\bf g}_{[1]}} ) \\
                  \vdots \\
                  \end{array} \right] \\
         &\!\!\!= \left[ \begin{array}{c}
                  \lmiip {\bf w}_{[1,2],(0)}, {\bf w}_{[0,2]\bullet}, {\bf x}_{[2]\bullet} \rmiip{3,{\bf g}_{[2]\bullet}} \\
                  \lmiip {\bf w}_{[1,2],(1)}, {\bf w}_{[0,2]\bullet}, {\bf x}_{[2]\bullet} \rmiip{3,{\bf g}_{[2]\bullet}} \\
                  \vdots \\
                  \end{array} \right] 
        \end{array}
        }}
       \]
       where ${\bf x}_{[1]}$ has been propogated through layer $1$ to obtain 
       ${\bf x}_{[2]}$, and likewise for the weights (${\bf w}_{[i,j]}$ is the 
       result of propogating weights ${\bf W}_{[i]}$ through layers $i,i+1, 
       \ldots,j-1$).  So:
       \[
        \!\!\!\!\!\!\!\!\!\!\!\!\!\!\!\!\!\!{\small{
        \begin{array}{l}
         \left.
         \begin{array}{rl}
          {\bf x}_{[2]\bullet}   &\!\!\!= {\bg{\featmap}} \left( {\bf x}_{[1]} \right) \\
          {\bf g}_{[2]\bullet}   &\!\!\!= {\bg{\gamma}}_{[1]} \odot {\bg{\featmap}} \left( {\bf g}_{[1]} \right) \\
          {\bf w}_{[0,2]\bullet} &\!\!\!= {\bg{\featmap}} \left( {\bf w}_{[0,1]} \right) \\
          {\bf w}_{[1,2],(i)}    &\!\!\!= {\bg{\featmap}} \left( {\bf w}_{[1],(i)} \otimes {\bf 1}_{m_1} \right)\!\!\! \\
         \end{array}
         \right|\!\!\!
         \begin{array}{rl}
          {\bf x}_{[2]}   &\!\!\!\!= {\bf 1}_{H_1} \otimes {\bf x}_{[2]  \bullet} \\
          {\bf g}_{[2]}   &\!\!\!\!= {\bf 1}_{H_1} \otimes {\bf g}_{[2]  \bullet} \\
          {\bf w}_{[0,2]} &\!\!\!\!= {\bf 1}_{H_1} \otimes {\bf w}_{[0,2]\bullet} \\
          {\bf w}_{[1,2]} &\!\!\!\!= [ {\bf w}_{[1,2],(0)}^\tsp \; {\bf w}_{[1,2],(1)}^\tsp \; \ldots ]^\tsp \\
         \end{array}
        \end{array}
        }}
       \] \\ $\ldots$
 \item[Layer q:] Repeating the same approach, at layer $q$ (where ${\bf 1}_{m_q} 
       = {\bf 1}_{H_{q-1}} \otimes \bg{\featmap} \left( {\bf 1}_{H_{q-2}} 
       \otimes \bg{\featmap} \left( \ldots {\bf 1}_{H_0} 
       \bg{\featmap} \left( {\bf 1}_D \right) \right) \right)$):
       \[
        \!\!\!\!\!\!\!\!\!\!\!\!\!\!\!\!\!\!{\small{
        \begin{array}{l}
         {\bf o}_{[q]} \left( {\bf x} \right) \\
         \;= \left[ \begin{array}{c} 
                  \sigma_{q} ( \sum_i w_{[q],(0)i} \lmiip {\bf w}_{[q-1,q]\bullet}, \ldots, {\bf w}_{[0,q]\bullet}, {\bf x}_{[q]\bullet} \rmiip{q+2,{\bf g}_{[q]\bullet}} ) \\
                  \sigma_{q} ( \sum_i w_{[q],(1)i} \lmiip {\bf w}_{[q-1,q]\bullet}, \ldots, {\bf w}_{[0,q]\bullet}, {\bf x}_{[q]\bullet} \rmiip{q+2,{\bf g}_{[q]\bullet}} ) \\
                  \vdots \\
                  \end{array} \right] \\
         \;= \left[ \begin{array}{c} 
                  \sigma_{q} ( \lmiip {\bf w}_{[q],(0)} \!\otimes\! {\bf 1}_{m_q}, {\bf w}_{[q-1,q]}, \ldots, {\bf w}_{[0,q]}, {\bf x}_{[q]} \rmiip{q+2,{\bf g}_{[q]}} ) \\
                  \sigma_{q} ( \lmiip {\bf w}_{[q],(1)} \!\otimes\! {\bf 1}_{m_q}, {\bf w}_{[q-1,q]}, \ldots, {\bf w}_{[0,q]}, {\bf x}_{[q]} \rmiip{q+2,{\bf g}_{[q]}} ) \\
                  \vdots \\
                  \end{array} \right] \\
         \;= \left[ \!\!\!\begin{array}{c}
                  \lmiip {\bf w}_{[q,q+1],(0)}, {\bf w}_{[q-1,q+1]\bullet}, \ldots, {\bf w}_{[0,q+1]\bullet}, {\bf x}_{[q+1]\bullet} \rmiip{q+2,{\bf g}_{[q+1]\bullet}} \\
                  \lmiip {\bf w}_{[q,q+1],(1)}, {\bf w}_{[q-1,q+1]\bullet}, \ldots, {\bf w}_{[0,q+1]\bullet}, {\bf x}_{[q+1]\bullet} \rmiip{q+2,{\bf g}_{[q+1]\bullet}} \\
                  \vdots \\
                  \end{array}\!\!\! \right] 
        \end{array}
        }}
       \]
       where propogation through layer $q$ gives:
       \begin{equation}
        \!\!\!\!\!\!\!\!\!\!\!\!\!\!\!\!\!\!{\small{
        \begin{array}{l}
         \left.
         \begin{array}{rl}
          {\bf x}_{[q+1]\bullet}     &\!\!\!= {\bg{\featmap}} \left( {\bf x}_{[q]} \right) \\
          {\bf g}_{[q+1]\bullet}     &\!\!\!= {\bg{\gamma}}_{[q]} \odot {\bg{\featmap}} \left( {\bf g}_{[q]} \right) \\
          {\bf w}_{[0,q+1]\bullet}   &\!\!\!= {\bg{\featmap}} \left( {\bf w}_{[0,q]} \right) \\
          {\bf w}_{[1,q+1]\bullet}   &\!\!\!= {\bg{\featmap}} \left( {\bf w}_{[1,q]} \right) \\
          \ldots \\
          {\bf w}_{[q-1,q+1]\bullet} &\!\!\!= {\bg{\featmap}} \left( {\bf w}_{[q-1,q]} \right) \\
          {\bf w}_{[q,q+1],(i)}      &\!\!\!= {\bg{\featmap}} \left( {\bf w}_{[q],(i)} \otimes {\bf 1}_{m_{q}} \right) \\
         \end{array} 
         \!\!\!\right|\!\!\! 
         \begin{array}{rl}
          {\bf x}_{[q+1]}     &\!\!\!\!= {\bf 1}_{H_q} \otimes {\bf x}_{[q+1]    \bullet} \\
          {\bf g}_{[q+1]}     &\!\!\!\!= {\bf 1}_{H_q} \otimes {\bf g}_{[q+1]    \bullet} \\
          {\bf w}_{[0,q+1]}   &\!\!\!\!= {\bf 1}_{H_q} \otimes {\bf w}_{[0,q+1]  \bullet} \\
          {\bf w}_{[1,q+1]}   &\!\!\!\!= {\bf 1}_{H_q} \otimes {\bf w}_{[1,q+1]  \bullet} \\
          \ldots \\
          {\bf w}_{[q-1,q+1]} &\!\!\!\!= {\bf 1}_{H_q} \otimes {\bf w}_{[q-1,q+1]\bullet} \\
          {\bf w}_{[q,q+1]}   &\!\!\!\!= [ {\bf w}_{[q,q+1],(0)}^\tsp \; \ldots \\
                              & \ldots {\bf w}_{[q,q+1],(1)}^\tsp \; \ldots ]^\tsp \\
         \end{array}
        \end{array}
        }}
        \label{eq:propitq}
       \end{equation} $\ldots$
 \item[Output layer: ] Propogation through the output layer $d-1$ follows the 
       same formula, noting that $n_{d-1} = 1$ and so ${\bf x}_{[d]} = {\bf 
       x}_{[d]\bullet}$, ${\bf w}_{[d-1,d]} = {\bf w}_{[d-1,d],0\bullet}$ etc.  
       Hence:
       \[
        {\small{
        \begin{array}{l}
         o \left( {\bf x} \right) 
         = o_{[d-1]} \left( {\bf x} \right) \\
         \;= \lmiip {\bf w}_{[d-1,d],0}, {\bf w}_{[d-2,d]}, \ldots, {\bf w}_{[1,d]}, {\bf w}_{[0,d]}, {\bf x}_{[d]} \rmiip{d+1,{\bf g}_{[d]}}
        \end{array}
        }}
       \]
       where (\ref{eq:propitq}) applies with $q = d-1$.
\end{description}

To simplify our notation we define:
\begin{equation}
 {\!\!\!\small{
  \begin{array}{l}
   {\bg{\featmapmkern}}_{\NN} \left( {\bf x} \right) = {\bf 1}_{H_{d-1}} \otimes {\bg{\featmap}} \left( {\bf 1}_{H_{d-2}} \otimes {\bg{\featmap}} \left( \ldots {\bf 1}_{H_0} \otimes {\bg{\featmap}} \left( {\bf x} \right) \right) \right) \\
 \begin{array}{rl}
  \!\!\!{\bg{\featmap}}_{[q+]} \!\!\left( \!{\bf W}^\tsp \!\right) 
  &\!\!\!\!\!\!=\! {\bf 1}_{\!H_{d-1}} \!\!\!\otimes\! {\bg{\featmap}} \!\!\left(\!\! \ldots {\bf 1}_{\!H_{q+1}} \!\!\!\otimes\! {\bg{\featmap}} \!\!\left( \!\left[\!\!\! \begin{array}{c} {\bg{\featmap}} \left( {\bf W}_{\!0,:} \!\otimes \!{\bf 1}_{m_q} \right) \\ {\bg{\featmap}} \left( {\bf W}_{\!1,:} \!\otimes \!{\bf 1}_{m_q} \right) \\ \vdots \\ \end{array} \!\!\!\right] \!\right) \!\!\right) \\
 \end{array} \\
 \begin{array}{l}
 \end{array}
 \end{array}
 }\!\!\!\!\!\!\!\!}
\end{equation}
%
%
%
%
Using this 
notation, it is not difficult to see that $\forall q \in \infset{N}_d$:
\begin{equation}
 \begin{array}{rl}
  {\bf w}_{[q,d]} 
  &\!\!\!= {\bg{\featmap}}_{[q+]} \left( {\bf W}_{[q]}^\tsp \right) \\
  {\bf x}_{[d]} 
  &\!\!\!= {\bg{\featmapmkern}}_{\NN} \left( {\bf x} \right) \\
 \end{array}
 \label{eq:wmaps}
\end{equation}
and hence, defining:
\[
 \begin{array}{l}
  {\bf v}_{\NN} = {\bf w}_{[d-1,d]} \odot \ldots \odot {\bf w}_{[1,d]} \odot {\bf w}_{[0,d]} \\
  {\bf g}_{\NN} = {\bf g}_{[d]} = {\bg{\gamma}}_{[d-1]} \odot {\bg{\featmap}} ( {\bf 1}_{H_{d-2}} \otimes ( {\bg{\gamma}}_{[d-2]} \odot {\bg{\featmap}} ( \ldots \\
  \;\;\;\;\;\;\;\;\;\ldots {\bf 1}_{H_0} \otimes ( {\bg{\gamma}}_{[0]} \odot {\bg{\featmap}} ( {\bf 1}_D ) ) \ldots ) ) ) \\
 \end{array}
\]
the overall network may be written in the simple form:
\begin{equation}
 \begin{array}{rl}
  f \left( {\bf x} \right) 
  &\!\!\!= \liip {\bf v}_{\NN}, {\bg{\featmapmkern}}_{\NN} \left( {\bf x} \right) \riip{{\bf g}_{\NN}} \\
 \end{array}
 \label{eq:trained_machine}
\end{equation}

Finally, using the form of ${\bg{\featmapmkern}}_{\NN}$ (a monomial map with 
terms of the form $x_j^{i_j}$) we have that:
\[
 \begin{array}{l}
  {\bf 1}_{m_i} = {\bg{\featmap}} \left( \ldots {\bf 1}_{H_1} \otimes {\bg{\featmap}} \left( {\bf 1}_{H_0} \otimes {\bg{\featmap}} \left( {\bf 1}_D \right) \right) \right)
 \end{array}
\]
and also ${\bg{\featmap}} ({\bf a} \odot {\bf b}) = 
{\bg{\featmap}} ({\bf a}) \odot {\bg{\featmap}} ({\bf 
a})$ 
and 
${\bg{\featmap}} ({\bf a} \otimes {\bf b}) = {\bg{\featmap}} 
({\bf a}) \otimes {\bg{\featmap}} \left( {\bf b} \right)$ (recalling 
that ${\bg{\featmap}}$ is purely polynomial, and hence $({\bf a} \otimes 
{\bf b})^{\odot n} = {\bf a}^{\odot n} \otimes {\bf b}^{\odot n}$).  It follows 
that:
\[
 \small{
 \begin{array}{l}
   {\bg{\featmap}}_{[i+]} \left( {\bf W} \right) 
   = {\bf 1}_{H_j} \otimes {\bg{\featmap}} ( \ldots {\bf 1}_{H_{i+1}} \otimes {\bg{\featmap}} ( \left[ \begin{array}{c} {\bg{\featmap}} ( {\bf W}_{0,:} \otimes {\bf 1}_{m_i} ) \\ {\bg{\featmap}} ( {\bf W}_{1,:} \otimes {\bf 1}_{m_i} ) \\ \vdots \\ \end{array} \right] ) ) \\
   \;= {\bf 1}_{H_j} \otimes {\bg{\featmap}} ( \ldots {\bf 1}_{H_{i+1}} \otimes {\bg{\featmap}} ( \left[ \begin{array}{c} {\bg{\featmap}} ( {\bf W}_{0,:} \otimes {\bf e}_{[i]} ) \\ {\bg{\featmap}} ( {\bf W}_{1,:} \otimes {\bf e}_{[i]} ) \\ \vdots \\ \end{array} \right] ) ) \\
 \end{array}
 }
\]
which completes the proof.
\end{proof}

\subsection{Regularisation Properties}

{\bf Theorem \ref{cor:collapse} }
{\it
 Recalling that $\sigma_{q} (\xi) = \sum_i a_{[q]i} \xi^i$, for all $q \in 
 \infset{N}_d$, and $\bar{\sigma}_{q} (\xi) = \sum_i |a_{[q]i}| \xi^i$.  
 Defining $\liip {\bf v}_{[q]}, {\bf v}_{[q]} \riip{{\bf g}_{\NN}} = p_q({\bf W}_{[q]})$ 
 and $\lip {\bf v}_{[q]}, {\bf v}_{[q]} \rip{|{\bf g}_{\NN}|} = \bar{p}_q ({\bf 
 W}_{[q]})$, where:
 \[
  {\small\!\!\!\!\!\!\!\!\!{
  \begin{array}{l}
   p_q \left( {\bf W}_{[q]} \right)
   \!=\! \sigma_{d-1} \Big( H_{d-2} \sigma_{d-2} \Big( H_{d-3} \sigma_{d-3} \Big( \ldots \\
   H_{q+1} \sigma_{q+1} \Big( \sum_{i_{q}} \sigma_{q} \Big( \left\| {\bf W}_{[q]i_{q},:} \right\|_2^2 \sigma_{q-1} \Big( \ldots 
   \!H_0 \sigma_{0} \!\Big( D \Big) \!\ldots \!\Big) \\ 
  \end{array}
  }\!\!\!\!\!\!}
  \;\;\mbox{ (\ref{eq:vqiip})}
 \]
 \[
  {\small\!\!\!\!\!\!\!\!\!{
  \begin{array}{l}
   \bar{p}_q \left( {\bf W}_{[q]} \right)
   \!=\! \bar{\sigma}_{d-1} \Big( H_{d-2} \bar{\sigma}_{d-2} \Big( H_{d-3} \bar{\sigma}_{d-3} \Big( \ldots \\
   H_{q+1} \bar{\sigma}_{q+1} \Big( \sum_{i_{q}} \bar{\sigma}_{q} \Big( \left\| {\bf W}_{[q]i_{q},:} \right\|_2^2 \bar{\sigma}_{q-1} \Big( \ldots 
   \!H_0 \bar{\sigma}_{0} \!\Big( D \Big) \!\ldots \!\Big) \\ 
  \end{array}
  }\!\!\!\!\!\!}
  \;\;\mbox{ (\ref{eq:vqip})}
 \]
 we have that:
 \[
 {\!\!\small{
 \begin{array}{rcl}
  0 &\!\!\!\!\leq     {p}_q \left( {\bf W}_{[q]} \right) &\!\!\!\!\leq \left(H    {L} \right)^{d} \frac{D}{H_q H_{q-1}} \left\| {\bf W}_{[q]} \right\|^2_{F_{{\;}_{\;}}} \\
  0 &\!\!\!\!\leq \bar{p}_q
   \left( {\bf W}_{[q]} \right) &\!\!\!\!\leq \left(H\bar{L} \right)^{d} \frac{D}{H_q H_{q-1}} \left\| {\bf W}_{[q]} \right\|^2_{F_{{\;}_{\;}}} \\
 \end{array}
 }}
 \]
 where $H$, $L$ and $\bar{L}$ are geometric means of $H_q$, $L_q$ and $\bar{L}_q$.  
 Furthermore, if $\infset{W}_{q}$ is compact for a given $q \in \infset{N}_d$ then 
 $p_{[q]}^{1/2}, \bar{p}_{[q]}^{1/2} : \infset{W}_q \to \infset{R}$ are, 
 respectively, an $F$-norm and a quasi-$F$-norm on $\infset{W}_q$, both of 
 which are topologically equivalent to the Frobenius norm $\| \cdot \|_{F}$ 
 with bounds:
 \[
 {\!\!\small{
 \begin{array}{rcl}
  \left(\!\frac{H}{    {L}'}\!\right)^{\!d} \!\frac{D}{H_q \!H_{q-1}} \!\left\| \!{\bf W}_{[q]}\! \right\|_F^2 &\!\!\!\!\!\!\leq     {p}_{[q]} \!\left( {\bf W}_{[q]} \right) &\!\!\!\!\!\!\leq\! \left(\!H    {L} \right)^{\!d} \!\frac{D}{H_q \!H_{q-1}} \!\left\| \!{\bf W}_{[q]}\! \right\|^2_{F_{{\;}_{\;}}} \\
  \left(\!\frac{H}{\bar{L}'}\!\right)^{\!d} \!\frac{D}{H_q \!H_{q-1}} \!\left\| \!{\bf W}_{[q]}\! \right\|_F^2 &\!\!\!\!\!\!\leq \bar{p}_{[q]} \!\left( {\bf W}_{[q]} \right) &\!\!\!\!\!\!\leq\! \left(\!H\bar{L} \right)^{\!d} \!\frac{D}{H_q \!H_{q-1}} \!\left\| \!{\bf W}_{[q]}\! \right\|^2_{F_{{\;}_{\;}}} \\
 \end{array}
 }}
 \]
 where $L'$ and $\bar{L}$ are geometric means of $L'_q$ and $\bar{L}'_q$, 
 respectively, where $L'_q |a-b| \leq |\sigma_q(a)-\sigma_q(b)| \leq L_q |a-b|$ 
 and $\bar{L}'_q |a-b| \leq |\bar{\sigma}_q(a)-\bar{\sigma}_q(b)| \leq 
 \bar{L}_q |a-b|$ (that is, both $\sigma_q$ and $\bar{\sigma}_q$ are 
 bi-Lipschitz on the bounded domain implied by the compactness of 
 $\infset{W}_q$).
}
\begin{proof}
We first note that, using the properties of $\bg{\featmap}$:
\[
 {\small
 \begin{array}{rl}
  \bg{\featmap}_{[q+]} \left( {\bf W}^\tsp \odot {\bf U}^\tsp \right) &\!\!\!\!= \bg{\featmap}_{[q+]} \left( {\bf W}^\tsp \right) \odot \bg{\featmap}_{[q+]} \left( {\bf U}^\tsp \right) \\
  \bg{\featmap}_{[q+]} \left( \left| {\bf W} \right|^\tsp \right) &\!\!\!\!= \left| \bg{\featmap}_{[q+]} \left( {\bf W}^\tsp \right) \right| \\
 \end{array}
 }
\]
and so, using theorem \ref{th:collapse}, we can derive (\ref{eq:vqiip}):
\[
 {\small{
 \begin{array}{l}
 \begin{array}{rl}
  p_{[q]} \left( {\bf W}_{[q]} \right)
  &\!\!\!\!=
  \liip \bg{\featmap}_{[q]} \left( {\bf W}_{[q]}^\tsp \right), \bg{\featmap}_{[q]} \left( {\bf W}_{[q]}^\tsp \right) \riip{{\bf g}_{\NN}} \\
  &\!\!\!\!=
  \lmiip \bg{\featmap}_{[q]} \left( {\bf W}_{[q]}^\tsp \right)^{\odot 2} \rmiip{1,{\bf g}_{\NN}} \\
  &\!\!\!\!=
  \lmiip \bg{\featmap}_{[q]} \left( {\bf W}_{[q]}^{\odot 2 \tsp} \right) \rmiip{1,{\bf g}_{\NN}} 
 \end{array} \\
 \begin{array}{rl}
  &\!\!\!\!=
  \lmiip {\bf 1}_{\!H_{d-1}} \!\!\otimes\! {\bg{\featmap}} \!\!\left(\!\! \ldots {\bf 1}_{\!H_{q+1}} \!\!\otimes\! {\bg{\featmap}} \!\!\left( \!\left[\!\!\! \begin{array}{c} {\bg{\featmap}} \left( {\bf W}_{[q]0,:}^{\odot 2} \!\otimes \!{\bf e}_{[q]} \right) \\ {\bg{\featmap}} \left( {\bf W}_{[q]1,:}^{\odot 2} \!\otimes \!{\bf e}_{[q]} \right) \\ \vdots \\ \end{array} \!\!\!\right] \!\right) \!\!\right) \rmiip{1,{\bf g}_{\NN}} \\
  &\!\!\!\!=
  H_{d-1} \lmiip {\bg{\featmap}} \!\!\left(\!\! \ldots {\bf 1}_{\!H_{q+1}} \!\!\otimes\! {\bg{\featmap}} \!\!\left( \!\left[\!\!\! \begin{array}{c} {\bg{\featmap}} \left( {\bf W}_{[q]0,:}^{\odot 2} \!\otimes \!{\bf e}_{[q]} \right) \\ {\bg{\featmap}} \left( {\bf W}_{[q]1,:}^{\odot 2} \!\otimes \!{\bf e}_{[q]} \right) \\ \vdots \\ \end{array} \!\!\!\right] \!\right) \!\!\right) \ldots \\
  &\ldots \rmiip{1,{\bg{\scriptstyle\gamma}}_{[d-1]} \odot \bg{\scriptstyle\featmap} ({\bf 1}_{H_{d-2}} \otimes {\bf g}_{[d-2]})} \\
  &\!\!\!\!=
  H_{d-1} \sigma_{d-1} \Bigg( \lmiip \ldots {\bf 1}_{\!H_{q+1}} \!\!\otimes\! {\bg{\featmap}} \!\!\left( \!\left[\!\!\! \begin{array}{c} {\bg{\featmap}} \left( {\bf W}_{[q]0,:}^{\odot 2} \!\otimes \!{\bf e}_{[q]} \right) \\ {\bg{\featmap}} \left( {\bf W}_{[q]1,:}^{\odot 2} \!\otimes \!{\bf e}_{[q]} \right) \\ \vdots \\ \end{array} \!\!\!\right] \!\right) \!\!\ldots \\
  &\ldots \rmiip{1,{\bf 1}_{H_{d-2}} \otimes {\bf g}_{[d-2]}} \Bigg) \\
 \end{array}
 \end{array}
 }}
\]
\[
 {\small{
 \begin{array}{l}
 \begin{array}{rl}
  &\!\!\!\!=
  H_{d-1} \sigma_{d-1} \Bigg( H_{d-2} \sigma_{d-2} \Bigg( \ldots H_{q+1} \sigma_{q+1} \Bigg( \\
  &\ldots \lmiip \left[\!\!\! \begin{array}{c} {\bg{\featmap}} \left( {\bf W}_{[q]0,:}^{\odot 2} \!\otimes \!{\bf e}_{[q]} \right) \\ {\bg{\featmap}} \left( {\bf W}_{[q]1,:}^{\odot 2} \!\otimes \!{\bf e}_{[q]} \right) \\ \vdots \\ \end{array} \!\!\!\right] \rmiip{1,{\bf 1}_{H_q} \otimes {\bf g}_{[q]}} \Bigg) \Bigg) \Bigg) \\
  &\!\!\!\!=
  H_{d-1} \sigma_{d-1} \Bigg( H_{d-2} \sigma_{d-2} \Bigg( \ldots H_{q+1} \sigma_{q+1} \Bigg( \\
  &\ldots \sum_{i_q} \sigma_q \Bigg( \lmiip {\bf W}_{[q]i_q,:}^{\odot 2} \!\otimes \!{\bf e}_{[q]} \rmiip{1,{\bf g}_{[q]}} \Bigg) \Bigg) \Bigg) \Bigg) \\
  &\!\!\!\!=
  H_{d-1} \sigma_{d-1} \Bigg( H_{d-2} \sigma_{d-2} \Bigg( \ldots H_{q+1} \sigma_{q+1} \Bigg( \\
  &\ldots \sum_{i_q} \sigma_q \Bigg( \sum_{i_{q-1}} W_{[q]i_q,i_{q-1}}^2 \lmiip {\bf e}_{[q]} \rmiip{1,{\bf g}_{[q-1]}} \Bigg) \Bigg) \Bigg) \Bigg) \\
  &\!\!\!\!=
  H_{d-1} \sigma_{d-1} \big( H_{d-2} \sigma_{d-2} \big( \ldots H_{q+1} \sigma_{q+1} \big( \\
  &\ldots \sum_{i_q} \sigma_q \big( \left\| {\bf W}_{[q]i_q,:} \right\|_2^2 \sigma_{q-1} \big( H_{q-1} \ldots H_0 \sigma_0 \big( D \big) \big) \big) \big) \big) \big) \\
 \end{array}
 \end{array}
 }}
\]
The derivation of (\ref{eq:vqip}) follows the same procedure, except that in 
this case we use $|{\bf g}_{\NN}|$, and so the functions are $\bar{\sigma}_q$ 
with Taylor series coefficients $|a_{[q]i}|$.

Recall that an $F$-norm on $\infset{W}_q$ is a function $p:\infset{W}_q \to 
\infset{R}_+$ satisfying $p({\bf U}+{\bf W}) \leq p({\bf U}) + p({\bf W})$ and 
$p({\bf W}) = 0$ iff ${\bf W} = {\bf 0}$, and a quasi-$F$-norm satisfies the 
weaker conditions $p({\bf U}+{\bf W}) \leq c(p({\bf U}) + p({\bf W}))$ for some 
$c \in \infset{R}_+$ and $p({\bf W}) = 0$ iff ${\bf W} = {\bf 0}$.  Clearly the 
right-hand-sides of (\ref{eq:vqiip}) and (\ref{eq:vqip}) satisfy the positivity 
requirement, and it is not difficult to see from the concavity assumption on 
$\sigma_q$ (and hence convexity on $\bar{\sigma}_q$) that the condition $p({\bf 
W}) = 0$ iff ${\bf W} = {\bf 0}$ is satisfied.  The increasing, concave 
assumption on $\sigma_q$, combined with the fact that ${\bf W}_{[q]}$ on the 
right-hand-side of (ref{eq:vqiip}) only occurs in a norm (and hence satisfies 
the triangle inequality) suffices to show that the right-side of 
(\ref{eq:vqiip}) satisfies $p({\bf U}+{\bf W}) \leq p({\bf U}) + p({\bf W})$.  
So the right-side of (\ref{eq:vqiip}) is an $F$-norm on $\infset{W}_q$.  Note 
that $\sigma_q$ bi-Lipschitz implies $\bar{\sigma}_q$ bi-Lipschitz with 
${\bar{L}'}_q{}^{-1} |x-x'| \leq |\bar{\sigma}_q(x)-\bar{\sigma}_q(x')| \leq 
\bar{L}_q |x-x'|$ for some $\bar{L}_q, {\bar{L}'}_q$.  Then this, the 
previously noted facts, and the compactness (hence boundedness) of 
$\infset{W}_q$, then implies that there exists $c \in \infset{R}_+$ such that 
the right-hand-side of (\ref{eq:vqip}) will satisfy $p({\bf U}+{\bf W}) \leq 
c(p({\bf U}) + p({\bf W}))$, and hence is a quasi-$F$-norm on $\infset{W}_q$.

Finally, we recall that by assumption $\sigma_q$ is bi-Lipschitz for all $q$ 
with constant $L_q$.  Hence, trivially, recalling that $\sigma_q (x) > 0$ 
for all $x \in \infset{R}_+$:
\[
 {\small{
 \begin{array}{l}
  \sigma_{d-1} \Big( H_{d-2} \sigma_{d-2} \Big( H_{d-3} \sigma_{d-3} \Big( \ldots H_{q+1} \sigma_{q+1} \Big( \sum_{i_{q}} \ldots \\
  \sigma_{q} \Big( \left\| {\bf W}_{[q]i_{q},:} \right\|_2^2 \sigma_{q-1} \Big( H_{q-2} \ldots H_0 \sigma_{0} \Big( D \Big) \Big) \Big) \Big) \Big) \Big) \Big) \\
  \leq \sum_{i_{q}} L_{d-1} H_{d-2} L_{d-2} H_{d-3} L_{d-3} \ldots H_{q+1} L_{q+1} L_{q} \ldots \\
  \left\| {\bf W}_{[q]i_{q},:} \right\|_2^2 L_{q-1} H_{q-2} \ldots H_0 L_{0} D 
 \end{array}
 }}
\]
\[
 {\small{
 \begin{array}{l}
  \sigma_{d-1} \Big( H_{d-2} \sigma_{d-2} \Big( H_{d-3} \sigma_{d-3} \Big( \ldots H_{q+1} \sigma_{q+1} \Big( \sum_{i_{q}} \ldots \\
  \sigma_{q} \Big( \left\| {\bf W}_{[q]i_{q},:} \right\|_2^2 \sigma_{q-1} \Big( H_{q-2} \ldots H_0 \sigma_{0} \Big( D \Big) \Big) \Big) \Big) \Big) \Big) \Big) \\
  \geq \sum_{i_{q}} L_{d-1}^{-1} H_{d-2} L_{d-2}^{-1} H_{d-3} L_{d-3}^{-1} \ldots H_{q+1} L_{q+1}^{-1} L_{q}^{-1} \ldots \\
  \left\| {\bf W}_{[q]i_{q},:} \right\|_2^2 L_{q-1}^{-1} H_{q-2} \ldots H_0 L_{0}^{-1} D 
 \end{array}
 }}
\]
and so:
 \[
 {\small{
 \begin{array}{rcl}
  \frac{\left( H    {L}^{-1} \right)^d D}{H_q H_{q-1}} \left\| {\bf W}_{[q]} \right\|_F^2 &\!\!\!\!\leq     {p}_{[q]} \left( {\bf W}_{[q]} \right) &\!\!\!\!\leq \frac{\left( H    {L} \right)^d D}{H_q H_{q-1}} \left\| {\bf W}_{[q]} \right\|^2_{F_{{\;}_{\;}}} \\
 \end{array}
 }}
 \]
We also note that by the assumptions on $\sigma_q$ we have $\bar{\sigma}_q$ 
convex, bi-Lipschitz, and $\bar{\sigma}_q (0) = 0$.  Let $\bar{L}_q$ be the 
associated Lipschitz constant.  Hence:
\[
 {\small{
 \begin{array}{l}
  \bar{\sigma}_{d-1} \Big( H_{d-2} \bar{\sigma}_{d-2} \Big( H_{d-3} \bar{\sigma}_{d-3} \Big( \ldots H_{q+1} \bar{\sigma}_{q+1} \Big( \sum_{i_{q}} \ldots \\
  \bar{\sigma}_{q} \Big( \left\| {\bf W}_{[q]i_{q},:} \right\|_2^2 \bar{\sigma}_{q-1} \Big( H_{q-2} \ldots H_0 \bar{\sigma}_{0} \Big( D \Big) \Big) \Big) \Big) \Big) \Big) \Big) \\
  \leq \sum_{i_{q}} \bar{L}_{d-1} H_{d-2} \bar{L}_{d-2} H_{d-3} \bar{L}_{d-3} \ldots H_{q+1} \bar{L}_{q+1} \bar{L}_{q} \ldots \\
  \left\| {\bf W}_{[q]i_{q},:} \right\|_2^2 \bar{L}_{q-1} H_{q-2} \ldots H_0 \bar{L}_{0} D 
 \end{array}
 }}
\]
\[
 {\small{
 \begin{array}{l}
  \bar{\sigma}_{d-1} \Big( H_{d-2} \bar{\sigma}_{d-2} \Big( H_{d-3} \bar{\sigma}_{d-3} \Big( \ldots H_{q+1} \bar{\sigma}_{q+1} \Big( \sum_{i_{q}} \ldots \\
  \bar{\sigma}_{q} \Big( \left\| {\bf W}_{[q]i_{q},:} \right\|_2^2 \bar{\sigma}_{q-1} \Big( H_{q-2} \ldots H_0 \bar{\sigma}_{0} \Big( D \Big) \Big) \Big) \Big) \Big) \Big) \Big) \\
  \geq \sum_{i_{q}} \bar{L}_{d-1}^{-1} H_{d-2} \bar{L}_{d-2}^{-1} H_{d-3} \bar{L}_{d-3}^{-1} \ldots H_{q+1} \bar{L}_{q+1}^{-1} \bar{L}_{q}^{-1} \ldots \\
  \left\| {\bf W}_{[q]i_{q},:} \right\|_2^2 \bar{L}_{q-1}^{-1} H_{q-2} \ldots H_0 \bar{L}_{0}^{-1} D 
 \end{array}
 }}
\]
and so:
 \[
 {\small{
 \begin{array}{rcl}
  \frac{\left( H\bar{L}^{-1} \right)^d D}{H_q H_{q-1}} \left\| {\bf W}_{[q]} \right\|_F^2 &\!\!\!\!\leq \bar{p}_{[q]} \left( {\bf W}_{[q]} \right) &\!\!\!\!\leq \frac{\left( H\bar{L} \right)^d D}{H_q H_{q-1}} \left\| {\bf W}_{[q]} \right\|^2_{F_{{\;}_{\;}}} \\
 \end{array}
 }}
 \]
hence these are topologically equivalent to $\| \cdot \|_F$.  The final result 
follows from simple arithmetic.
\end{proof}

{\bf Theorem \ref{cor:collapsed} }
{\it
 Using the notation of theorem \ref{cor:collapse}, defining 
 $\liip {\bf v}_{\NN}, {\bf v}_{\NN} \riip{{\bf g}_{\NN}} = p_{\NN} ({\bf 
 W}_{[0]}, {\bf W}_{[1]}, \ldots)$ and $\lip {\bf v}_{\NN}, {\bf v}_{\NN} 
 \rip{|{\bf g}_{\NN}|} = \bar{p}_{\NN} ({\bf W}_{[0]}, {\bf W}_{[1]}, \ldots)$, 
 where:
 \[
  {\!\!\!\!\!\!\small{
  \begin{array}{l}
   p_{\NN} \left( {\bf W}_{[0]}, {\bf W}_{[1]}, \ldots \right)
   \!=\! \sigma_{d-1} \Big( {\sum}_{i_{d-2}} \left| W_{[d-1] 0,i_{d-2}} \right|^2 \ldots \\
   \sigma_{d-2} \Big( \ldots {\sum}_{i_0} \left| W_{[1]i_1,i_0} \right|^2 \sigma_{0} \Big( \left\| {\bf W}_{[0],i_0,:} \right\|_2^2 \Big) \Big) \Big) 
  \end{array}
  }\!\!\!\!\!\!}
  \mbox{ (\ref{eq:viip})}
 \]
 \[
  {\!\!\!\!\!\!\small{
  \begin{array}{l}
   \bar{p}_{\NN} \left( {\bf W}_{[0]}, {\bf W}_{[1]}, \ldots \right)
   \!=\! \bar{\sigma}_{d-1} \Big( {\sum}_{i_{d-2}} \left| W_{[d-1] 0,i_{d-2}} \right|^2 \ldots \\
   \bar{\sigma}_{d-2} \Big( \ldots {\sum}_{i_0} \left| W_{[1]i_1,i_0} \right|^2 \bar{\sigma}_{0} \Big( \left\| {\bf W}_{[0],i_0,:} \right\|_2^2 \Big) \Big) \Big) 
  \end{array}
  }\!\!\!\!\!\!}
  \mbox{ (\ref{eq:vip})}
 \]
 we have that:
 \[
  {\small{
  \begin{array}{rccl}
   0 &\!\!\!\!\leq \liip {\bf v}_{\NN},{\bf v}_{\NN} \riip{\bf g}    &\!\!\!\!\leq     {L}^d \prod_q \left\|{\bf W}_{[q]}\right\|_{[q]_{{\;}_{\;}}\!\!\!\!}^2 &\!\!\!\!\leq \left( \frac{    {L}}{d} \sum_q \left\| {\bf W}_{[q]} \right\|_{[q]}^2 \right)^d \\
   0 &\!\!\!\!\leq \lip  {\bf v}_{\NN},{\bf v}_{\NN} \rip{|{\bf g}|} &\!\!\!\!\leq \bar{L}^d \prod_q \left\|{\bf W}_{[q]}\right\|_{[q]                    }^2 &\!\!\!\!\leq \left( \frac{\bar{L}}{d} \sum_q \left\| {\bf W}_{[q]} \right\|_{[q]}^2 \right)^d \\
  \end{array}
  }}
 \]
 Furthermore:
 \[
 {\small{
 \begin{array}{rccl}
  \left( \frac{1}{    {L}'} \right)^d \prod_q \left\| {\bf W}_{[q]} \right\|_F^2 &\!\!\!\!\leq&\!\!\!\! \liip {\bf v}_{\NN}, {\bf v}_{\NN} \riip{{\bf g}_{\NN}} &\!\!\!\!\!\leq       L^d \prod_q \left\| {\bf W}_{[q]} \right\|_{F_{{\;}_{\;}}}^2 \\
  \left( \frac{1}{\bar{L}'} \right)^d \prod_q \left\| {\bf W}_{[q]} \right\|_F^2 &\!\!\!\!\leq&\!\!\!\! \lip {\bf v}_{\NN}, {\bf v}_{\NN} \rip{|{\bf g}_{\NN}|} &\!\!\!\!\!\leq \bar{L}^d \prod_q \left\| {\bf W}_{[q]} \right\|_F^2 \\
 \end{array}
 }}
 \]
}
\begin{proof}
The derivations of (\ref{eq:viip}) and (\ref{eq:vip}) follows a similar 
structure to the derivation of (\ref{eq:vqiip}) and (\ref{eq:vqip}) in the 
proof of theorem \ref{cor:collapse}, except that at every layer we encounter a 
weight matrix, so the result is as shown.  Subsequently we apply the 
bi-Lipschitz property to bound the activation functions, thereby obtaining the 
second result, and finally simple arithmetic and the AM-GM inequality for the 
final bounds.
\end{proof}

\subsection{Equivalent SVMs for Deep Networks}


{\bf Theorem \ref{th:kernels} } 
{\it
 Let the feature map ${\bg{\featmapmkern}} : \infset{R}^D \to \frkks{}$ and 
 metric ${\bf g}$ be defined by the deep network (\ref{eq:deepnetform}) as per 
 theorem \ref{th:collapse}.  Then:
 \begin{equation}
  \begin{array}{l}
   K_{\NN} \left( {\bf x}, {\bf x}' \right)
   = \liip {\bg{\featmapmkern}}_{\NN} \left( {\bf x} \right), {\bg{\featmapmkern}}_{\NN} \left( {\bf x}' \right) \riip{{\bf g}_{\NN}}
   = \sigma_{d-1} ( \ldots \\
   \!\!\!H_{d-2} \sigma_{d-2} ( H_{d-3} \ldots H_1 \sigma_{1} ( H_0 \sigma_{0} ( \lip {\bf x}, {\bf x}' \rip{{\bf 1}} ) ) ) ) \\
  \end{array}
  \label{eq:whatisK}
 \end{equation}
 is the corresponding {\krein} kernel, and:
 \begin{equation}
  {\!\!
  \begin{array}{l}
   \bar{K}_{\NN} \left( {\bf x}, {\bf x}' \right)
   = \lip {\bg{\featmapmkern}}_{\NN} \left( {\bf x} \right), {\bg{\featmapmkern}}_{\NN} \left( {\bf x}' \right) \rip{|{\bf g}_{\NN}|}
   = \bar{\sigma}_{d-1} ( \ldots \\
   \!\!\!H_{d-2} \bar{\sigma}_{d-2} ( H_{d-3} \ldots H_1 \bar{\sigma}_{1} ( H_0 \bar{\sigma}_{0} ( \lip {\bf x}, {\bf x}' \rip{{\bf 1}} ) ) ) ) \\
  \end{array}
  \!\!}
  \label{eq:whatisbarK}
 \end{equation}
 the associated kernel where, if $\sigma_{q} (\xi) = \sum_i a_i \xi^i$ 
 then $\bar{\sigma}_{q} (\xi) = \sum_i |a_i| \xi^i$ (e.g. see table 1 in the 
 supplementary).
}
\begin{proof}
%
%
Using the notation and definitions in the proof of theorem \ref{th:collapse}, 
and applying the definitions (theorem \ref{th:collapse} and lemma 
\ref{th:pushforward}):
\[
 \small{
 \begin{array}{l}
  K_{\NN} \left( {\bf x}, {\bf x}' \right)
    = \liip {\bg{\featmapmkern}} \left( {\bf x} \right), {\bg{\featmapmkern}} \left( {\bf x}' \right) \riip{{\bf g}} \\
  \;= \liip {\bg{\featmap}}_{[0,d-1]} \left( {\bf x} \right), {\bg{\featmap}}_{[0,d-1]} \left( {\bf x}' \right) \riip{{\bf g}} \\
  \;= \liip {\bg{\featmap}} \left( {\bg{\featmap}}_{[0,d-2]} \left( {\bf x} \right) \right), {\bg{\featmap}} \left( {\bg{\featmap}}_{[0,d-2]} \left( {\bf x}' \right) \right) \riip{{\bg{\scriptstyle\gamma}}_{[d-1]} \odot {\bg{\scriptstyle\featmap}} (\ldots)} \\
  \!= \sigma_{d-1} \left( \liip {\bg{\featmap}}_{[0,d-2]} \left( {\bf x} \right), {\bg{\featmap}}_{[0,d-2]} \left( {\bf x}' \right) \riip{{\bf 1}_{H_{d-2}} \otimes ({\bg{\scriptstyle\gamma}}_{[d-2]} \odot {\bg{\scriptstyle\featmap}} (\ldots))} \right) \\
  \!= \sigma_{d-1} \left( H_{d-2} \sigma_{d-2} \left( \liip {\bg{\featmap}}_{[0,d-3]} \left( {\bf x} \right), {\bg{\featmap}}_{[0,d-3]} \left( {\bf x}' \right) \riip{{\bf 1}_{H_{d-3}} \otimes ({\bg{\scriptstyle\gamma}}_{[d-3]} \odot {\bg{\scriptstyle\featmap}} (\ldots))} \right) \right) \\
  \!= \ldots \\
  \!= \sigma_{d-1} \left( H_{d-2} \sigma_{d-2} \left( \ldots H_1 \sigma_{1} \left( H_0 \sigma_{0} \left( \lip {\bf x}, {\bf x}' \rip{{\bf 1}} \right) \right) \right) \right) \\
 \end{array}
 }
\]
and likewise, the associated kernel $\bar{K}_{\NN}$ is:
\[
 \small{
 \begin{array}{l}
  \bar{K}_{\NN} \left( {\bf x}, {\bf x}' \right)
    = \lip {\bg{\featmapmkern}} \left( {\bf x} \right), {\bg{\featmapmkern}} \left( {\bf x}' \right) \rip{|{\bf g}|} \\
  \;= \lip {\bg{\featmap}}_{[0,d-1]} \left( {\bf x} \right), {\bg{\featmap}}_{[0,d-1]} \left( {\bf x}' \right) \rip{|{\bf g}|} \\
  \;= \lip {\bg{\featmap}} \left( {\bg{\featmap}}_{[0,d-2]} \left( {\bf x} \right) \right), {\bg{\featmap}} \left( {\bg{\featmap}}_{[0,d-2]} \left( {\bf x}' \right) \right) \rip{|{\bg{\scriptstyle\gamma}}_{[d-1]} \odot {\bg{\scriptstyle\featmap}} (\ldots)|} \\
  \!= \bar{\sigma}_{d-1} \left( \lip {\bg{\featmap}}_{[0,d-2]} \left( {\bf x} \right), {\bg{\featmap}}_{[0,d-2]} \left( {\bf x}' \right) \rip{|{\bf 1}_{H_{d-2}} \otimes ({\bg{\scriptstyle\gamma}}_{[d-2]} \odot {\bg{\scriptstyle\featmap}} (\ldots))|} \right) \\
  \!= \bar{\sigma}_{d-1} \left( H_{d-2} \bar{\sigma}_{d-2} \left( \lip {\bg{\featmap}}_{[0,d-3]} \left( {\bf x} \right), {\bg{\featmap}}_{[0,d-3]} \left( {\bf x}' \right) \rip{|{\bf 1}_{H_{d-3}} \otimes ({\bg{\scriptstyle\gamma}}_{[d-3]} \odot {\bg{\scriptstyle\featmap}} (\ldots))|} \right) \right) \\
  \!= \ldots \\
  \!= \bar{\sigma}_{d-1} \left( H_{d-2} \bar{\sigma}_{d-2} \left( \ldots H_1 \bar{\sigma}_{1} \left( H_0 \bar{\sigma}_{0} \left( \lip {\bf x}, {\bf x}' \rip{{\bf 1}} \right) \right) \right) \right) \\
 \end{array}
 }
\]
where, if $\sigma_{q} (\xi) = \sum_i a_i \xi^i$ (recall that $\sigma_{q}$ 
is entire, so this Taylor series exists) then $\bar{\sigma}_{q} (\xi) = 
\sum_i |a_i| \xi^i$.  See table 1 in the supplementary for examples.
\end{proof}

\subsection{Rademacher Complexity Analysis}

{\bf Theorem \ref{th:boundsvm} } 
{\it
 Let $K_{\NN}$ be a {\krein} kernel and $\bar{K}_{\NN}$ be its associated 
 kernel such that ${\bf x} \to \bar{K}_{\NN} ({\bf x}, {\bf x}) \in L_1 
 (\infset{X}, \nu)$ and $\bar{K}_{\NN} ({\bf x}, {\bf x}) \geq 0$ $\forall {\bf x} 
 \in \infset{X}$.  Then:
 \[
  \begin{array}{l}
   \mathcal{R}_{N} \left( \unitnormballkrein{\SVM} \,\right) \leq \frac{1}{\sqrt{N}} \left( R_{\SVM} \int_{{\bf x} \in \infset{X}} \bar{K}_{\NN} \left( {\bf x}, {\bf x} \right) d\nu ({\bf x}) \right)^{\frac{1}{2}}
  \end{array} \mbox{   (\ref{eq:radboundsvm})}
 \]
}
\begin{proof}
Following \cite{Men2}, we first prove the following, where we use the 
Cauchy-Schwarz inequality at step $\#$, the fact that the Rademacher complexity 
of a ball in {\RKKS} is the same as the Rademacher complexity of the same ball 
in the associated {\RKHS} at step $*$, Jensen's inequality at step $\wedge$, 
and independence at step $\vee$:
\[
{\small
 \begin{array}{l}
  \mathbb{E}_\epsilon \left[ \mathop{\sup}\limits_{f \in \subrkks{K_{\NN}} | \liip f,f \riip{\subsubrkks{K_{\NN}}} \leq R_{\SVM}} \left| \frac{1}{N} \sum_i \epsilon_i f \left( {\bf x}_i \right) \right| \right] \\
   = \mathbb{E}_\epsilon \left[ \mathop{\sup}\limits_{\ldots} \left| \frac{1}{N} \liip \sum_i \epsilon_i \bg{\featmapmkern}_{\NN} \left( {\bf x}_i \right), {\bf v} \riip{\bf g} \right| \right] \\
   = \frac{1}{N} \mathbb{E}_\epsilon \left[ \mathop{\sup}\limits_{\ldots} \left| \lip \sum_i \epsilon_i |{\bf g}|^{\odot \frac{1}{2}} \odot \bg{\featmapmkern}_{\NN} \left( {\bf x}_i \right), \sgn \left( {\bf g} \right) \odot |{\bf g}|^{\odot \frac{1}{2}} \odot {\bf v} \rip{\bf 1} \right| \right] \\
   \leq^\# \frac{1}{N} \mathbb{E}_\epsilon \left[ \mathop{\sup}\limits_{\ldots} \left\| \sum_i \epsilon_i |{\bf g}|^{\odot \frac{1}{2}} \odot \bg{\featmapmkern}_{\NN} \left( {\bf x}_i \right) \right\|_2 \left\| |{\bf g}|^{\odot \frac{1}{2}} \odot {\bf v} \right\|_2 \right] \\
   = \frac{1}{N} \mathbb{E}_\epsilon \left[ \mathop{\sup}\limits_{\ldots} \left\| \sum_i \epsilon_i |{\bf g}|^{\odot \frac{1}{2}} \odot \bg{\featmapmkern}_{\NN} \left( {\bf x}_i \right) \right\|_2 \lip {\bf v}, {\bf v} \rip{|{\bf g}|}^{\frac{1}{2}} \right] \\
   \leq^* \frac{1}{N} \mathbb{E}_\epsilon \left[ \left\| \sqrt{R_{\SVM}} \sum_i \epsilon_i |{\bf g}|^{\odot \frac{1}{2}} \odot \bg{\featmapmkern}_{\NN} \left( {\bf x}_i \right) \right\|_2 \right] \\
   \leq^\wedge \frac{1}{N} \left[ R_{\SVM} \mathbb{E}_\epsilon \left\| \sum_i \epsilon_i |{\bf g}|^{\odot \frac{1}{2}} \odot \bg{\featmapmkern}_{\NN} \left( {\bf x}_i \right) \right\|_2^2 \right]^{\frac{1}{2}} \\
   = \frac{1}{N} \left[ R_{\SVM} \mathbb{E}_\epsilon \sum_{ij} \epsilon_i \epsilon_j \lip \bg{\featmapmkern}_{\NN} \left( {\bf x}_i \right), \bg{\featmapmkern}_{\NN} \left( {\bf x}_j \right) \rip{|{\bf g}|} \right]^{\frac{1}{2}} \\
   =^\vee \frac{1}{N} \left( R_{\SVM} \sum_i \lip \bg{\featmapmkern}_{\NN} \left( {\bf x}_i \right), \bg{\featmapmkern}_{\NN} \left( {\bf x}_i \right) \rip{|{\bf g}|} \mathbb{E}_\epsilon \left[ \epsilon_i^2 \right] \right)^{\frac{1}{2}} \\
   = \frac{1}{\sqrt{N}} \sqrt{R_{\SVM} \sum_i \bar{K}_{\NN} \left( {\bf x}_i, {\bf x}_i \right)} \\
 \end{array}
}
\]
Then, using the properties of Rademacher complexity (again following 
\cite{Men2}), we have that:
\[
 \begin{array}{rl}
  \mathcal{R}_N \left( \unitnormballkrein{\SVM} \,\right) &\!\!\!\!= \mathbb{E}_\nu \mathbb{E}_\epsilon \left[ \mathop{\sup}\limits_{f \in \subrkks{K_{\NN}} | \liip f,f \riip{\subsubrkks{K_{\NN}}} \leq R_{\SVM}} \left| \frac{1}{N} \sum_i \epsilon_i f \left( {\bf x}_i \right) \right| \right] \\
  &\!\!\!\!\leq \mathbb{E}_\nu \frac{1}{\sqrt{N}} \sqrt{R_{\SVM} \sum_i \bar{K}_{\NN} \left( {\bf x}_i, {\bf x}_i \right)} \\
  &\!\!\!\!\leq^\wedge \frac{1}{\sqrt{N}} \sqrt{R_{\SVM} \mathbb{E}_\nu \sum_i \bar{K}_{\NN} \left( {\bf x}_i, {\bf x}_i \right)} \\
  &\!\!\!\!\leq \frac{1}{\sqrt{N}} \sqrt{R_{\SVM} \int_{{\bf x} \in \infset{X}} \bar{K}_{\NN} \left( {\bf x},{\bf x} \right) d\nu\left({\bf x}\right)} \\
 \end{array}
\]
\end{proof}

{\bf Theorem \ref{th:tightboundnn} } 
{\it
 Let $\sigma_q$ be concave on $\infset{R}_+$, $\sigma_q(0) = 0$ and 
 $\sigma_q(-\xi) = -\sigma_q(\xi)$ in addition to the usual assumptions.  
 Let:
 \[
  {\small{
  \begin{array}{l}
   \chi_{\NN} \left( \xi \right) = \sigma_{d-1} \left( d \sqrt{H_{d-2}} \sigma_{d-2} \left( d \sqrt{H_{d-3}} \sigma_{d-3} \left( \ldots d \sqrt{H_0} \sigma_{0} \left( \xi \right) \right) \right) \right) \\
  \end{array}
  }}
 \]
 If ${\bf x} \to \chi_{\NN} (\| {\bf x} \|_2) \in L_1 (\infset{X}, \nu)$ then:
 \[
  {\small{
  \begin{array}{rl}
   \mathcal{R}_N \left( \unitnormballkrein{\NN} \,\right) &\!\!\!\!= \leq \frac{\max \{ 1,R_{\NN}^d \}}{\sqrt{N}} \sqrt{\int \chi_{\NN}^2 \left( \left\| {\bf x}_i \right\|_2 \right) d\nu \left( {\bf x} \right)} \\
  \end{array}
  }}
 \]
 Moreover if $\sigma_q$ is unbounded for all $q$ then:
 \[
  {\small{
   \begin{array}{l}
    \mathcal{R}_N \left( \unitnormballkrein{\NN} \right) \leq \frac{\max \left\{ 1,R_{\NN}^d \right\}}{\sqrt{N}} \left( d\sqrt{H}L \right)^d \sqrt{\int \left\| {\bf x}_i \right\|_2^2 d\nu ({\bf x})}
   \end{array}
  }}
 \]
 and otherwise, if $\sigma_{q} (\xi) \leq 1$ $\forall \xi \in \infset{R}_+$ for 
 some $q \in \infset{N}_d$ then:
 \[
  {\small{
  \begin{array}{rl}
   \mathcal{R}_N \left( \unitnormballkrein{\NN} \,\right) 
   &\!\!\!\!\leq \frac{\max \left\{ 1,R_{\NN}^d \right\}}{\sqrt{N}} \left( d \sqrt{H_{[q+]}} L_{[q+]} \right)^{d-q-1} 
  \end{array}
  }}
 \]
 where $H_{[q+]} = \GM (H_{q+1},\ldots,H_{d-1})$ and $L_{[q+]} = \GM (L_{q+1}, 
 \ldots,L_{d-1})$ are geometric means.
}

\begin{proof}
We start by considering:
\[
{\small
 \begin{array}{l}
  \mathbb{E}_\epsilon \left[ \mathop{\sup}\limits_{f = \liip {\bf v}_{\NN}, \bg{\scriptstyle\featmapmkern} (\cdot) \riip{\bf g} | {\bf v}_{\NN} \in \frkks{\NN}, \liip {\bf v}_{\NN}, {\bf v}_{\NN} \riip{\bf g} \leq R_{\NN}} \left| \frac{1}{N} \sum_i \epsilon_i f \left( {\bf x}_i \right) \right| \right] \\
  = \mathbb{E}_\epsilon \left[ \mathop{\sup}\limits_{{\bf W}_{[q]} \in \infset{W}_q | \frac{1}{d} \sum_q \left\| {\bf W}_{[q]} \right\|_F^2 \leq R_{\NN}} \left| \frac{1}{N} \sum_i \epsilon_i \liip {\bf v}_{\NN}, \bg{\featmapmkern} \left( {\bf x}_i \right) \riip{\bf g} \right| \right] \\
  = \mathbb{E}_\epsilon \Big[ \mathop{\sup}\limits_{\ldots} \Big| \frac{1}{N} \sum_i \epsilon_i 
    \sigma_{d-1} \Big( {\sum}_{i_{d-2}} W_{[d-1] 0,i_{d-2}} \sigma_{d-2} \Big( \ldots \\
    \;\;\;\;\;\;\;\;\;\;\ldots {\sum}_{i_0} W_{[1]i_1,i_0} \sigma_{0} \Big( {\bf W}_{[0],i_0,:}^\tsp {\bf x}_i \Big) \Big) \Big)
    \Big| \Big] \\
 \end{array}
}
\]
Using our assumptions on $\sigma_q$ and subsequently the Cauchy-Schwarz 
inequality it follows that:
\[
{\small
 \begin{array}{l}
  \mathbb{E}_\epsilon \left[ \mathop{\sup}\limits_{f = \liip {\bf v}_{\NN}, \bg{\scriptstyle\featmapmkern} (\cdot) \riip{\bf g} | {\bf v}_{\NN} \in \frkks{\NN}, \liip {\bf v}_{\NN}, {\bf v}_{\NN} \riip{\bf g} \leq R_{\NN}} \left| \frac{1}{N} \sum_i \epsilon_i f \left( {\bf x}_i \right) \right| \right] \\
  \leq \mathbb{E}_\epsilon \Big[ \mathop{\sup}\limits_{\ldots, s_i = \pm 1} \Big| \frac{1}{N} \sum_i \epsilon_i 
    s_i \sigma_{d-1} \Big( {\sum}_{i_{d-2}} \left| W_{[d-1] 0,i_{d-2}} \right| \sigma_{d-2} \Big( \ldots \\
    \;\;\;\;\;\;\;\;\;\;\ldots {\sum}_{i_0} \left| W_{[1]i_1,i_0} \right| \sigma_{0} \Big( \left| {\bf W}_{[0],i_0,:}^\tsp {\bf x}_i \right| \Big) \Big) \Big)
    \Big| \Big] \\
  \leq \mathbb{E}_\epsilon \Big[ \mathop{\sup}\limits_{\ldots} \Big| \frac{1}{N} \sum_i \epsilon_i 
    s_i \sigma_{d-1} \Big( {\sum}_{i_{d-2}} \left| W_{[d-1] 0,i_{d-2}} \right| \sigma_{d-2} \Big( \ldots \\
    \;\;\;\;\;\;\;\;\;\;\ldots {\sum}_{i_0} \left| W_{[1]i_1,i_0} \right| \sigma_{0} \Big( \left\| {\bf W}_{[0],i_0,:} \right\|_2 \left\| {\bf x}_i \right\|_2 \Big) \Big) \Big)
    \Big| \Big] \\
 \end{array}
}
\]
Then, by Jensen's inequality on concave functions and subsequently the 
Cauchy-Schwarz inequality and that $\| \cdot \|_2 \leq \| \cdot \|_1 \leq 
\sqrt{n} \| \cdot \|_2$ for the $1$- and $2$-norms on $\infset{R}^n$:
\[
{\small
 \begin{array}{l}
  \mathbb{E}_\epsilon \left[ \mathop{\sup}\limits_{f = \liip {\bf v}_{\NN}, \bg{\scriptstyle\featmapmkern} (\cdot) \riip{\bf g} | {\bf v}_{\NN} \in \frkks{\NN}, \liip {\bf v}_{\NN}, {\bf v}_{\NN} \riip{\bf g} \leq R_{\NN}} \left| \frac{1}{N} \sum_i \epsilon_i f \left( {\bf x}_i \right) \right| \right] \\
  \leq \mathbb{E}_\epsilon \Big[ \mathop{\sup}\limits_{\ldots} \Big| \frac{1}{N} \sum_i \epsilon_i 
    s_i \sigma_{d-1} \Big( {\sum}_{i_{d-2}} \left| W_{[d-1] 0,i_{d-2}} \right| \sigma_{d-2} \Big( \ldots \\
    \;\;\;\;\;\;\;\;\;\;\ldots \left( {\sum}_{i_0} \left| W_{[1]i_1,i_0} \right| \right) \sigma_{0} \Big( \frac{{\sum}_{i_0} \left| W_{[1]i_1,i_0} \right| \left\| {\bf W}_{[0],i_0,:} \right\|_2}{{\sum}_{i_0} \left| W_{[1]i_1,i_0} \right| } \left\| {\bf x}_i \right\|_2 \Big) \Big) \Big)
    \Big| \Big] \\
  \leq \mathbb{E}_\epsilon \Big[ \mathop{\sup}\limits_{\ldots} \Big| \frac{1}{N} \sum_i \epsilon_i 
    s_i \sigma_{d-1} \Big( {\sum}_{i_{d-2}} \left| W_{[d-1] 0,i_{d-2}} \right| \sigma_{d-2} \Big( \ldots \\
    \;\;\;\;\;\;\;\;\;\;\ldots \left\| {\bf W}_{[1]i_1,:} \right\|_1 \sigma_{0} \Big( \frac{\left\| {\bf W}_{[1]i_1,:} \right\|_2}{\left\| {\bf W}_{[1]i_1,:} \right\|_1} \left\| {\bf W}_{[0]} \right\|_F \left\| {\bf x}_i \right\|_2 \Big) \Big) \Big)
    \Big| \Big] \\
  \leq \mathbb{E}_\epsilon \Big[ \mathop{\sup}\limits_{\ldots} \Big| \frac{1}{N} \sum_i \epsilon_i 
    s_i \sigma_{d-1} \Big( {\sum}_{i_{d-2}} \left| W_{[d-1] 0,i_{d-2}} \right| \sigma_{d-2} \Big( \ldots \\
    \;\;\;\;\;\;\;\;\;\;\ldots \left\| {\bf W}_{[1]i_1,:} \right\|_2 \sqrt{H_{0}} \sigma_{0} \Big( \frac{\left\| {\bf W}_{[1]i_1,:} \right\|_1}{\left\| {\bf W}_{[1]i_1,:} \right\|_1} \left\| {\bf W}_{[0]} \right\|_F \left\| {\bf x}_i \right\|_2 \Big) \Big) \Big)
    \Big| \Big] \\
  = \mathbb{E}_\epsilon \Big[ \mathop{\sup}\limits_{\ldots} \Big| \frac{1}{N} \sum_i \epsilon_i 
    s_i \sigma_{d-1} \Big( {\sum}_{i_{d-2}} \left| W_{[d-1] 0,i_{d-2}} \right| \sigma_{d-2} \Big( \ldots \\
    \;\;\;\;\;\;\;\;\;\;\ldots \left\| {\bf W}_{[1]i_1,:} \right\|_2 \sqrt{H_{0}} \sigma_{0} \Big( \left\| {\bf W}_{[0]} \right\|_F \left\| {\bf x}_i \right\|_2 \Big) \Big) \Big)
    \Big| \Big] \\
 \end{array}
}
\]
Repeating the same procedure at each level of the nested activation functions 
and using the definition of $\unitnormballkrein{\NN}$:
\[
{\small
 \begin{array}{l}
  \mathbb{E}_\epsilon \left[ \mathop{\sup}\limits_{f = \liip {\bf v}_{\NN}, \bg{\scriptstyle\featmapmkern} (\cdot) \riip{\bf g} | {\bf v}_{\NN} \in \frkks{\NN}, \liip {\bf v}_{\NN}, {\bf v}_{\NN} \riip{\bf g} \leq R_{\NN}} \left| \frac{1}{N} \sum_i \epsilon_i f \left( {\bf x}_i \right) \right| \right] \\
  \leq \mathbb{E}_\epsilon \Big[ \mathop{\sup}\limits_{\ldots} \Big| \frac{1}{N} \sum_i \epsilon_i 
    s_i \sigma_{d-1} \Big( \left\| {\bf W}_{[d-1]} \right\|_F \sqrt{H_{d-2}} \sigma_{d-2} \Big( \ldots \\
    \;\;\;\;\;\;\;\;\;\;\ldots \left\| {\bf W}_{[1]} \right\|_F \sqrt{H_{0}} \sigma_{0} \Big( \left\| {\bf W}_{[0]} \right\|_F \left\| {\bf x}_i \right\|_2 \Big) \Big) \Big)
    \Big| \Big] \\
  \leq \mathbb{E}_\epsilon \Big[ \mathop{\sup}\limits_{s_i = \pm 1} \Big| \frac{1}{N} \sum_i \epsilon_i 
    s_i \sigma_{d-1} \Big( d R_{\NN} \sqrt{H_{d-2}} \sigma_{d-2} \Big( \ldots \\
    \;\;\;\;\;\;\;\;\;\;\ldots d R_{\NN} \sqrt{H_{0}} \sigma_{0} \Big( d R_{\NN} \left\| {\bf x}_i \right\|_2 \Big) \Big) \Big)
    \Big| \Big] \\
 \end{array}
}
\]
Note that for any concave increasing function $g : \infset{R}_+ \cup \{0\} \to 
\infset{R}_+$ and $a \in [1,\infty)$, $b \in [0,\infty)$ we have that $g(ab) 
\leq a g(b)$.  Using this, and subsequently applying Jensen's inequality and 
the Cauchy-Schwarz inequality, we see that:
\[
{\small
 \begin{array}{l}
  \mathbb{E}_\epsilon \left[ \mathop{\sup}\limits_{f = \liip {\bf v}_{\NN}, \bg{\scriptstyle\featmapmkern} (\cdot) \riip{\bf g} | {\bf v}_{\NN} \in \frkks{\NN}, \liip {\bf v}_{\NN}, {\bf v}_{\NN} \riip{\bf g} \leq R_{\NN}} \left| \frac{1}{N} \sum_i \epsilon_i f \left( {\bf x}_i \right) \right| \right] \\
  \leq \frac{\max \{ 1,R_{\NN}^d \}}{N} \mathbb{E}_\epsilon \Big[ \mathop{\sup}\limits_{\ldots} \Big| \sum_i 
    \epsilon_i s_i \sigma_{d-1} \Big( d \sqrt{H_{d-2}} \sigma_{d-2} \Big( \ldots \\
    \;\;\;\;\;\;\;\;\;\;\ldots d \sqrt{H_{0}} \sigma_{0} \Big( d \left\| {\bf x}_i \right\|_2 \Big) \Big) \Big)
    \Big| \Big] \\
  \leq \frac{\max \{ 1,R_{\NN}^d \}}{N} \Big( \mathbb{E}_\epsilon \Big[ \Big( \mathop{\sup}\limits_{\ldots} \sum_i 
    \epsilon_i s_i \sigma_{d-1} \Big( d \sqrt{H_{d-2}} \sigma_{d-2} \Big( \ldots \\
    \;\;\;\;\;\;\;\;\;\;\ldots d \sqrt{H_{0}} \sigma_{0} \Big( d \left\| {\bf x}_i \right\|_2 \Big) \Big) \Big)
    \Big)^2 \Big] \Big)^{\frac{1}{2}} \\
  \leq \frac{\max \{ 1,R_{\NN}^d \}}{N} \Big( \mathbb{E}_\epsilon \Big[ \mathop{\sup}\limits_{\ldots} 
    \left( \sum_i \left( \epsilon_i s_i \right)^2 \right) \Big( \sum_i \Big( \sigma_{d-1} \Big( d \sqrt{H_{d-2}} \sigma_{d-2} \Big( \ldots \\
    \;\;\;\;\;\;\;\;\;\;\ldots d \sqrt{H_{0}} \sigma_{0} \Big( d \left\| {\bf x}_i \right\|_2 \Big) \Big) \Big) \Big)^2 \Big)
    \Big] \Big)^{\frac{1}{2}} \\
  \leq \frac{\max \{ 1,R_{\NN}^d \}}{\sqrt{N}} \sqrt{\sum_i \chi_{\NN}^2 \left( \left\| {\bf x}_i \right\|_2 \right)} \\
 \end{array}
}
\]
where $\chi_{\NN}$ is as defined in the theorem.  It follows that:
\[
 {\small{
 \begin{array}{rl}
  \mathcal{R}_N \left( \unitnormballkrein{\NN} \,\right) &\!\!\!\!= \mathbb{E}_\nu \mathbb{E}_\epsilon \left[ \mathop{\sup}\limits_{f = \liip {\bf v}_{\NN}, \bg{\scriptstyle\featmapmkern} (\cdot) \riip{\bf g} \Big| {{{\bf v}_{\NN} \in \frkks{\NN}}, \atop {\liip {\bf v}_{\NN}, {\bf v}_{\NN} \riip{\bf g} \leq R_{\NN}}}} \left| \frac{1}{N} \sum_i \epsilon_i f \left( {\bf x}_i \right) \right| \right] \\
  &\!\!\!\!\leq \mathbb{E}_\nu \frac{\max \{ 1,R_{\NN}^d \}}{\sqrt{N}} \sqrt{\sum_i \chi_{\NN}^2 \left( \left\| {\bf x}_i \right\|_2 \right)} \\
  &\!\!\!\!\leq \frac{\max \{ 1,R_{\NN}^d \}}{\sqrt{N}} \sqrt{\mathbb{E}_\nu \left[ \sum_i \chi_{\NN}^2 \left( \left\| {\bf x}_i \right\|_2 \right) \right]} \\
  &\!\!\!\!\leq \frac{\max \{ 1,R_{\NN}^d \}}{\sqrt{N}} \sqrt{\int \chi_{\NN}^2 \left( \left\| {\bf x}_i \right\|_2 \right) d\nu \left( {\bf x} \right)} \\
 \end{array}
 }}
\]

Finally, we consider the special cases.  In the fully bounded case 
$\sigma_{d-1} (\xi) \leq 1$ we may bound the integral by replacing $\chi_{\NN}$ 
with $1$, so it follows that:
\[
 {\small{
 \begin{array}{rl}
  \mathcal{R}_N \left( \unitnormballkrein{\NN} \,\right) 
  &\!\!\!\!\leq \frac{\max \{ 1,R_{\NN}^d \}}{\sqrt{N}} \sqrt{\int d\nu \left( {\bf x} \right)} 
  = \frac{\max \{ 1,R_{\NN}^d \}}{\sqrt{N}} \\
 \end{array}
 }}
\]
In the unbounded case we use that $\sigma_q$ is Lipschitz and positive on 
$\infset{R}_+$ to obtain $0 \leq \sigma_q (\xi) \leq L_q \xi$ for all $\xi 
\in \infset{R}_+$, and hence:
\[
 {\small{
 \begin{array}{rl}
  \mathcal{R}_N \left( \unitnormballkrein{\NN} \,\right) 
  &\!\!\!\!\leq \left( d \sqrt{H} L \right)^d \frac{\max \{ 1,R_{\NN}^d \}}{\sqrt{N}} \sqrt{\int \left\| {\bf x}_i \right\|_2^2 d\nu \left( {\bf x} \right)} \\
 \end{array}
 }}
\]
and in the partially bounded case we observe that he definition of $\chi_{\NN}$ 
can be ``pinched off'' at $\sigma_q$ in a manner similar to the fully bounded 
case, while the remainder of the activation functions to the output contribute 
a Lipschitz constant and a width term, so:
\[
 {\small{
 \begin{array}{rl}
  \mathcal{R}_N \left( \unitnormballkrein{\NN} \,\right) 
  &\!\!\!\!\leq \left( d \sqrt{H_{[q+]}} L_{[q+]} \right)^{d-q-1} \frac{\max \{ 
  1,R_{\NN}^d \}}{\sqrt{N}} \\ 
 \end{array}
 }}
\]
where $H_{[q+]} = \GM (H_{q+1},\ldots,H_{d-1})$ and $L_{[q+]} = \GM (L_{q+1}, 
\ldots,L_{d-1})$ are geometric means.
\end{proof}

\subsection{Sparsity Analysis}

{\bf Theorem \ref{th:lownormbound} }
{\it
 For a given deep network satisfying our assumptions, using the notations 
 described, we have that $\| {\bf v}_{[q]} \|_\infty \leq dR_{\NN}$, $\| {\bf 
 v}_{\NN} \|_\infty \leq R_{\NN}^d$, and:
 \[
  \begin{array}{l}
   \left\| {\bf g}_{\NN} \odot {\bf v}_{\NN} \right\|_{2/d} \leq \left( H\bar{L} \right)^d \frac{DR_{\NN}}{H^2} \;\;\;\; \mbox{ (\ref{eq:coolnormbound})} \\
  \end{array}
 \]
 where $\| {\bf a} \|_\gamma = \sum_i |a_i|^\gamma$ is the $L_\gamma$-``norm'' 
 $\forall \gamma \in [0,1]$.
}
\begin{proof}
Using the Cauchy-Schwarz and theorem \ref{cor:collapse}:
\[
 {\small{
 \begin{array}{r}
  \sum_i \left| g_{\NN i} v_{\NN i} \right|^{\frac{2}{d}} 
  = \lmip \left| {\bf v}_{\NN} \right|^{\odot \frac{2}{d}} \rmip{1,|{\bf g}_{\NN}|} 
  = \lmip \bigodot_q \left| {\bf v}_{[q]} \right|^{\odot \frac{2}{d}} \rmip{1,|{\bf g}_{\NN}|} \\
  \begin{array}{l}
   = \lmip \left| {\bf v}_{[0]} \right|^{\odot \frac{2}{d}}, \left| {\bf v}_{[1]} \right|^{\odot \frac{2}{d}}, \ldots, \left| {\bf v}_{[d-1]} \right|^{\odot \frac{2}{d}} \rmip{d,|{\bf g}_{\NN}|} \\
   \leq \left( \prod_q \left| \lmip \left| {\bf v}_{[q]} \right|^{\odot \frac{2}{d}}, \left| {\bf v}_{[q]} \right|^{\odot \frac{2}{d}}, \ldots, \left| {\bf v}_{[q]} \right|^{\odot \frac{2}{d}} \rmip{d,|{\bf g}_{\NN}|} \right| \right)^{1/d} \\
   = \left( \prod_q \lip {\bf v}_{[q]}, {\bf v}_{[q]} \rip{|{\bf g}_{\NN}|} \right)^{1/d} \\
   \leq \left( \prod_q \frac{(H\bar{L})^d D}{H_q H_{q-1}} \left\| {\bf W}_{[q]} \right\|_{[q]}^2 \right)^{1/d} \\
   = \left( H\bar{L} \right)^d D \frac{1}{H^{2}} \frac{1}{D^{1/d}} \left( \prod_q \left\| {\bf W}_{[q]} \right\|_{[q]}^2 \right)^{1/d} \\
   = H^{d-2} \bar{L}^{d} D^{1-\frac{1}{d}} \left( \prod_q \left\| {\bf W}_{[q]} \right\|_{[q]}^2 \right)^{1/d} \\
   \leq \left( H\bar{L} \right)^d \left( \frac{1}{H} \right)^2 D^{1-\frac{1}{d}} \frac{1}{d} \sum_q \left\| {\bf W}_{[q]} \right\|_{[q]}^2 \\
   = \left( H\bar{L} \right)^d \left( \frac{1}{H} \right)^2 D^{1-\frac{1}{d}} R_{\NN} \\
   \leq \left( H\bar{L} \right)^d \frac{DR_{\NN}}{H^2} \\
  \end{array}
 \end{array}
 }}
\]
Likewise:
\[
 \begin{array}{l}
  \left\| {\bf v}_{\NN} \odot \left| {\bf g}_{\NN} \right|^{\odot \frac{1}{m}} \right\|_m 
   \!=\! \lmip {\bf v}_{\NN}, {\bf v}_{\NN}, \ldots \rmip{m,|{\bf g}_{\NN}|}^{1/m} \\
   \!=\! \Big( \bar{\sigma}_{d-1} \Big( {\sum}_{i_{d-2}} \left| W_{[d-1] 0,i_{d-2}} \right|^m \bar{\sigma}_{d-2} \ldots \\ 
   \bar{\sigma}_{d-2} \Big( \ldots {\sum}_{i_0} \left| W_{[1]i_1,i_0} \right|^m \bar{\sigma}_{0} \Big( \left\| {\bf W}_{[0],i_0,:} \right\|_m^m \Big) \Big) \Big) \Big)^{1/m} \\
   \!=\! \left( \bar{L}^d \prod_q \left\| {\bf W}_{[q]} \right\|_m^m \right)^{1/m} 
   \!=\! \bar{L}^{d/m} \prod_q \left\| {\bf W}_{[q]} \right\|_m \\
   \!\leq\! \left( \frac{\bar{L}^{1/m}}{d} \sum_q \left\| {\bf W}_{[q]} \right\|_m \right)^d 
   \!\leq\! \left( \bar{L}^{\frac{1}{m}} R_{\NN} \right)^d \\
 \end{array}
\]
so:
\[
 \begin{array}{l}
  \left\| {\bf v}_{\NN} \right\|_\infty = \mathop{\lim}\limits_{m \to \infty} \left\| {\bf v}_{\NN} \odot \left| {\bf g}_{\NN} \right|^{\odot \frac{1}{m}} \right\|_m
  \leq R_{\NN}^d
 \end{array}
\]
and:
\[
 \begin{array}{l}
  \left\| {\bf v}_{[q]} \odot \left| {\bf g}_{\NN} \right|^{\odot \frac{1}{m}} \right\|_m 
   \!=\! \lmip {\bf v}_{[q]}, {\bf v}_{[q]}, \ldots \rmip{m,|{\bf g}_{\NN}|}^{1/m} \\
   \!=\! \Big( \bar{\sigma}_{d-1} \Big( H_{d-2} \bar{\sigma}_{d-2} \Big( H_{d-3} \bar{\sigma}_{d-3} \Big( \ldots \\ 
         \ldots H_{q+1} \bar{\sigma}_{q+1} \Big( \sum_{i_q} \bar{\sigma}_q \Big( \left\| {\bf W}_{[q] i_q,:} \right\|_m^m \sigma_{q-1} \Big( H_{q-2} \ldots \\
         \ldots H_0 \bar{\sigma}_0 \Big( D \Big) \Big) \Big) \Big) \Big) \Big) \Big) \Big)^{1/m} 
   \!=\! \left( \frac{D\bar{L}^d \bar{H}^d}{H_q H_{q-1}} \left\| {\bf W}_{[q]} \right\|_m^m \right)^{1/m} \\
   \!=\! \frac{D^{1/m}\bar{L}^{d/m} \bar{H}^{d/m}}{(H_q H_{q-1})^{1/m}} \left\| {\bf W}_{[q]} \right\|_m 
   \!\leq\! \left( \frac{D\bar{L}^d \bar{H}^d}{H_q H_{q-1}} \right)^{\frac{1}{m}} d R_{\NN} \\
 \end{array}
\]
so:
\[
 \begin{array}{l}
  \left\| {\bf v}_{[q]} \right\|_\infty = \mathop{\lim}\limits_{m \to \infty} \left\| {\bf v}_{[q]} \odot \left| {\bf g}_{\NN} \right|^{\odot \frac{1}{m}} \right\|_m
  \leq d R_{\NN}
 \end{array}
\]
\end{proof}

{\bf Corollary \ref{cor:sparsity} }
{\it
 The total weight vector ${\bf g}_{\NN} \odot {\bf v}_{\NN}$ of the flat 
 representation is $\epsilon$-sparse - that is, there are at most $\lfloor 
 (H\bar{L} )^d \frac{DR_{\NN}}{H^2} \epsilon^{-\frac{2}{d}} \rfloor$ elements 
 in this vector with magnitude $| g_{\NN i} v_{\NN i} |$ greater than $\epsilon 
 \in (0, dR_{\NN} \| {\bf g}_{\NN} \|_\infty]$.
}
\begin{proof}
Suppose $n_{\max}$ elements of the total weight vector ${\bf g}_{\NN} \odot 
{\bf v}_{\NN}$ have $|g_{\NN i} v_{\NN i}| \geq \epsilon$.  To satisfy theorem 
\ref{th:lownormbound} we must have:
\[
 \begin{array}{l}
  n_{\max} \epsilon^{\frac{2}{d}} \leq \left( H\bar{L} \right)^d \frac{DR_{\NN}}{H^2}
 \end{array}
\]
and so:
\[
 \begin{array}{l}
  n_{\max} \leq \left( H\bar{L} \right)^d \frac{DR_{\NN}}{H^2} \epsilon^{-\frac{2}{d}}
 \end{array}
\]

Using theorem \ref{th:lownormbound} we have that $\| {\bf g}_{\NN} \odot {\bf 
v}_{\NN} \|_\infty \leq \| {\bf g}_{\NN} \|_\infty \| {\bf v}_{\NN} \|_\infty 
\leq \| {\bf g}_{\NN} \|_\infty R_{\NN}^d$, so we see that $|g_{\NN i} v_{\NN 
i}| \leq \| {\bf g}_{\NN} \|_\infty R_{\NN}^d$, which provides our upper bound 
on $\epsilon$.
\end{proof}

\end{document}